\documentclass{article}

\usepackage{amsmath,amsfonts,bm}

\def\ie{{\em i.e.,~}}
\def\eg{{\em e.g.,~}}

\newcommand{\wrt}{w.r.t.}

\def\eqref#1{equation~\ref{#1}}
\def\1{\bm{1}}

\def\ve{{\bm{e}}}

\def\vx{{\bm{x}}}
\def\vy{{\bm{y}}}

\def\mA{{\bm{A}}}

\def\mF{{\bm{F}}}

\def\mI{{\bm{I}}}

\def\mM{{\bm{M}}}
\def\mN{{\bm{N}}}

\def\mP{{\bm{P}}}
\def\mQ{{\bm{Q}}}
\def\mR{{\bm{R}}}

\def\mU{{\bm{U}}}
\def\mV{{\bm{V}}}

\def\mX{{\bm{X}}}
\def\mY{{\bm{Y}}}
\def\mZ{{\bm{Z}}}

\DeclareMathAlphabet{\mathsfit}{\encodingdefault}{\sfdefault}{m}{sl}
\SetMathAlphabet{\mathsfit}{bold}{\encodingdefault}{\sfdefault}{bx}{n}
\newcommand{\tens}[1]{\bm{\mathsfit{#1}}}
\def\tA{{\tens{A}}}

\def\tE{{\tens{E}}}

\def\tT{{\tens{T}}}

\def\gA{{\mathcal{A}}}

\def\gE{{\mathcal{E}}}
\def\gF{{\mathcal{F}}}
\def\gG{{\mathcal{G}}}

\def\gL{{\mathcal{L}}}

\def\gS{{\mathcal{S}}}

\def\gV{{\mathcal{V}}}

\def\sA{{\mathbb{A}}}

\def\sM{{\mathbb{M}}}

\def\sS{{\mathbb{S}}}

\def\emA{{A}}

\def\emM{{M}}
\def\emN{{N}}

\def\emP{{P}}
\def\emQ{{Q}}
\def\emR{{R}}

\def\emV{{V}}

\newcommand{\etens}[1]{\mathsfit{#1}}

\def\etA{{\etens{A}}}

\def\etE{{\etens{E}}}

\def\etT{{\etens{T}}}

\newcommand{\R}{\mathbb{R}}

\DeclareMathOperator*{\argmin}{arg\,min}

\usepackage{amsthm}
\usepackage[pagebackref=true]{hyperref}
\renewcommand*{\backrefalt}[4]{%
    \ifcase #1 \footnotesize{(Not cited.)}%
    \or        \footnotesize{(Cited on page~#2.)}%
    \else      \footnotesize{(Cited on pages~#2.)}%
    \fi}
    
\newtheorem{lemma}{Lemma}

\newtheorem{proposition}{Proposition}
\newtheorem{definition}{Definition}

\newcommand{\cF}{\mathcal{F}}

\newcommand{\cS}{\mathcal{S}}

\usepackage{fontawesome}
\usepackage[final]{neurips_2025}

\usepackage[utf8]{inputenc} % allow utf-8 input
\usepackage[T1]{fontenc}    % use 8-bit T1 fonts
\usepackage{hyperref}       % hyperlinks
\usepackage{url}            % simple URL typesetting
\usepackage{booktabs}       % professional-quality tables
\usepackage{amsfonts}       % blackboard math symbols
\usepackage{nicefrac}       % compact symbols for 1/2, etc.
\usepackage{microtype}      % microtypography
\usepackage{xcolor}         % colors

\usepackage{graphicx}
\usepackage{multirow}
\usepackage{multicol}
\usepackage{enumitem}
\usepackage{tcolorbox}
\usepackage{colortbl}
\usepackage[table]{xcolor}
\usepackage{url}
\usepackage{float}
\usepackage{subcaption}

\usepackage{hyperref}
\usepackage{caption}
\usepackage{ragged2e} 
\usepackage{tikz}
\definecolor{cyan}{cmyk}{.3,0,0,0}
\definecolor{LightCyan}{rgb}{0.88,1,1}
\definecolor{Gray}{gray}{0.95}

\usepackage{amsmath}
\usepackage{amssymb}
\usepackage{mathtools}
\usepackage{amsthm}
\usepackage{wrapfig}
\usepackage{adjustbox}
\usepackage{array}
\usepackage{tabularx}
\usepackage{longtable}
\usepackage[font=small]{caption}
\usepackage{mdframed}
\newmdtheoremenv[linecolor=black,linewidth=1pt,skipabove=10pt,skipbelow=10pt]{boxedtheorem}{Theorem}

\title{\textsc{ExGra-Med}: Extended Context Graph Alignment for Medical Vision-Language Models}
\usepackage[subfigure]{tocloft}

\vspace{-0.1in}
\author{%
  \textbf{Duy M. H. Nguyen} \textsuperscript{1,2,3}
  \quad
  \textbf{Nghiem T. Diep} \textsuperscript{1}$^*$
  \quad 
  \textbf{Trung Q. Nguyen} \textsuperscript{1}$^*$
  \quad
  \textbf{Hoang-Bao Le} \textsuperscript{1}\\
  \textbf{Tai Nguyen} \textsuperscript{1}
  \quad 
  \textbf{Tien Nguyen} \textsuperscript{4,5}
  \quad
  \textbf{TrungTin Nguyen} \textsuperscript{6,7}
  \quad
  \textbf{Nhat Ho} \textsuperscript{9}
  \quad
  \textbf{Pengtao Xie} \textsuperscript{10, 11} \\
  \textbf{Roger Wattenhofer} \textsuperscript{12}
  \textbf{James Zou} \textsuperscript{13}
  \quad
  \textbf{Daniel Sonntag} \textsuperscript{1,8}$^\dagger$
  \quad
  \textbf{Mathias Niepert} \textsuperscript{2,3}$^\dagger$ \vspace{0.1in} \\
  \textsuperscript{1}\,German Research Centre for Artificial Intelligence (DFKI),  \\
  \textsuperscript{2}\,Max Planck Research School for Intelligent Systems (IMPRS-IS),
\textsuperscript{3}\,University of Stuttgart,\\
\textsuperscript{4}\,University Medical Center Gottingen, 
\textsuperscript{5}\,Max Planck Institute for Multidisciplinary Sciences, \\
\textsuperscript{6}\,ARC Centre of Excellence for the Mathematical Analysis of Cellular Systems,\\ 
\textsuperscript{7}\,School of Mathematical Sciences, Queensland University of Technology,\\
\textsuperscript{8}\,University of Oldenburg,
\textsuperscript{9}\,University of Texas at Austin, \\
\textsuperscript{10}\,University of California San Diego,
\textsuperscript{11}\,MBZUAI,  
\textsuperscript{12}\,ETH Zurich,  \textsuperscript{13}\,Stanford University \\[3pt]
* Co-second contribution \quad $^\dagger$ Co-senior authors. \\
\faGithubSquare \enspace \texttt{\href{https://exgra-med.github.io/}{Exgra-Med}}
}

\begin{document}

\vspace{-0.5in}
\maketitle

\vspace{-0.2in}
\begin{abstract}
\vspace{-0.05in}
State-of-the-art medical multi-modal LLMs (med-MLLMs), such as \textsc{LLaVA-Med} and \textsc{BioMedGPT}, primarily depend on scaling model size and data volume, with training driven largely by autoregressive objectives. However, we reveal that this approach can lead to weak vision-language alignment, making these models overly dependent on costly instruction-following data. To address this, we introduce \textsc{ExGra-Med}, a novel multi-graph alignment framework that jointly aligns images, instruction responses, and extended captions in the latent space, advancing semantic grounding and cross-modal coherence. To scale to large LLMs (e.g., LLaMa-7B), we develop an efficient end-to-end training scheme using black-box gradient estimation, enabling fast and scalable optimization. Empirically, \textsc{ExGra-Med} matches \textsc{LLaVA-Med}’s performance using just 10\% of pre-training data, achieving a 20.13\% gain on VQA-RAD and approaching full-data performance. It also outperforms strong baselines like \textsc{BioMedGPT} and \textsc{RadFM} on visual chatbot and zero-shot classification tasks, demonstrating its promise for efficient, high-quality vision-language integration in medical AI. 
\end{abstract}
\vspace{-0.1in}
\addtocontents{toc}{\protect\setcounter{tocdepth}{-1}}
\section{Introduction}
\vspace{-0.05in}
Generic Multi-Modal Large Language Models (MLLMs) such as GPT-4V \citep{achiam2023gpt}, LLaVa \citep{llava}, and Next-GPT \citep{wu2023next} unify text, image, and audio processing for tasks like captioning and visual reasoning. A key component in training MLLMs is instruction-following (IF) data \citep{lou2023comprehensive}, which involves complex, often multi-turn interactions grounded in image content \citep{sun2024parrot}. In the medical domain, specialized IF datasets, including medical images, clinical notes, and diagnostic criteria, have been curated to adapt general-purpose MLLMs while leveraging their pre-learned knowledge and minimizing training costs \citep{xie2024medtrinity}. For example, LLaVA-Med \citep{llava-med} samples 600K image-text pairs from PMC-15M \citep{pmc-15m}, using GPT-4 to generate around 60K multi-modal IF examples. The training involves two pretraining steps: (i) aligning vision encoders and language decoders via projection layers, and (ii) jointly training the model (excluding the vision encoder) on medical IF data using an auto-regressive objective. The resulting model is then fine-tuned for downstream medical tasks.

\begin{wrapfigure}{r}{0.48\textwidth}
    \centering % Centers the figure within the wrapfigure
    \vspace{-0.2in}
    \includegraphics[width=0.48\textwidth]{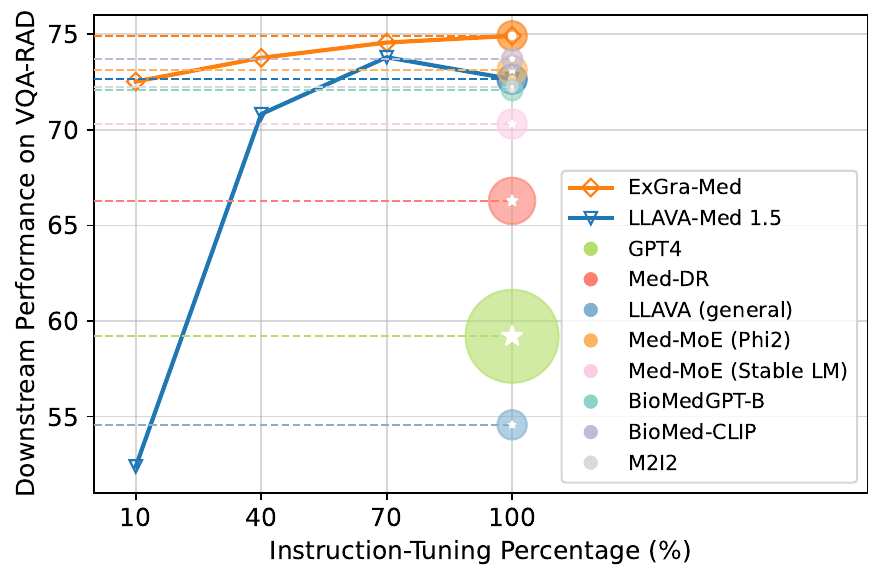}
    \vspace{-0.13in}
    % \caption{\textit{Illustration of the data-hungry behavior of auto-regressive modeling} in LLaVA-Med against ExGra-Med when varying pre-training instruction following (IF) data size. \textcolor{blue}{After being pre-trained at each IF rate, two models are fine-tuned on the same VQA-RAD training set}. At 100\%  IF pretraining, LLaVa-Med and ExGra-Med are benchmarked with other architectures that are also fine-tuned on the same VQA-RAD training set, except the GPT-4. The radius at each circle illustrates its model parameters.}
    \caption{Our \textsc{ExGra-Med} versus LLaVA-Med across varying instruction-following (IF) pre-training data sizes, \textbf{highlighting the data-hungry behavior of auto-regressive modeling}. Both \textit{models are fine-tuned on the same VQA-RAD training set} after the pre-training stage at each IF rate. At 100\% IF pre-training, ExGra-Med and LLaVA-Med are benchmarked against other state-of-the-art models, \textit{all fine-tuned on the same VQA-RAD training set} (except GPT-4, which is evaluated without fine-tuning). Circle radius represents the number of model parameters.}
    \vspace{-0.1in}
    \label{fig:auto-regressive-less}
\end{wrapfigure}

Following the above approach, most later works have focused on scaling up the amount of medical IF data \citep{xie2024medtrinity,zhang2023biomedgpt,he2024meddr} or increasing the model size by incorporating larger vision encoders or language decoders \citep{radfm,jiang2024moe} while relying on the same standard autoregressive learning scheme. Contrary to this, we question the effectiveness of autoregressive objective functions when learning medical-MLLM with IF data. Surprisingly, \textit{our findings reveal that autoregressive learning is highly data-hungry during pre-training}, i.e., without sufficient medical IF samples, model performance plummets for downstream tasks, \textit{even after fine-tuning}. To illustrate this,
we pre-trained LLAVA-Med using only $10\%$ of the data and compared it to the version trained on $100\%$. Both models were fine-tuned on two medical visual question-answering tasks - VQA-RAD \citep{lau2018dataset} and PathVQA \citep{he2020pathvqa} - and their average performance on open- and close-ended questions are compared. The results show a dramatic decline: from $72.64\%$ to $52.39\%$ on VQA-RAD (Figure \ref{fig:auto-regressive-less}) and from $64.06\%$ to $56.15\%$ on PathVQA (Table \ref{tab:model_comparison_10percent}). This underscores the instability of medical-MLLM trained with autoregressive methods and highlights the problem that these methods require the curation of enough medical IF data to achieve satisfactory performance.

To address the limitations of autoregressive training under limited instruction-following data, we propose \textsc{ExGra-Med}, a novel multi-graph alignment framework that strengthens cross-modal understanding in multi-modal large language models (MLLMs). At the core of our approach is the construction of three modality-specific graphs: one for visual features extracted by a vision encoder, and two for different textual variants of the instruction. These graphs represent semantic relationships within and across modalities, and we formulate a combinatorial multi-graph alignment problem to learn consistent triplet-level associations between the image, its instruction, and a semantically enriched variant. This alignment objective is jointly optimized with the autoregressive language modeling loss, enabling the model to enhance semantic depth, coherence, and instruction-following ability. To generate the enriched instruction variant, we use a frozen LLM (GPT-4 \cite{achiam2023gpt}) to produce a contextually extended version of each instruction that highlights key concept relationships without altering the original intent. The vision encoder and language model (LLaMa \cite{touvron2023llama}) then process the image, instruction, and extension independently to produce node embeddings used in the alignment step. Unlike naive data augmentation, our use of GPT-4 enriches supervision and facilitates fine-grained graph-based correspondence learning across modalities (Figure~\ref{fig:overview}).

Our method stands apart from prior multi-modal alignment approaches for LLMs \citep{park2024bridging,li2023blip,chen2023videollm} in two key ways. (i) Instead of merely learning projection layers between frozen vision and language models, we \textit{train the LLM directly} via a multi-graph alignment framework. (ii) We also \textit{extend pairwise contrastive learning} by integrating global structural graph constraints, enabling alignment not just between individual image-caption pairs but across entire datasets. This graph-based design captures both feature and relational consistency, critical for handling similar entities in medical data. While multi-graph alignment is typically non-differentiable \citep{rolinek2020deep} and computationally intensive \citep{pevzner1992multiple}, we address these challenges using implicit maximum likelihood estimation \citep{niepert2021implicit,minervini2023adaptive}. This enables efficient gradient-based training over large LLMs (e.g., LLaMa-7B) using a barycenter graph \citep{agueh2011barycenters} for alignment, allowing our model to scale effectively while preserving strong alignment performance.

In summary, we make the following key contributions:
\begin{itemize}[itemsep=0.02in]
\vspace{-0.1in}
    \item We reveal the data-demanding nature of autoregressive modeling in pre-training medical-MLLM (LLaVa-Med), showing that insufficient instruction-following data leads to significant performance drops on downstream tasks, even after fine-tuning.
    \item We introduce a new multi-graph alignment objective that establishes triplet correlations among images, their instruction-following context, and their enriched versions. Additionally, we developed an efficient solver for training with LLMs and outlined theoretical properties related to distance and the shortest path in the geodesic space of multi-modal graphs.
    \item We empirically demonstrate that using a small amount of pre-training data, \textsc{ExGra-Med} can achieve performance comparable to LLaVa-Med trained on $100\%$ data. Additionally, when trained on larger datasets, \textsc{ExGra-Med} \textit{outperforms} several state-of-the-art \textit{med-MLLMs} and \textit{multi-modal pre-training} algorithms across three Medical VQA tasks, medical visual chat, and the average zero-shot image classification performance on 23 datasets.
\end{itemize}

\vspace{-0.1in}
\section{Related Work}
\vspace{-0.1in}
\textbf{Medical Multi-Modal LLMs.}
% Recent developments in medical MLLMs like Biomed-GPT \citep{zhang2023biomedgpt}, MedFlamingo~\citep{moor2023med}, Med-Dr~\citep{he2024meddr}, LLAVA-Med~\citep{llava-med}, or close-models as Med-PaLMs \citep{singhal2023towards,tu2024towards} are transforming healthcare by integrating diverse data types and scaling medical instruction data. Biomed-GPT excels in handling multiple biomedical modalities, while MedFlamingo focuses on few-shot learning for medical visual question answering in data-limited settings. Med-Dr aids clinical decision-making by synthesizing multi-modal data, and LLAVA-Med leverages large-scale biomedical image-text pairs for enhanced performance. 
Recent developments in medical-MLLM like Biomed-GPT \citep{zhang2023biomedgpt}, MedFlamingo \citep{moor2023med}, Med-Dr \citep{he2024meddr}, LLAVA-Med \citep{llava-med}, and Med-PaLMs \citep{singhal2023towards,tu2024towards} are transforming healthcare by integrating diverse data types and scaling medical instruction data. Biomed-GPT excels with multiple biomedical modalities, MedFlamingo focuses on few-shot learning for medical visual question answering, 
% Med-Dr supports clinical decision-making with multi-modal data, 
and LLAVA-Med leverages large-scale biomedical image-text pairs for improved performance. Commonly, these models emphasize scaling medical instruction data and increasing model parameters to enhance accuracy and applicability in real-world medical scenarios. In contrast, \textit{our approach examines the widely used autoregressive pre-training algorithms} and demonstrates that incorporating enriched context multi-graph alignment of existing instruction samples can significantly enhance medical-MLLM performance without requiring larger models or extensive datasets.

%%%%%%%%%%%%%%%%%%%%
%\vspace{-0.1in}
\textbf{Visual Instruction Tuning.}
%\vspace{-0.1in}
Visual instruction tuning techniques aim to bridge the gap between frozen vision-language models and frozen LLMs trained on unimodal data, enabling them to work effectively in a multi-modal context. These methods involve (i) learning a multi-layer perceptron (MLP) layer to map embeddings from the vision model to the language model as LLaVa \citep{llava}, VideoLLM \citep{chen2023videollm}; (ii) using adapter-based adjustment as LLaMa-adapter \citep{zhang2024llama}, Voxposer \citep{huang2023voxposer}, or (iii) learning multi-modal perceiver by gated cross-attention \citep{alayrac2022flamingo} or Q-Former as in BLIP-2 \citep{li2023blip}. Pre-training algorithms to train these models can be combined with both auto-regressive and contrastive learning \citep{park2024bridging,zhai2023sigmoid} or image-text matching as in \citep{li2022blip,li2023blip}. Our algorithm differs from those by \textit{focusing on directly training LLMs rather than lightweight projectors}. This requires a fast solver capable of efficiently handling forward and backward passes through large-scale LLMs with extensive parameters.

%\vspace{-0.1in}
\textbf{Vision-language Pretraining Algorithm.}
% Pre-training algorithms in the literature are used to train vision-language models such as CLIP. They include generative-based algorithms such as masking prediction, which is used in training the BERT model, or auto-regressive algorithms to predict sequential text in the LLMs model. There is another direction based on discriminative, which learns a contrastive distance between image-text pairs, optimal transport, or clustering conditions. Our objective function differs and generalizes those works through combinatorial graph-matching formulation defined on cross-domain graphs. In that sense, LVM-Med is the most relevant to us; however, this method is designed to match problems in vision domains while we work on the alignment between images, instruction data, and the longer context of instruction data.
%\vspace{-0.1in}
Pre-training algorithms commonly applied for vision-language models, like CLIP \citep{radford2021learning}, employ various strategies.  Among these, generative approaches are widely used, including masked prediction in language models \citep{devlin2018bert,song2020mpnet}, or autoregressive algorithms that predict sequential text in LLMs \citep{llava,zhang2024llama}.  Another prominent direction focuses on discriminative methods, which learn contrastive distances between image-text pairs \citep{liu2023contrastive,zhai2023sigmoid,khan2023contrastive}, optimal transport \citep{chen2022plot,nguyen2024dude}, or impose clustering constraints \citep{park2024bridging}. \textit{Toward multi-modal learning across three or more modalities}, there also exist works such as PAC-S\cite{sarto2023positive}, GeoCLAP \cite{khanal2023learning}, and IMAGEBIND \cite{girdhar2023imagebind}, extending the CLIP or InfoNCE \cite{oord2018representation}  to align embeddings across multiple modalities simultaneously. 

Our function departs from these by \textit{generalizing them into a combinatorial graph-matching formulation across cross-domain graphs}. While LVM-Med \citep{mh2024lvm} is the most similar to our approach, it targets alignment within vision tasks, whereas we align images, instruction-following data, and extended contextual information. We provide a more comprehensive comparison between \textsc{ExGra-Med} and related works in the Appendix.

% Solving graph alignment between $k$ domains ($k \geq 3$) is computationally expensive in general. Current works have applied multi-marginal in optimal transport, Wasserstein barycenter, or making assumptions on the multi-adjacency matrix to relax the problem. However, these approaches are restricted to small-scale problems or require several steps in the solver to produce matching solutions, which makes them unsuitable in our case when training with LLMs. In contrast, our algorithm leverages heuristic solvers and modern gradient estimation techniques for black-box solvers to make the formulation scale on LLMs.

% Graph alignment across $K$ domains ( $K \geq 3$ ) is notoriously computationally intensive. Existing approaches, such as multi-marginal optimal transport \citep{complexity_multimarginalOT,multimarginalOT_equivalence, piran2024contrasting}, Wasserstein barycenters \citep{nguyen2024structure}, or assumptions on multi-adjacency matrices ~\citep{bernard2019hippi,swoboda2019convex}, relax the problem but are limited to small-scale tasks and involve multiple solver steps. This makes them impractical for our case, where we need to train LLMs efficiently. In contrast, our method leverages heuristic solvers \citep{swoboda2017study,rolinek2020deep} and modern gradient estimation techniques for black-box optimization \citep{niepert2021implicit,minervini2023adaptive}, allowing it to scale effectively for large language models.

\textbf{Scalable Multi-Graph Alignment.} Graph alignment across $K$ domains ($K \geq 3$) is highly computationally intensive. Current methods, such as multi-marginal optimal transport \citep{complexity_multimarginalOT, piran2024contrasting}, Wasserstein barycenters \citep{nguyen2024structure}, and multi-adjacency matrix assumptions \citep{bernard2019hippi,swoboda2019convex}, relax the problem but are limited to small-scale tasks and require multiple solver steps, making them inefficient for LLM training. In contrast, our algorithm leverages heuristic solvers \citep{swoboda2017study,rolinek2020deep} and modern gradient estimation techniques for black-box optimization \citep{niepert2021implicit,minervini2023adaptive}, enabling scalable and efficient performance for large language models. A deeper analysis of this factor is discussed in the Ablation study.
\vspace{-0.1in}

\section{Multi-graph Alignment Learning}
\vspace{-0.1in}
We denote the vision encoder, projector, and large-language  model (LLM) models are $f_{\theta}(.),\, h_{\phi}(.),\, g_{\sigma}(.)$, respectively.  Figure \ref{fig:overview} illustrates our \textsc{ExGra-Med} algorithm, which learns parameters for these models by solving a triplet alignment between modalities in instruction tuning data. Below, we summarize the notations used before describing each component in detail.

{\bf Notation.} Given any tensor $\tT=(\etT_{i,j,k,l})$ and matrix $\mM = (\emM_{k,l})$, we use $\tT\otimes \mM$ to denote the tensor-matrix multiplication, \ie~the matrix $(\sum_{k,l}\etT_{i,j,k,l}\emM_{k,l})_{i,j}$. Given
$\mY = [\vy_1, \vy_2,...,\vy_N] \in \R^{N\times d}$, we define $\mathbb{E}(\mY) = \frac{1}{N}\sum_{i=1}^{N} \vy_{i} \in \R^{d}$.
Moreover, we define the matrix scalar (or inner) product associated with the Frobenius norm between two matrices $\mM = (\emM_{i,j})$ and  $\mN = (\emN_{i,j})$ as $\langle \cdot, \cdot \rangle$, \ie~$\langle \mM, \mN\rangle = \sum_{i,j} \emM_{i,j} \emN_{i,j}$. 
We write $[M] = \{1, 2, \ldots, M\}$ for any natural number $M$.
\vspace{-0.05in}
\subsection{Extended context enriched medical instruction following data}
\vspace{-0.05in}
Recent research has demonstrated that incorporating longer context significantly enhances LLMs’ ability to process complex inputs and improves instruction-following by retaining more relevant information \citep{liu2024lost,an2024training,pawar2024and}. Building on this insight, we expand medical instruction-following data by generating \textit{contextually enriched paraphrased versions of existing samples}, offering a complementary perspective to the original dataset. There are two key motivations for incorporating both original and extended captions in our multi-graph alignment framework. (i) First, aligning with original captions preserves precise, domain-specific details, while extended captions enhance semantic richness, leading to more robust image embeddings. (ii) Second, this approach helps the LLM generate contextually rich yet semantically consistent responses, improving alignment across diverse linguistic forms (Table \ref{tab:logra_ablation}).

% First, this dual-caption strategy enhances the robustness of image embeddings: aligning images with original captions preserves precise, domain-specific details, while extended captions encourage richer semantic understanding for a more comprehensive visual representation. Second, it improves the LLM’s capacity to generate contextually rich responses while maintaining strong semantic consistency with the original content, promoting better alignment across varied linguistic formulations.
% providing an additional enriched perspective of the original data. 
\vspace{-0.1in}
\begin{wrapfigure}{r}{0.4\textwidth}
    \centering % Centers the figure within the wrapfigure
    \vspace{-0.2in}
    \includegraphics[width=0.4\textwidth]{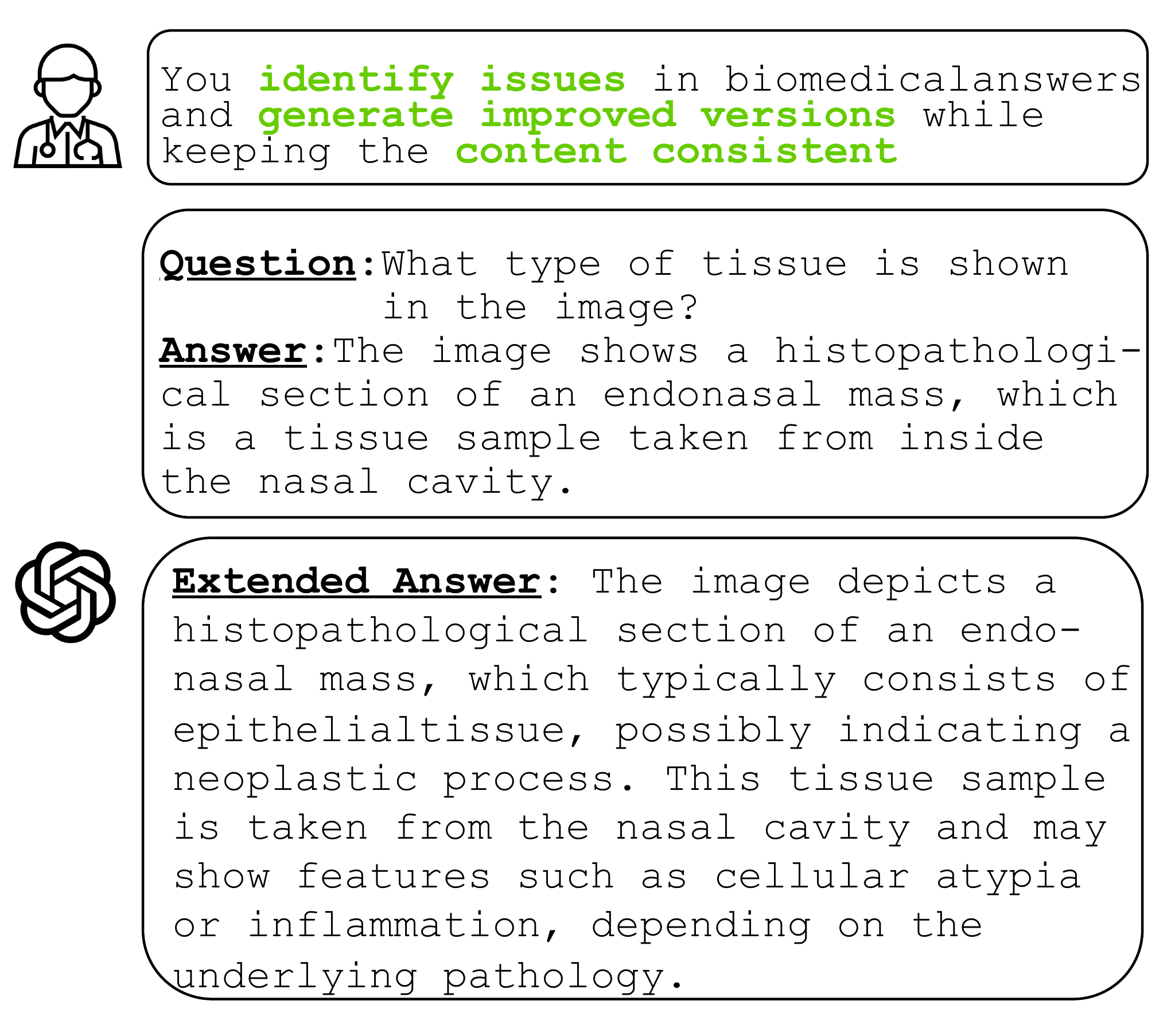}
    \caption{Illustration for creating the extended context instruction-following data powered by GPT-4o.} 
    \vspace{-0.1in }
    \label{fig:long-context-example}
\end{wrapfigure}
In particular, a typical instruction sample includes $\{\mX_{v}, [\mX_{q}^{1},\,\mX_{a}^{1}],...,[\mX_{q}^{L},\,\mX_{a}^{L}]\}$ where $\mX_{v}$ is an input image, $\mX_{q}^{l}$ a question, and $\mX_{a}^{l}$ an answer at round $l$ in multi-round $L$ of a conversation. In the medical domain, most of the questions are generic, and the information answer usually covers the question, so we only focus on extending the answer $\mX_{a}$. We leverage the \texttt{GPT} API  with a \texttt{prompt} to form an extended context for each $\mX_{a}^{l}$ by:
\setlength{\abovedisplayskip}{3pt}
\setlength{\belowdisplayskip}{3pt}
\begin{equation}
    \mX_{ae}^{l} = \mathrm{GPT}\left(\mX_{q}^{l}, \mX_{a}^{l},\mathrm{prompt}\right),\  \forall l \in [L].
    \label{eq:gpt}
\end{equation}
% \[\mX_{at}^{l} = \mathrm{GPT}\left(\mX_{q}^{l}, \mX_{a}^{l},\mathrm{prompt}\right)\  \forall l \in L. \]\ 
The details for $\mathrm{prompt}$ are presented in the Appendix. In short, we ask GPT to provide additional explanations for concepts that appeared in the original answer $\mX_a$ while keeping the content consistent. An example output 
for $\mX_{ae}^{l}$ is illustrated in Figure \ref{fig:long-context-example} and Table \ref{tab:extend_long_context_examples}. It's worth noting that other frozen LLMs like Gemini are also valid in our method (Table \ref{tab:exgra-med-other-llms}). 

\vspace{-0.1in}
\subsection{Multi-graph construction on vision-language  embedding}
\label{sec:graph_construction}
\vspace{-0.05in}
Each \textit{image} $\mX_{v} \in \R^{3\times H \times W}$, where $(H, W)$ are the original spatial dimensions, is divided it into a sequence of visual patches $\mU = [u_{i}]_{i=1}^{N}$ with $N = (H \times W)/U$
with $U$ the patch size. Using a pre-trained ViT model $f_{\theta}$, we extract patch-wise features as $\mV = f_{\theta}(\mU) \in \R^{N\times d_v}$ and apply another projector to map it into the projected embedding $\mZ = h_{\phi}(\mV) \in \R^{N \times d}$. We then pool the features from the image patches to define a global description as 
$\mZ_{v} = \mathbb{E}(\mZ) \in \R^{d}$. For each \textit{language input} $\mX_{c}^{l} \in \{\mX_{a}^{l},\mX_{ae}^{l}\}$ with $c \in \{a,ae\}$, we assume it has $M$ tokens, i.e., $\mX_{c}^{l} = [\vx_j]_{j=1}^{M} \in \R^{M}$, and feed it into the \texttt{LLM} model to extract a set of embedding $\mZ_{c}^{l} = g_{\sigma}([\vx_j]_{j=1}^{M}) = [\ve_j]_{j=1}^{M} \in \R^{M\times d}$. We subsequently concatenate all multi-round $L$ in each single instruction tuning to define $\mZ_{c} = \textstyle \frac{1}{L} \sum_{l=1}^{L} \mathbb{E}(\mZ_{c}^{l})$ which collects average text embedding of original answers ($c=a$) and their longer-context extended versions ($c=ae$) respectively. Though we adapt simple average pooling feature mechanisms, it remains an effective approach with a clear observed margin of separation between the distinct distributions (Table \ref{tab:logra_ablation} in the Ablation study).

Given a batch size of $B$ instruction-tuning samples, we now construct three graphs $\gG_{v} =(\gV_{v}, \gE_{v})$,  $\gG_{a} =(\gV_{a}, \gE_{a})$, and $\gG_{ae} =(\gV_{ae}, \gE_{ae})$ representing for visual image features, text embedding encoded by LLM for original answers and their extended context embedding extended by GPT. Specifically, for each triplet pair $\{\mX_{v}^{(k)}, [\mX_{a}^{l}]^{(k)}, [\mX_{ae}^{l}]^{(k)}\}_{k},\, (k \in [B])$, we add a node representing $\mX_{v}^{(k)}$ to $\gV_{v}$, a node for $[\mX_{a}^{l}]^{(k)}$ to $\gV_{e}$, and finally a node for $[\mX_{ae}^{l}]^{(k)}$ to $\gV_{ae}$. This results in a set of nodes $\gV_{v} = \{\mX_{v}^{(1)},...,\mX_{v}^{(B)}\}$; $\gV_{c} = \{[\mX_{c}^{l}]^{(1)},...,[\mX_{c}^{l}]^{(B)}\}$ for each $c \in \{a,ae\}$. We equip node-level feature matrices for these graphs using their embedding computed above, i.e., $\mF_{v} = \{\mZ_{v}^{(1)},...,\mZ_{v}^{(B)}\}$, $\mF_{c} = \{\mZ_{c}^{(1)},...,\mZ_{c}^{(B)}\}$. The edges for $\gE_{v}, \gE_{c}$ afterward can be created through the k-nearest neighbors algorithm given the feature node matrices $\mF_{v},\,\mF_{c}$. Finally, we can run a message-passing network $m_{\alpha}(.)$ on three built graphs to learn richer node representations. This approach has proven effective for representation learning \citep{tang2022unifying,ju2024comprehensive}, resulting in aggregated feature-node matrices as $\{\hat{\mZ}_{s}^{(1)},...,\hat{\mZ}_{s}^{(B)}\} = m_{\alpha}(\mF_{s}, \gE_{s})$, with $s \in \{v,a, ae\}$.
\vspace{-0.1in}
\begin{figure}[!hbt]
\centering
\includegraphics[width=1.1\textwidth, trim=0 0 2.2cm 0, clip]{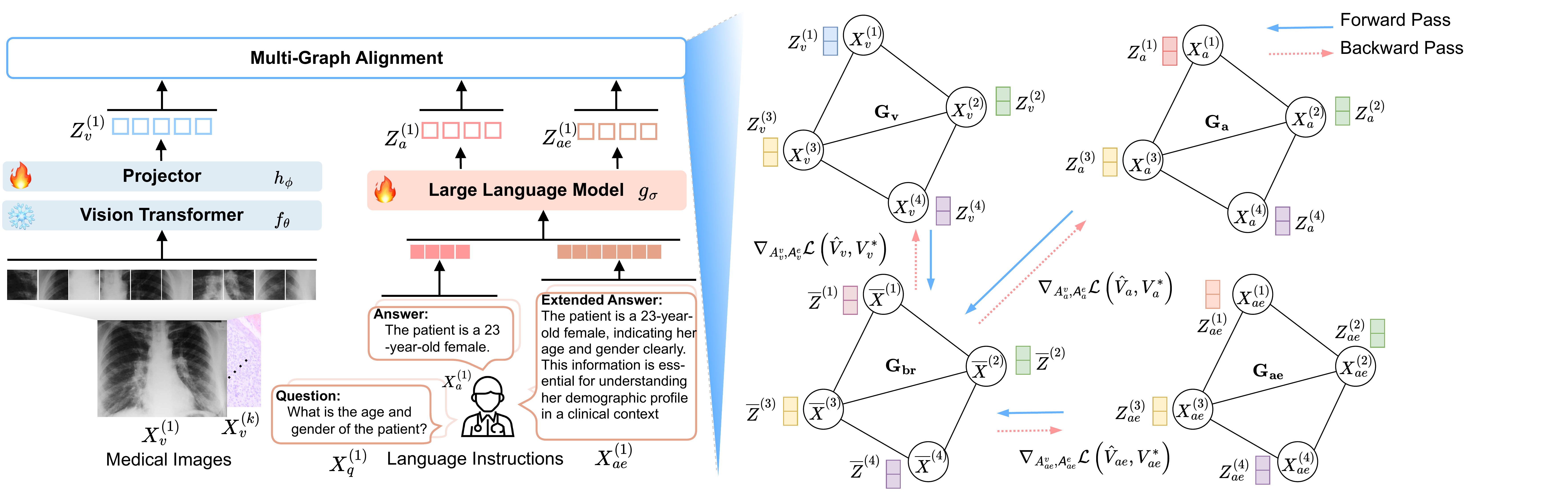}
\vspace{-0.1in}
\caption{Overview of \textsc{ExGra-Med}: The large language model $g_{\sigma}$ and the projector $h_{\phi}$ are trained jointly by aligning a triplet of modalities - input image, instruction-following data, and extended captions - through a structure-aware multigraph alignment (Eq.(\ref{eq_GM})). This alignment operates over graphs $\mathcal{G}_v$, $\mathcal{G}a$, and $\mathcal{G}{ae}$, representing the visual, instruction, and extended textual information, respectively, via a shared barycenter graph. The entire model is optimized end-to-end using modern black-box gradient estimation techniques to enable efficient learning across modalities \cite{niepert2021implicit,minervini2023adaptive}.}
\label{fig:overview}
\vspace{-0.2in}
\end{figure}
\subsection{Second-order graph alignment problem}
\vspace{-0.05in}
We first provide background information about the second-order graph alignment problem between two arbitrary graphs $\gG_1=(\gV_1,\gE_1)$ and $\gG_2=(\gV_2,\gE_2)$, which is known as quadratic assignment problem. It occurs in several problems in vision and computer graphics to find correspondences between two graph structures under constraints on the \textit{consistency of node features and the graph structures }\citep{zanfir2018deep,haller2022comparative,ehm2024partial}.

We denote by $\mV \in \{0,1\}^{|\gV_1||\gV_2|}$, with $|\gV_1| = M$ and $|\gV_2| = N$, the indicator matrix  of matched vertices, that is, $\emV_{i,j} = 1$ if a vertex $v_i \in \gV_1$ is matched with $v_j\in \gV_2$ and  $\emV_{i,j} = 0$ otherwise. That is, $\mV$ is a binary matrix with exactly one non-zero entry in each row and column.
Similarly, we set $\tE \in \{0,1\}^{|\gE_1||\gE_2|}$ as the indicator tensor of match edges, that is, $\etE_{i,k,j,l} = 1$ if  $\emV_{i,j} = 1$ and  $\emV_{k,l} = 1$ and $\etE_{i,k,j,l} = 0$ otherwise. This implies that the tensor $\tE$ is fully determined by the matrix $\mV$, that is, $\etE_{i,k,j,l} = \emV_{i,j} \emV_{k,l}$. 
We also define the vertex affinity matrix and edge affinity tensor as $\mA^v \in \R^{|\gV_1||\gV_2|}$ and $\tA^e \in \R^{|\gE_1||\gE_2|}$, respectively. 
The set $\gA(\gG_1,\gG_2)$ indicates for all admissible pairs $(\mV,\tE)$ that encode a valid matching between $\gG_1$ and $\gG_2$. 
\iffalse
If graphs $\gG_1$ and $\gG_2$ have $M$ and $N$ vertices, respectively, then the set of all admissible pairs is defined as follows:
\fi
\setlength{\abovedisplayskip}{2pt}
\setlength{\belowdisplayskip}{2pt}
% \begin{equation}
% {\footnotesize
%     \begin{align}
%     \gA(\gG_1,\gG_2) = \left\{\mV \in \{0,1\}^{M\times N}:\sum_{i=1}^M\emV_{i,j} = 1,\sum_{j=1}^N\emV_{i,j} = 1\right\}.
%     \end{align}
% \end{equation}
\begin{equation}
    {\footnotesize
    \begin{aligned}
    \gA(\gG_1,\gG_2) = \left\{\mV \in \{0,1\}^{M\times N}:\sum_{i=1}^M\emV_{i,j} = 1,\sum_{j=1}^N\emV_{i,j} = 1\right\}.
    \end{aligned}
    }
\end{equation}

% \vspace{-0.1in}
The second-order graph alignment (SGA) problem now is defined as: 
\setlength{\abovedisplayskip}{4pt}
\setlength{\belowdisplayskip}{4pt}
\begin{align}\label{eq_GM}
    \text{SGA}(\mA^v,\tA^e) = \argmin_{\mV \in \gA(\gG_1,\gG_2)} \langle\mA^v+\tA^e\otimes \mV, \mV\rangle \nonumber \\=\argmin_{\mV \in \gA(\gG_1,\gG_2)}\sum_{i,j}\emA^v_{i,j}\emV_{i,j}+\sum_{i,j,k,l}\etA^e_{i,j,k,l}\emV_{i,j} \emV_{k,l}.
\end{align}
\iffalse
We now establish the second-order graph alignment (SGM) problem as: 
\setlength{\abovedisplayskip}{4pt}
\setlength{\belowdisplayskip}{4pt}
\begin{align}\label{eq_GM}
    \text{SGA}(\mA^v,\tA^e) &= \argmin_{\mV \in \gA(\gG_1,\gG_2)} O(\mA^v,\tA^e,\mV) \text{ where } O(\mA^v,\tA^e,\mV) = \langle\mA^v+\tA^e\otimes \mV, \mV\rangle.
\end{align}
Note that the objective function in Eq.~(\ref{eq_GM}) is expanded as sum of vertex and edge costs:
\begin{align}\label{eq_GM_expa}
   O(\mA^v,\tA^e,\mV) = \sum_{i,j}\emA^v_{i,j}\emV_{i,j}+\sum_{i,j,k,l}\etA^e_{i,j,k,l}\emV_{i,j} \emV_{k,l}.
\end{align}
\fi
\vspace{-0.26in}
\subsection{Scalable Multi-graph Alignment}
\label{sec:scalable-solver}
\vspace{-0.1in}
Our aim is to solve the graph alignment between $\gG_{v}, \gG_{a}$, and $\gG_{ae}$ to form a triplet constraint between input image embedding, its original instruction embedding, and the extended version ones. However, solving a structure-aware graph alignment between $K$ domains ($K \geq 3$) is computationally expensive. One potential solution is to perform pairwise graph alignments $\binom{K}{2}$ times, as shown in Eq.~(\ref{eq_GM}), while applying specific constraints to maintain consistency between correspondences \citep{bernard2019hippi,swoboda2019convex} (Table \ref{tab:logra_ablation}). However, these approaches become impractical as  $K$  increases, making them unsuitable for large-scale problems or the integration of multi-modal inputs such as text, images, audio, or medical health records.

Another direction leverages the barycenter concept from optimal transport \cite{guo2020fast,altschuler2022wasserstein}, which identifies a central distribution that minimizes the weighted sum of Wasserstein distances to the given input distributions. We follow this idea to reformulate the alignment of $K$  graphs into $K$  separate alignments with a barycenter graph. Unlike previous unsupervised methods that estimate the barycenter before aligning, 
we directly define the barycenter using known triplet pairs across the three graphs. This significantly reduces complexity, making our solver more efficient in LLM settings.

Specifically, we define a new barycenter graph $\gG_{br} = (\gV_{br},\gE_{br})$ where $\gV_{br} = \left\{v_{br}^{(1)},...,v_{br}^{(B)}\right\}$ with $v_{br}^{(k)} = \overline{\mX}^{(k)} =\{\mX_{v}^{(k)}, [\mX_{a}^{l}]^{(k)}, [\mX_{ae}^{l}]^{(k)}\}$ and  a correspondence feature node as $\mF_{br} =  \dfrac{1}{3}\left\{\sum_{s}\hat{\mZ}_{s}^{(1)},...,\sum_{s}\hat{\mZ}_{s}^{(B)}\right\}$ with $s \in \{v,a,ae\}$. The edge set $\gE_{br}$ is formed similarly to another graph by running the k-nearest neighbor on feature node $\mF_{br}$. We now state the multi-graph alignment as:
\setlength{\abovedisplayskip}{2pt}
\setlength{\belowdisplayskip}{2pt}
\begin{equation}\label{eq:multi-graph}
    \text{SGA}(\mA_{s}^v,\tA_{s}^e) = \argmin_{\mV_{s} \in \gA(\gG_{s},\gG_{br})}  \sum_{s \in \{v,a,ae\}} \, \langle\mA_{s}^v+\tA_{s}^e\otimes \mV_{s}, \mV_{s}\rangle,
\end{equation}
where $\mV_{s}$ is the indicator matrix representing for valid mapping between $\gG_{s}$ and $\gG_{br}$, $\mA_{s}^v \in \R^{|\gV_s||\gV_{br}|}$ and $\tA_{s}^e \in \R^{|\gE_s||\gE_{br}|}$ be vertex affinity matrix and edge affinity tensor between $\gG_{s}$ and $\gG_{br}$. For e.g., $\left(\mA_{s}^v\right)_{ij} = d\left(\hat{\mZ}_{s}^{(i)}, \dfrac{1}{3}\sum_{s}\hat{\mZ}_{s}^{(j)}\right)$ with $d(.)$ be a distance metric (e.g., cosine distance) measuring similarity between node $i^{th}$ in $\gG_{s}$ and node $j^{th}$ in $\gG_{br}$. 

To address the NP-Hard nature of aligning each graph to the barycenter graph $\gG_{c}$, which arises from its combinatorial complexity, we employ efficient heuristic solvers utilizing Lagrange decomposition techniques \citep{swoboda2017study,rolinek2020deep}.
\vspace{-0.1in}
\subsection{Backpropagation with Black-box Gradient Estimation}
\vspace{-0.1in}
Given $\hat{\mV}_{s} = \text{SGA}(\mA_{s}^v,\tA_{s}^e)$ be solution obtained from the solver, we aim to learn feature representation for LLMs such that $\hat{\mV}_{s}$ be identical to true triplet alignments explicitly indicated by the barycenter graph. By denoting $\mV_{s}^{*}$ be an optimal mapping between the graph $\gG_{c}$ to $\gG_{br}$, we compute the following total of hamming loss function:  
\setlength{\abovedisplayskip}{2pt}
\setlength{\belowdisplayskip}{2pt}
\begin{equation}\label{eq:training_loss}
{\footnotesize
    \mathcal{L}(\hat{\mV}_{s}, \mV_{s}^{*}) = \sum_{s \in \{v,a,ae\}} \langle \hat{\mV}_{s}, (1-\mV_{s}^{*})\rangle + \langle \mV_{s}^{*} , (1 - \hat{\mV}_{s})\rangle.}
\end{equation}
However, computing the gradient of the loss function with respect to the matching problem inputs \(\left(\mA_{s}^{v}, \tA_{s}^{e}\right)\), i.e., \(\nabla_{\mA_{s}^{v}, \tA_{s}^{e}} \mathcal{L}\left(\hat{\mV}_{s}, \mV_{s}^{*}\right)\), poses a challenge due to the piecewise constant nature of the graph matching objective in Eq.~(\ref{eq:multi-graph}) \citep{poganvcic2020differentiation, rolinek2020optimizing}. To address this, we resort to the IMLE \citep{niepert2021implicit,minervini2023adaptive}, a method that estimates gradients and enables backpropagation through the alignment algorithm by taking the difference between solutions of noise-perturbed alignments.

In particular, given $(\epsilon, \epsilon') \sim\mathrm{Gumble}(0,1)$ and for each $s \in \{v, a, ae\}$, we compute:
\setlength{\abovedisplayskip}{4pt}
\setlength{\belowdisplayskip}{4pt}
\allowdisplaybreaks
\begin{align}
    \left(\dfrac{\partial\mathcal{L}}{\partial \mA_{s}^{v}}, \dfrac{\partial\mathcal{L}}{\partial \tA_{s}^{e}}\right) & \approx \Tilde{\mV}_{s} - \text{SGA}\left(\mA_{s,\lambda}^v,\tA_{s,\lambda}^e\right) \nonumber 
     \text{ where }   \Tilde{\mV}_{s}   = \text{SGA}\left(\mA_{s}^v + \epsilon,\tA_{s}^e + \epsilon'\right), \\
    \left(\mA_{s,\lambda}^v,\tA_{s,\lambda}^e\right) & = \left(\mA_{s}^v + \epsilon,\tA_{s}^e + \epsilon'\right) - \lambda \nabla_{\Tilde{\mV}_{s}} \mathcal{L}(\Tilde{\mV}_{s}, \mV_{s}^{*}), \nonumber 
     \text{ with } \lambda \text{ is a step size.}
\end{align}

\subsection{Graph Distance Properties via Structure Alignment}
% We present theoretical insights into the graph-matching problem outlined in Eq.~(\ref{eq_GM}). Specifically, we demonstrate that once the optimal matching between two graphs is established, it defines a valid metric distance. Additionally, the geodesic path (i.e., the shortest path distance) connecting the two graphs in the manifold space can be derived based on the computed matching alignments. We define a discrete between two graphs given a solution of matching alignment as:

% We provide theoretical insights into the graph-matching problem outlined in Eq.~(\ref{eq_GM}). Specifically, we show that once the optimal matching between two graphs is determined, it defines a valid metric distance (Theorem \ref{theorem_SGM_distance}), i.e., we can compare the distance between two graphs $\gG_1$ and $\gG_2$ based on their joint node-edge representations. Furthermore, the geodesic path (Theorem \ref{theorem_constant_geodesic}), or shortest path distance, connecting the two graphs in the manifold space can be derived from the matching alignments. This property is useful in optimal transport to define dynamic formulations and in representation learning by sampling new data along the geodesic between two endpoint graphs.
\vspace{-0.1in}
We provide theoretical insights into the graph-matching problem in Eq.(\ref{eq_GM}). Once the optimal matching is found, it defines a valid distance metric between graphs (Theorem\ref{theorem_SGM_distance}) based on joint node-edge representations. Additionally, the geodesic path connecting two graphs (Theorem~\ref{theorem_constant_geodesic}) can be derived from these alignments, enabling dynamic formulations in optimal transport and sampling strategies in representation learning. In particular, we define a discrete distance between two graphs given a solution to the matching alignment as follows:
\setlength{\abovedisplayskip}{3pt}
\setlength{\belowdisplayskip}{3pt}
% \begin{align}\label{eq_SGM}
%     d_{\text{SGA}}(\gG_1,\gG_2) &= \min_{\mV \in \gA(\gG_1,\gG_2)} \Big(\sum_{i,j}\emA^v_{i,j}\emV_{i,j}+ \nonumber \\
%     & \sum_{i,j,k,l}\etA^e_{i,j,k,l}\emV_{i,j} \emV_{k,l}\Big).
%     % = \min_{\mV \in \gA(\gG_1,\gG_2)}  O(\mA^v,\tA^e,\mV).
% \end{align}
\begin{equation}
{
\begin{aligned}\label{eq_SGM}
    d_{\text{SGA}}(\gG_1,\gG_2) &= \min_{\mV \in \gA(\gG_1,\gG_2)} \Big(\sum_{i,j}\emA^v_{i,j}\emV_{i,j}+ 
    \sum_{i,j,k,l}\etA^e_{i,j,k,l}\emV_{i,j} \emV_{k,l}\Big).
    % = \min_{\mV \in \gA(\gG_1,\gG_2)}  O(\mA^v,\tA^e,\mV).
\end{aligned}}
\end{equation}
We provide an informal definition of the space of structured graphs below to introduce the main theorems. A more formal treatment, including proofs and definitions, is presented in the Appendix.
\vspace{0.05in}
\begin{definition}[\textbf{Space of all structured graphs}]
$\sS(\gF)$ is the space of all structured graphs defined over a node metric feature space $\left(\gF, d_f\right)$, where each graph is associated with an edge structure space $\left(\cS, d_s \right)$ and a mixing measure $\mu = \sum_{i=1}^N w_i \delta_{(f_i, s_i)}$  over the product space $\left(\gF \times \gS \right)$.

\end{definition}
\vspace{-0.1in}
\begin{boxedtheorem}[\textbf{Metric properties}]\label{theorem_SGM_distance} 
% \begin{theorem}[\textbf{Metric properties}]\label{theorem_SGM_distance}      
    % If the vertex affinity matrix  $\emA^v_{i,j}=(d_f(f_i,f_j))_{i,j}$ and edge affinity tensor $\etA^e_{i,j,k,l} = |d_{s}(s_i,s_k)-d_{s}(s_j,s_l)|$ are distances,
    The distance $d_{\text{SGA}}$ in Eq.~(\ref{eq_SGM}) defines a metric in $\sS(\cF)$.
    %%
    \iffalse
    More precisely, for any two graphs $\gG_1$ and $\gG_2$ in the structured graph space $\sS(\gF)$,
    % described respectively by their mixing measure $\mu_1 = \sum_{i=1}^M w_{1i} \delta_{(f_{1i},s_{1i})}$ and $\mu_2 = \sum_{j=1}^N w_{2j} \delta_{(f_{2j},s_{2j})}$, 
    it holds 
    \begin{enumerate}
        \item Positivity: $d_{\text{SGA}}(\gG_1,\gG_2) > 0$ for any $\gG_1 \neq \gG_2$.

        \item Equality relation: $d_{\text{SGA}}(\gG_1,\gG_2) = 0$ if and only if $\gG_1 = \gG_2$.

        \item Symmetry: $d_{\text{SGA}}(\gG_1,\gG_2) = d_{\text{SGA}}(\gG_2,\gG_1)$.

        \item Triangle inequality: $d_{\text{SGA}}(\gG_1,\gG_3) \le d_{\text{SGA}}(\gG_1,\gG_2) + d_{\text{SGA}}(\gG_2,\gG_3)$ for any graph $\gG_3$.
    \end{enumerate}
    \fi
   
\end{boxedtheorem}
% More precisely, SGM satisfies the triangle inequality and is null if and only if $N=M$ and there exists a bijection \ldots
% \end{theorem}
%%
% Theorem \ref{theorem_SGM_distance} is proved in Appendix \ref{proof_theorem_SGM_distance}. Intuitively, we can compare two graphs, $\gG_1$ and $\gG_2$, described by their mixing measures $\mu_1 = \sum_{i=1}^M w_{1i} \delta_{(f_{1i},s_{1i})}$ and $\mu_2 = \sum_{j=1}^N w_{2j} \delta_{(f_{2j},s_{2j})}$, with uniform vertex weights (\ie~$w_{1i} = 1, \forall i \in [M]$, $w_{2j} = 1, \forall j \in [N]$) and shortest path structure matrices. 
\vspace{-0.05in}
The $d_{\text{SGA}}$ distance above is zero if and only if there exists a one-to-one mapping between the graph vertices that preserves both shortest paths and features and both graphs have the same number of vertices.
\vspace{-0.05in}
\begin{boxedtheorem}[\textbf{Geodesic space}]\label{theorem_constant_geodesic}
        The space $\sS(\cF)$ equipped with the $d_{\text{SGA}}$ distance is geodesic.
\end{boxedtheorem}
\section{Experiments}
\vspace{-0.1in}
\subsection{Implementation Details}
\begin{table}[H]
\begin{minipage}{0.68\textwidth}
\vspace{-0.2in}
\begin{minipage}{\textwidth}
    \centering
    \caption{\small Fine-tuning performance on MedVQA  tasks (\textbf{pre-trained 10\%}). \textbf{Bold} indicates the best values among \colorbox{cyan!12}{pre-training algorithms (Sec. \ref{sec:multi-modal-pretraining})}, excluding \texttt{LLaVA-Med}.}
    
    \begin{adjustbox}{width=0.98\textwidth}
    \setlength{\tabcolsep}{2.0 pt}
    \vspace{0.1in}
    \begin{tabular}{l ccc ccc ccc l}
        \toprule
        \multirow{2}{*}{\textbf{Method}} & \multicolumn{3}{c}{\textbf{VQA-RAD}} & \multicolumn{3}{c}{\textbf{SLAKE}} & \multicolumn{3}{c}{\textbf{PathVQA}} & \multirow{2}{*}{\textbf{Overall}} \\
        \cmidrule(lr){2-4} \cmidrule(lr){5-7} \cmidrule(lr){8-10}
        & Open & Closed & Avg. & Open & Closed & Avg. & Open & Closed & Avg. & \\
        \midrule
        LLaVA-Med (100\%) & 63.65&81.62&72.64&83.44&83.41&83.43&36.78&91.33&64.06& 73.37 \\
        LLaVA-Med (10\%) & 43.38\textcolor{red}{$\downarrow$\scriptsize{20.3}}&61.4\textcolor{red}{$\downarrow$\scriptsize{20.2}}&52.39\textcolor{red}{$\downarrow$\scriptsize{20.3}}&80.94\textcolor{red}{$\downarrow$\scriptsize{2.5}}&80.29\textcolor{red}{$\downarrow$\scriptsize{3.1}}&80.62\textcolor{red}{$\downarrow$\scriptsize{2.8}}&24.26\textcolor{red}{$\downarrow$\scriptsize{13.69}}&88.03\textcolor{red}{$\downarrow$\scriptsize{3.18}}&56.15\textcolor{red}{$\downarrow$\scriptsize{7.91}}&63.05\textcolor{red}{$\downarrow$\scriptsize{10.3}}\\
        InfoNCE&59.39&77.57&68.48&82.4&83.17&82.78&34.59&\textbf{91.45}&63.02&71.43 \\
        PLOT&16.86&26.47&21.67&37.81&56.25&47.03&11.79&81.36&46.58&38.42 \\
        SigLIP&56.99&77.94&67.47&80.86&80.53&80.69&18.08 &50.85 &34.465 &60.88\\
        VLAP&57.49&76.47&66.98&80.05&82.21&81.13&32.21 &91.16 &61.685 &69.93\\
        \midrule       GeoCLAP&60.68&75.37&68.03&82.64&\textbf{85.10}&83.87&35.12 &91.15 &63.14 &71.68\\
        PAC-S&57.72&72.79&65.26&83.78&81.49&82.64&35.01 &91.36 &63.19 &70.36\\   IMAGEBIND&57.31&75.74&66.53&80.79&84.13&82.46&34.61 &91.42 &63.02 &70.67\\
        \rowcolor{Gray}
        \textbf{ExGra-Med}&\textbf{66.02}&\textbf{79.04}&\textbf{72.52}&\textbf{84.92}&\textbf{85.10}&\textbf{85.01}&\textbf{37.25}&\textbf{91.45}&\textbf{64.34}&\textbf{73.96} \\
        \bottomrule
    \end{tabular}
    \end{adjustbox}
    \label{tab:model_comparison_10percent}
\end{minipage}
\begin{minipage}{\textwidth}
    \centering
    \caption{\small Fine-tuning performance on MedVQA  tasks (\textbf{pre-trained 40\%}). \textbf{Bold} indicate for best values among \colorbox{cyan!12}{pre-training algorithms (Sec. \ref{sec:multi-modal-pretraining})} excluding \texttt{LLaVA-Med}.}
    % \vspace{0.1in}
    \begin{adjustbox}{width=0.98\textwidth}
    \setlength{\tabcolsep}{0.5 pt}
    \vspace{-0.2in}
    \begin{tabular}{l ccc ccc ccc l}
        \toprule
        \multirow{2}{*}{\textbf{Method}} & \multicolumn{3}{c}{\textbf{VQA-RAD}} & \multicolumn{3}{c}{\textbf{SLAKE}} & \multicolumn{3}{c}{\textbf{PathVQA}} & \multirow{2}{*}{\textbf{Overall}} \\
        \cmidrule(lr){2-4} \cmidrule(lr){5-7} \cmidrule(lr){8-10}
        & Open & Closed & Avg. & Open & Closed & Avg. & Open & Closed & Avg. & \\
        \midrule
        LLaVA-Med (100\%) & 63.65&81.62&72.64&83.44&83.41&83.43&36.78&91.33&64.06&73.37 \\
        LLaVA-Med (40\%) & 62.23\textcolor{red}{$\downarrow$\scriptsize{1.4}}&79.41\textcolor{red}{$\downarrow$\scriptsize{2.2}}&70.82\textcolor{red}{$\downarrow$\scriptsize{1.8}}&84.42\textcolor{red}{$\uparrow$\scriptsize{1.0}}&83.65\textcolor{red}{$\downarrow$\scriptsize{0.2}}&84.04\textcolor{red}{$\uparrow$\scriptsize{0.6}}&31.86\textcolor{red}{$\downarrow$\scriptsize{4.9}}&84.99\textcolor{red}{$\downarrow$\scriptsize{6.3}}&58.43\textcolor{red}{$\downarrow$\scriptsize{5.6}}&71.09\textcolor{red}{$\downarrow$\scriptsize{2.3}}\\
        InfoNCE&63.11&77.57&70.34&82.68&83.89&83.29&33.58 &89.62 &61.6 &71.74 \\
        PLOT&64.36&79.41&71.89&83.38&82.93&83.16&35.11 &89.59 &62.35 &72.46 \\
        SigLIP&63.02&81.25&72.14&81.26&80.29&80.77&36.01 &90.86 &63.435 &72.12\\
        VLAP&63.17&79.04&71.11&83.38&83.89&83.64&35.62 &90.83 &63.225 &72.66\\
        \midrule               GeoCLAP&62.28&\textbf{82.72}&72.5&82.64&85.2&83.92&33.2 &75.05 &54.13 &70.18\\
        PAC-S&63.77&79.41&71.59&\textbf{84.52}&85.58&85.05&27.11 &85.34&56.23 &70.96\\   IMAGEBIND&64.73&78.68&71.71&82.31&84.62&83.47&35.76 &87.08 &61.42 &72.20\\
        \rowcolor{Gray}
        \textbf{ExGra-Med}&\textbf{66.01}&\textbf{82.72}&\textbf{74.37}&{84.47}&\textbf{85.82}&\textbf{85.15}&\textbf{37.41} &\textbf{91.27} &\textbf{64.34} &\textbf{74.57} \\
        \bottomrule
    \end{tabular}
    \end{adjustbox}
    \vspace{-0.1in}
    \label{tab:model_comparison_40percent}
\end{minipage}
\end{minipage}
\begin{minipage}{0.30\textwidth}
\vspace{-0.1in}
    \begin{minipage}{\textwidth}
    \caption{{\small ExGra-Med versus \colorbox{cyan!12}{LLaVa-Med when pre-trained} \colorbox{cyan!12}{with extended captions (ext.cap)}. Performance is reported on VQA-RAD downstream after fine-tuning.}}
            \resizebox{\columnwidth}{!}{
    % \begin{tabular}{@{\hspace{-0pt}}l@{\hspace{8pt}}c@{\hspace{8pt}}c@{\hspace{10pt}}c}%rrrrrrrrrrrr
    % \toprule
    % Method         & VQA-RAD  & SLAKE \\
    % \midrule
    % ExGra-Med (Full)                    & \textbf{74.37}     & \textbf{84.99} \\
    % \rowcolor{cyan!50}
    % LVM-Med with synonyms               &  72.39   & 82.93  \\
    % ExGra-Med in two stages                   &  72.81  & 84.14  \\
    % \rowcolor{cyan!50}
    % LVM-Med w/o long-context             &  72.12   & 81.95 \\
    % LVM-Med w/o message passing             &  73.90   & 84.29 \\
    % LVM-Med w/o message passing             &  73.90   & 84.29 \\
    % LVM-Med w/o message passing             &  73.90   & 84.29 \\
    % LVM-Med w/o message passing             &  73.90   & 84.29 \\
    % LVM-Med w/o message passing             &  73.90   & 84.29 \\
    % LVM-Med w/o message passing             &  73.90   & 84.29 \\
    % \bottomrule
    % \end{tabular}
    % }
    \setlength{\tabcolsep}{1.0 pt}
    
    \begin{tabular}{l ccc}
        \toprule
        \multirow{2}{*}{\textbf{Method}} & \multicolumn{3}{c}{\textbf{VQA-RAD}}  \\
        \cmidrule(lr){2-4} 
        & Open & Closed & Avg.  \\
        \midrule
        LLaVA-Med (100\%) & 63.65&81.62&72.64 \\
        LLaVA-Med (10\%) & 43.38\textcolor{red}{$\downarrow$\scriptsize{20.3}}&61.4\textcolor{red}{$\downarrow$\scriptsize{20.2}}&52.39\textcolor{red}{$\downarrow$\scriptsize{20.3}}\\
        \rowcolor{Gray} LLaVA-Med (10\%) ext.cap & 63.07&75.74&69.41\\
        % \midrule
        \rowcolor{Gray}
        \textbf{ExGra-Med} (10\%)&\textbf{66.02}&\textbf{79.04}&\textbf{72.52}\\
        \midrule
        LLaVA-Med (40\%) & 62.23\textcolor{red}{$\downarrow$\scriptsize{1.4}}&79.41\textcolor{red}{$\downarrow$\scriptsize{2.2}}&70.82\textcolor{red}{$\downarrow$\scriptsize{1.8}}\\
\rowcolor{Gray} LLaVA-Med (40\%) ext.cap &63.68 &79.78&71.73\\
        \rowcolor{Gray}
        \textbf{ExGra-Med} (40\%)&\textbf{66.01}&\textbf{82.72}&\textbf{74.37}\\
        \bottomrule
    \end{tabular}
    }
    \vspace{-0.05in}
    \label{tab:llava-med-ext-cap}  
    \end{minipage}
    
    \begin{minipage}{\textwidth}
    \caption{\small \textsc{ExGra-Med} results with \colorbox{cyan!12}{different frozen LLMs}. It shows that Gemini and Qwen is also effective within our method.}
    \vspace{0.1in}
    \resizebox{1.0\columnwidth}{!}{
    \setlength{\tabcolsep}{0.1 pt}
    \vspace{0.1in}
        \begin{tabular}{@{\hspace{-0pt}}l@{\hspace{8pt}}c@{\hspace{8pt}}c@{\hspace{10pt}}c}%rrrrrrrrrrrr
\toprule
\textbf{Method}         & \textbf{VQA-RAD}  & \textbf{SLAKE} \\
\midrule
ExGra-Med (GPT-4), $10 \%$                    & \textbf{72.52}     & \textbf{85.01} \\
\rowcolor{Gray} ExGra-Med (Gemini), $10 \%$                    & \underline{71.09}     & 83.98 \\
\rowcolor{Gray} ExGra-Med (Qwen), $10 \%$   & 70.13 & \underline{84.7} \\
LLaVa-Med (Baseline) $10 \%$                    & {52.39}     & {80.62} \\
\midrule
ExGra-Med (GPT-4), $40 \%$                    & \textbf{74.37}     & \textbf{85.15} \\
\rowcolor{Gray} ExGra-Med (Gemini), $40 \%$                    & \underline{73.26}     & \underline{85.10} \\
\rowcolor{cyan!50} ExGra-Med with synonyms,  $40 \%$               &  72.39   & 82.93  \\
LLaVa-Med (Baseline) $40 \%$                    & {70.82}     & {84.04} \\
\bottomrule
\end{tabular}}
\vspace{-0.05in}
\label{tab:exgra-med-other-llms}
    \end{minipage}
\end{minipage}
\vspace{-0.1in}
\end{table}
\vspace{0.05in}
\textbf{Model Architecture Configuration.}
% We use \texttt{LLaMA} \citep{touvron2023llama} large language model (LLM), \texttt{CLIP-ViT-L-Patch14} \citep{radford2021learning} visual encoder, and an MLP projection the same as \texttt{LLaVA 1.5}'s architecture \citep{llava_15}. We keep both visual encoder and LLM weights frozen in stage 1 and keep only visual encoder weights frozen in stage 2. We train the model in stage 1 following the standard stage 1 of \texttt{LLaVA-Med} \citep{llava-med} and combine multi-graph alignment with autoregressive in stage 2. To apply multi-graph alignment in stage 2, we also apply a graph convolutional network model with 2 layers to the output of Projection and LLM Decoder (image and text modalities) before multi-graph alignment. The setting of autoregressive pre-training is the same as \texttt{LLaVA-Med} \citep{llava-med}. We trained the model with 1 epoch in stage 1 and 3 epochs in stage 2 on the dataset used in \texttt{LLava-Med} \citep{llava-med}. The model is optimized with \texttt{Adam} \cite{kingma2014adam} with an initial learning rate $2e-3$ and $2e-5$ for stage 1 and stage 2, respectively. We also use \texttt{CosineAnnealingLR} for learning rate scheduler. These hyperparameter settings are also used for original \texttt{LLaVA-Med 1.5} 
% \citep{llava-med}.
We use the \texttt{LLaMA-7B} large language model \citep{touvron2023llama}, the \texttt{CLIP-ViT-L-Patch14} visual encoder \citep{radford2021learning}, and an MLP projection similar to \texttt{LLaVA 1.5} \citep{llava_15}.
% In Stage 1, we freeze both the visual encoder and LLM weights; in Stage 2, only the visual encoder weights remain frozen. 
\textit{Stage 1} follows the standard \texttt{LLaVA-Med} \citep{llava-med} setup, while \textit{stage 2 incorporates our multi-graph alignment with autoregressive training}. For multi-graph alignment, a 2-layer graph convolutional network is applied to the output of the Projection and LLM Decoder (handling both image and text modalities). We train for 1 epoch in stage 1 and 3 epochs in stage 2 using the same dataset as \texttt{LLava-Med}. The model is optimized using \texttt{Adam} \citep{kingma2014adam} with CosineAnnealingLR scheduler and learning rates of $2e-3$ and $2e-5$ for stages 1 and 2, respectively.
% and a \texttt{CosineAnnealingLR} scheduler, following the same hyperparameters as \texttt{LLaVA-Med 1.5}

% \vspace{-0.05in}
\textbf{Pre-training data.}
% We follow the same dataset being used in \texttt{LLaVA-Med} \citep{llava-med}. For stage 1, there are 600K image-text pairs filtered from \texttt{PMC-15M}. They are all converted to instruction-following data with simple instructions for describing the image. For stage 2, there are 60K image-text pairs extracted from \texttt{PMC-15M} with 5 modalities: CXR (chest X-ray), CT (computed tomography), MRI (magnetic resonance imaging), histopathology, and gross (i.e., macroscopic) pathology. Then, the language-only \texttt{GPT-4} is used to generate multi-round questions and answers in a tone as if it could see the image to convert these pairs to an instruction-following format. 
We use the same dataset as \texttt{LLaVA-Med} \citep{llava-med}. Stage 1 includes 600K image-text pairs filtered from \texttt{PMC-15M}, converted into instruction-following data with simple image descriptions. Stage 2 comprises 60K image-text pairs from \texttt{PMC-15M} across five modalities: CXR, CT, MRI, histopathology, and gross pathology. \texttt{GPT-4} is then employed to generate multi-round Q\&A in a tone mimicking visual interpretation, converting these pairs into an instruction-following format.

%In addition, before doing zero-shot classification task, we also continue to pre-train the model on \texttt{PMC-VQA} dataset, which having 227k image-question pairs with question and modality diversity.
% \vspace{-0.05in}
\textbf{Running-time.}
We train \textsc{ExGra-Med} using 4 A100-GPUs per with 80GB for both stages and complete the training process for stage 1 in 6.5 hours and for stage 2 in 7.5 hours. With original \texttt{LLaVA-Med} (version 1.5) \citep{llava-med}, the training process for stage 1 finishes in 6.5 hours, and for stage 2 finishes in 7 hours. In total, we need an extra 0.5 hour to complete the whole pre-training process compared to the LLaVa-Med.
\vspace{-0.1in}

\subsection{Autoregressive Training is Data-hungry}
\vspace{-0.05in}
We highlight the data-intensive nature of autoregressive training by evaluating \texttt{LLaVA-Med}, a state-of-the-art biomedical multimodal language model. \texttt{LLaVA-Med} follows a two-stage training process: Stage 1 aligns image-text tokens with biomedical concepts, and Stage 2 fine-tunes the model for instruction tasks. We pre-trained \texttt{LLaVA-Med} on varying data amounts (10\%, 40\%, 70\%) and fine-tuned it on the VQA-RAD dataset. As shown in Figure~\ref{fig:auto-regressive-less}, performance drops significantly at 10\% pre-training compared to full training, revealing the autoregressive mechanism’s data dependency issue in medical-MLLMs. This highlights the challenge of weak connections between visual features and text embeddings without sufficient instruction-tuning data.

In contrast, our \textsc{ExGra-Med} specifically learns image-text alignment by enforcing semantic consistency across images, instruction responses, and extended contexts. Using the same setup as \texttt{LLaVA-Med}, 
% it is pretrained on varying data sizes and fine-tuned with pre-trained checkpoints. 
\textsc{ExGra-Med} excels even with limited data, achieving $72.52\%$ at just $10\%$ pretraining, far above \texttt{LLaVA-Med}’s $52.39\%$ and consistently outperforms \texttt{LLaVA-Med} across instruction tuning rates ($40\% - 100\%$).
\newline
\\
\textbf{Impact of Longer Training and Enriched Captions on \textsc{LLaVa-Med} Performance}. 
We conduct a deeper analysis of the data-hungry phenomenon by examining \textsc{LLaVa-Med} when (i) \textit{trained longer in Stage 2} (with an additional hour) and when (ii) \textit{incorporating extended captions} in an autoregressive manner, as done in \textsc{ExGra-Med}. The experiments were performed using 10\% and 40\% pre-training settings, followed by fine-tuning on the VQA-RAD dataset. From the results in Table \ref{tab:llava-med-ext-cap}, we draw two key conclusions. First, extended captions improve \textsc{LLaVa-Med} performance, particularly in the data-scarce 10\% pre-training scenario. However, in both settings, \textsc{ExGra-Med} demonstrates superior performance with significant margins, \textit{highlighting the effectiveness of its multi-graph alignment strategy} in mitigating the data-hungry issues of the autoregressive mechanism.

% This suggests that the multi-graph alignment strategy effectively mitigates the data-intensive demands of autoregressive mechanisms in mLLMs.
\vspace{-0.05in}
\subsection{Multi-modal Pre-training Comparison}
\label{sec:multi-modal-pretraining}
\vspace{-0.05in}
To further demonstrate the strengths of \textsc{ExGra-Med}, we compare its performance against other vision-language pre-training methods currently used for visual instruction tuning to enhance frozen vision-language models

\textbf{Datasets.} We test pre-trained models on three prominent biomedical VQA datasets: \texttt{VQA-RAD} \citep{lau2018dataset}, \texttt{SLAKE} \citep{liu2021slake}, and \texttt{PathVQA} \citep{he2020pathvqa}. \texttt{VQA-RAD} includes 3,515 questions across 315 radiology images, while \texttt{SLAKE} contains 642 radiology images from various body parts and over 7k QA pairs. \texttt{PathVQA}, focused on pathology, features 5k images and 32.8k questions. All datasets include \textbf{open-ended} (e.g., what, why, where) and \textbf{closed-ended }(yes/no or two-option) question types. We report performance using recall for open-ended, which evaluates the proportion of ground-truth tokens that appear in the generated sequences and accuracy for the closed-ended questions.

% We provide more details in the Appendix. 
% We test pre-trained models on 
% three well-known biomedical visual question answering (VQA) datasets: VQA-RAD\citep{lau2018dataset}, SLAKE \citep{liu2021slake} and PathVQA \citep{he2020pathvqa}. VQA-RAD \citep{lau2018dataset} focuses on radiology images and includes 3,515 total questions with 315 image semantically-Labeled Knowledge-Enhanced dataset or SLAKE \citep{liu2021slake} contains 642 radiology images across many human body parts such as brain, neck, chest, abdomen, and pelvic cavity, and have over 7k QA pairs. PathVQA \citep{he2020pathvqa}, which is a VQA dataset concentrating on the pathology domain, comprises about 5k images and 32.8k questions. All three datasets contain two different types: open-ended for example what, why, where, how, \textit{etc.} and closed-ended (yes/no or two-option question).  
% \vspace{-0.05in}
\textbf{Baselines.} 
We compare \textit{seven approaches}, including: 
\begin{itemize}
\vspace{-0.1in}
\item \textbf{two modalities-based methods} such as \texttt{InfoNCE}-based methods \citep{khan2023contrastive,liu2023contrastive}, \texttt{SigLIP} \citep{zhai2023sigmoid}, \texttt{PLOT} \citep{chen2022plot}, \texttt{VLAP} \citep{park2024bridging}. Among this, \texttt{SigLIP} adapts the Sigmoid loss on image-text pairs to break the global view of the pairwise similarities for normalization, resulting in scaling in large batch sizes. \texttt{PLOT} defines optimal transport as a distance between visual image patches and text embedding. In contrast, \texttt{VLAP} uses assignment prediction to bridge the modality gap between the visual and LLM embeddings.
\item \textbf{multi-modal learning across three modalities} such as image, text, voice, or augmented image—is explored by methods like \texttt{PAC-S} \cite{sarto2023positive}, \texttt{GeoCLAP} \cite{khanal2023learning}, and \texttt{IMAGEBIND} \cite{girdhar2023imagebind}.
\texttt{PAC-S} integrates contrastive losses across multiple modality pairs: (image–text), (image–augmented text), and (text–augmented text).
\texttt{GeoCLAP} applies CLIP-style contrastive learning to all cross-domain pairs, such as audio–text and text–image.
Similarly, \texttt{IMAGEBIND} generalizes the InfoNCE loss to learn unified embeddings across diverse modalities.

\end{itemize}
% (i) \textbf{two modalities-based methods} such as \texttt{InfoNCE}-based methods \citep{khan2023contrastive,liu2023contrastive}, \texttt{SigLIP} \citep{zhai2023sigmoid}, \texttt{PLOT} \citep{chen2022plot}, \texttt{VLAP} \citep{park2024bridging}, and (ii) \textbf{three modalities (image, text, voice, or augmented image)} such as \texttt{PAC-S} \cite{sarto2023positive}, \texttt{GeoCLAP} \cite{khanal2023learning}, and \texttt{IMAGEBIND} \cite{girdhar2023imagebind}. 
We train the baselines under the same settings as \textsc{ExGra-Med} with varying pre-training data rates and compare their performance on downstream tasks.

\textbf{Results.} Tables~\ref{tab:model_comparison_10percent}–\ref{tab:model_comparison_100percent} show our performance and baselines under $10\%$, $40\%$, and $100\%$ instruction-tuning data. While most contrastive baselines improve over \texttt{LLaVA-Med} at $10\%$, \textsc{ExGra-Med} consistently outperforms all methods across settings. It excels particularly on open-ended questions requiring external knowledge and maintains stable gains across all VQA datasets. In contrast, some methods like \texttt{SigLIP} peak at $40\%$ (e.g., $72.14\%$ Avg on VQA-RAD) but drop notably at $100\%$ (down over $6\%$), whereas \textsc{ExGra-Med} improves further to $74.91\%$ (Avg) and $74.75\%$ (Overall).
% In Tables \ref{tab:model_comparison_10percent}, \ref{tab:model_comparison_40percent} and \ref{tab:model_comparison_100percent}, 
% we show the performance of \textsc{ExGra-Med} and the baselines when pre-trained with $10\%$, $40\%$, and $100\%$ of instruction-tuning data. While most contrastive baselines improve \texttt{LLaVA-Med} at 10\%, \textsc{ExGra-Med} consistently outperforms \texttt{LLaVA-Med} and other methods overall. In open-ended questions, which require external knowledge, \textsc{ExGra-Med} delivers the best results. Notably, it maintains stable improvements across all three VQA datasets, unlike other methods that peak at $40\%$ pre-training and decline afterward. For example, \texttt{SigLIP} scores $72.14\%$ (Average) and $72.12\%$ (Overall) on VQA-RAD at $40\%$ but drops over $6\%$ and $1\%$ at $100\%$. In contrast, \textsc{ExGra-Med} continues to improve, reaching $74.91\%$ (Average) and $74.75\%$ (Overall).
\vspace{-0.1in}
\begin{table*}
\begin{minipage}{0.62\textwidth}
    \vspace{-0.1in}
    \centering
    \caption{Comparison with \colorbox{cyan!12}{other Med-MLLMs on MedVQA tasks}. \textit{All models (except GPT-4) are fine-tuned on the same training set in each VQA task}. These Med-MLLMs differ notably in model size, training data volume, and pre-training strategies - e.g., ExGra-Med (7B, 60K GPT-4 augmented samples) vs. MedDR (40B, 2M samples).}
    \begin{adjustbox}{width=\textwidth}
    \setlength{\tabcolsep}{2.0 pt}
    \vspace{0.1in}
    \begin{tabular}{l l ccc ccc ccc l}
        \toprule
        \multirow{2}{*}{\textbf{Method}} & \#\textbf{Para} & \multicolumn{3}{c}{\textbf{VQA-RAD}} & \multicolumn{3}{c}{\textbf{SLAKE}} & \multicolumn{3}{c}{\textbf{PathVQA}} & \multirow{2}{*}{\textbf{Overall}} \\
        \cmidrule(lr){3-5} \cmidrule(lr){6-8} \cmidrule(lr){9-11}
        & & Open & Closed & Avg. & Open & Closed & Avg. & Open & Closed & Avg. & \\
        \midrule
        LLaVA-Med & 7B & 63.65&81.62&72.64&83.44&83.41&83.43&{36.78}&{91.33}&\underline{64.06}& {73.37} \\
        BiomedGPT-B & 182M & 60.9&81.3&71.1&84.3&\textbf{89.9}&\textbf{87.1}&28&88&58&72.07\\
        M2I2 & - &61.8&81.6&71.7&74.7&91.1&82.9&36.3&88&62.15&72.25 \\
        BioMed-CLIP & 422M & \textbf{67.6}&79.8&73.7&82.5&89.7&\underline{86.1}& & & & \\
        Med-Dr  & 40B & 37.5&78.9&58.2&74.2&83.4&78.8&33.5&90.2&61.85&66.28 \\
        LLaVA (general) & 7B & 50&65.1&57.55&78.2&63.2&70.7&7.7&63.2&35.45&54.57 \\
        GPT-4 & 200B & 39.5&78.9&59.2&33.6&43.6&38.6 & & & & \\
        Med-MoE (Phi2) & 3.6B &58.55&\underline{82.72}&70.64&\underline{85.06}&{85.58}&85.32&34.74&\textbf{91.98}&63.36&73.11 \\
        Med-MoE (Stable LM) & 2B &50.08&80.07&65.08&83.16&83.41&83.29&33.79&91.30&62.55&70.3 \\
        % \midrule
        \rowcolor{Gray}
        \textbf{ExGra-Med} & 7B &{66.35}&\textbf{83.46}&\underline{74.91}&\textbf{85.34}&{85.58}&85.46&\underline{36.82}&90.92&{63.87}&\underline{74.75}\\
        \rowcolor{Gray}
        \textbf{ExGra-Med (DCI)} & 7B &\underline{67.03}&\textbf{83.46}&\textbf{75.25}&{84.88}&{85.58}&85.23&\textbf{37.77}&\underline{91.86}&\textbf{64.82}&\textbf{75.1}\\
        \bottomrule
    \end{tabular}
    \end{adjustbox}
    \label{tab:model_comparison}
\end{minipage}
\hfill
\begin{minipage}{0.36\textwidth}
\vspace{-0.1in}
    \caption{{\colorbox{cyan!12}{ExGra-Med ablation study}. Results are presented as average scores on VQA-RAD and SLAKE, using pre-trained weights on 10\%, 40\%, 100\%.}}
    \setlength{\tabcolsep}{0.5 pt}
    \resizebox{\columnwidth}{!}{
    \begin{tabular}{@{\hspace{-0pt}}l@{\hspace{8pt}}c@{\hspace{8pt}}c@{\hspace{10pt}}c}%rrrrrrrrrrrr
    \toprule
    \textbf{Method}         & \textbf{VQA-RAD}  & \textbf{SLAKE} \\
    \midrule
    ExGra-Med (Full, 10\%, $\alpha = 1.0$)                    & \textbf{72.52}     & \textbf{85.01} \\
    \rowcolor{cyan!50}
    - (ii) ExGra-Med (Full, 10\%, $\alpha = 0.1$)                    & {65.95}     & {82.9} \\
    - (ii) ExGra-Med (Full, 10\%, $\alpha = 0.5$)                    & {67.72}     & {82.33} \\
    \midrule
    ExGra-Med (Full, 40\%)                    & \textbf{74.37}     & \textbf{84.99} \\
    % \rowcolor{cyan!50}
    % LVM-Med with synonyms               &  72.39   & 82.93  \\
    % ExGra-Med in two stages                   &  72.81  & 84.14  \\
    \rowcolor{cyan!50}
    - (iii) ExGra-Med w/o ext. context             &  72.12   & 81.95 \\
    \rowcolor{cyan!50}
    - (iv) ExGra-Med w/o ori. caption             &  72.58   & 82.31 \\
    - (v) ExGra-Med w/o message passing             &  73.90   & 84.29 \\\rowcolor{cyan!50}
    - (vi) ExGra-Med in two stages             &  72.81   & 84.14 \\
    \midrule
    ExGra-Med (Full, 100\%)                    & \textbf{74.91}     & \textbf{85.46} \\
    - (vii) ExGra-Med w/o barycenter graph             &  73.88   & 84.34 \\
    \bottomrule
    \end{tabular}}
    \label{tab:logra_ablation}
\end{minipage}
\vspace{-0.1in}
\end{table*}
\subsection{Med-VQA Comparison with Medical MLLMs}
\vspace{-0.05in}
We now compare \textsc{ExGra-Med} pre-trained with $100\%$ data against other medical foundation models, each trained on varying datasets and employing different architectures or model sizes.
\\
\textbf{Baselines.} 
% We use three aforementioned datasets: VQA-RAD, SLAKE and PathVQA to evaluate our approach against popular methods 
We benchmark \textit{eight competitors}, both generic or medical foundation models, including
\texttt{LLaVA} \citep{llava}, \texttt{LLaVA-Med} \citep{llava-med}, \texttt{Med-Flamingo} \citep{moor2023med}, \texttt{Med-Dr} \citep{he2024meddr}, \texttt{Biomed-GPT} \citep{zhang2023biomedgpt}, \texttt{M2I2} \citep{10230743}, \texttt{GPT-4o} \citep{achiam2023gpt} and \texttt{Med-MoE} \citep{jiang2024moe}. Whilst LLaVA and GPT-4o have no medical background, the others are pre-trained on a variety of biomedical knowledge. With the exception of \texttt{LLaVa}, which we reproduced, the results for the other baselines are taken from the literature. Moreover, we also present an enhanced version, \textsc{ExGra-Med + DCI}, which integrates multi-scale visual features from vision encoders \citep{yao2024dense}, potentially benefiting medical image analysis by considering both local (detailed) and global (contextual) features.
% we also introduce another \textsc{ExGra-Med} version, namely \textsc{ExGra-Med + DCI} by fusing different multi-scale visual features of vision encoders \citep{yao2024dense} which might be useful in medical images.

% - a simple plug-and-play module. We utilize Dense Channel Integration (DCI), that enhances visual-learning ability of MLLMs by densely concatenating multi-layer visual features.
\textbf{Results.}
% Overall, two \textsc{ExGra-Med} versions perform better than the baseline models (Table \ref{tab:model_comparison}), especially the version combined with DCI gains the best performance for PathVQA on Average score (64.82\%) and for Overall result (75.1\%). Compared to \texttt{LLaVA-Med}, \textsc{ExGra-Med} exhibits significant improvements on every task result, such as 2.01\% on VQA-RAD, 2.03\% on SLAKE and 0.76\% on PathVQA. Furthermore, it is important to note that both the plain and DCI versions of \textsc{ExGra-Med} demonstrate competitive performance across all VQA datasets, despite having significantly fewer parameters. For instance, the two 7B-parameter \textsc{ExGra-Med} versions outperform the 40B-parameter \texttt{Med-Dr} across all three datasets.
Overall, both versions of \textsc{ExGra-Med} outperform baseline models (Table~\ref{tab:model_comparison}), with the DCI variant achieving the best results on PathVQA ($64.82\%$ average, $75.1\%$ overall). Compared to \texttt{LLaVA-Med}, it shows notable gains: +$2.01\%$ on VQA-RAD, +$2.03\%$ on SLAKE, and +$0.76\%$ on PathVQA. Despite using only 7B parameters, both \textsc{ExGra-Med} models surpass the 40B \texttt{Med-Dr} across all datasets.

% Overall, both versions of \textsc{ExGra-Med} perform better than baseline models (Table \ref{tab:model_comparison}), with the DCI setting achieving the highest performance on PathVQA (64.82\% average score, 75.1\% overall). Compared to \texttt{LLaVA-Med}, \textsc{ExGra-Med} shows significant improvements across tasks, including 2.01\% on VQA-RAD, 2.03\% on SLAKE, and 0.76\% on PathVQA. Notably, both versions of \textsc{ExGra-Med} perform competitively across all VQA datasets, despite having fewer parameters. For example, the two 7B-parameter versions outperform the 40B-parameter \texttt{Med-Dr} on all three datasets.
% both \textsc{ExGra-Med} plain and DCI-version have competitive performance in every score of each VQA dataset, although they own the much lower parameter amount. For example, compared to \texttt{Med-Dr} having up to \textit{40B parameters}, two 7B-parameter \textsc{ExGra-Med} versions all exceed all the scores on three datasets.
% \begin{figure}[h!]  % Positioning of the figure
%     \centering  % Center the figure
%     \includegraphics[width=0.5\textwidth]{./figures/perf_vqarad.pdf}  % Adjust width as needed
%     \caption{A comparison of our method with competitive methods at various instruction tuning percentages on the VQA-RAD dataset.}  % Caption for the figure
%     \label{fig:perf_vqarad}  % Label for referencing
% \end{figure}
\vspace{-0.05in}
\subsection{Medical Visual Chatbot Evaluation \& Zero-shot Image Classification}
\vspace{-0.05in}
\paragraph{Medical Visual Chatbot.} We present in Section \ref{sec:medical_visual_chatbot} Appendix \textsc{ExGra-Med} results compared against several SOTA general and Med-MLLMs such as LLaVA, GPT-4o, LLaVA-Med, Med-Flamingo, Med-Dr, and Biomed-GPT. Among these, we observe that \textsc{ExGra-Med} is the best model across question types.
\vspace{-0.05in}
\paragraph{Zero-shot Image Classification as MedVQA.}
Results are presented in Section \ref{sec:zero-shot-img-class} of the Appendix. In short, we outperform other models across all datasets, particularly excelling in microscopy, where it surpasses RadFM by $8.2\%$
\vspace{-0.05in}
\subsection{Additional Analysis}
\vspace{-0.05in}
% \textbf{ExGra-Med generalization with other base models}

% \begin{table}[H]
%     \centering
%     \caption{Performance comparison for integrating multi-graph alignment framework (ExGra-Med) into Qwen-VL 7B model}
%     \vspace{0.05in}
% \begin{adjustbox}{width=0.7\textwidth}
% \begin{tabular}{@{\hspace{-0pt}}l@{\hspace{8pt}}c@{\hspace{8pt}}c@{\hspace{8pt}}c}%rrrrrrrrrrrr
% \toprule
% \textbf{Method}         & \textbf{VQA-RAD}  & \textbf{Slake}\\
% \midrule
% LLaVA-Med (medical pre-trained 10\%)             & 61.40     & 80.30\\
% \textbf{ExGra-Med} (LLaVA architecture, pre-trained 10\%) & \textbf{79.40}     & \textbf{85.10}\\
% \midrule
% Qwen-VL (general weights)  & 53.85     & 70.50\\
% Qwen-VL Med (medical pre-trained 10\%)  & 52.73     & 86.53\\
% \textbf{ExGra-Med} (Qwen-VL, pre-trained 10\%) & \textbf{58.10}     & \textbf{89.75}\\
% \bottomrule
% \end{tabular}
% \end{adjustbox}
% \label{tab:qwen}
% \end{table}

% \textbf{Performance comparison between fully finetuning and LoRA}
\vspace{-0.2in}
\begin{table}[H]
    \centering
    \caption{Results of fully finetuning vs LoRA finetuning on VQA-RAD dataset.}
    \vspace{0.05in}
\begin{adjustbox}{width=0.84\textwidth}
\begin{tabular}{@{\hspace{-0pt}}l@{\hspace{8pt}}c@{\hspace{8pt}}c@{\hspace{8pt}}c@{\hspace{8pt}}c@{\hspace{8pt}}c}%rrrrrrrrrrrr
\toprule
\textbf{Method}         & \textbf{VQA-RAD Open}  & \textbf{VQA-RAD Closed} & \textbf{VQA-RAD Avg} & \textbf{Param}\\
\midrule
LLaVa-Med (LoRa)              & 62.06     & 75.00 & 68.53 & 2.1B \\
ExGra-Med (LoRa)              & 63.55     & 79.41 & 71.48 & 2.1B \\
LLaVa-Med (Fully-finetuning)  & 63.65     & 81.62 & 72.64 & 7B \\
ExGra-Med (Fully-finetuning)  & 66.35     & 83.46 & 74.91 & 7B   \\
\bottomrule
\end{tabular}
\end{adjustbox}
\label{tab:lora}
\end{table}

\textbf{Potentially Hallucination in Extended Captions.} We conducted a user study with \textit{five general practitioners} from top public hospitals (Appendix Section K). In Stage 2 of pre-training, each expert evaluated 200 image-text pairs (1,000 total) across five modalities - chest X-ray, CT, MRI, histology, and others - rating the completeness and accuracy of GPT-4-generated extended captions. As shown in Figures 10–14, most scores ranged from 3 to 5, with few low ratings, confirming the overall consistency and quality of the extended outputs. Also, these extended captions are used only during pre-training to guide latent space alignment. \textit{They are excluded during fine-tuning} on downstream tasks. As such, we argue that a small amount of noise in the extended captions should have minimal impact on overall performance, since they do not directly affect the model’s task-specific adaptations.

\textbf{Other Factors.} We examine ExGra-Med results under following settings:
\begin{itemize}
    \item (i) \textit{generalization to other frozen LLMs} (GPT-4, Gemini \cite{team2023gemini}, and Qwen3-8B LLM \cite{yang2025qwen3}) to generate extended captions and how the method works with \textit{simple synonyms} (Table \ref{tab:exgra-med-other-llms}).
    \item ii) \textit{contribution of coefficient ($\alpha$)} \textit{combine multi-graph alignment with auto-regressive}.
    \item iii) \textit{without using extended contexts}, i.e., simplifying from the three graph alignment to two cross-domain alignments (image vs. original captions).
    \item (iv) \textit{without using original captions}, i.e., only extended ones are used.
    \item (v) applying \textit{message passing to enhance node features}.
    \item (vi) \textit{using multi-graph alignment} in both steps (default uses only Step-2).
    \item (vii) \textit{solving three pair-wise graph alignments} in multi-graph alignment \textit{rather than solving through a barycenter graph} (Eq.(\ref{eq:multi-graph}) in Sec. 
    \ref{sec:scalable-solver}).
    \item (viii) using \textit{parameter-efficient finetuning} with LoRa \cite{hu2022lora} rather than fully finetuning on downstream tasks. 
\end{itemize}
% (i) \textit{generalization to other frozen LLMs} (GPT-4, Gemini \cite{team2023gemini}) to generate extended captions and how the method works with \textit{simple synonyms} (Table \ref{tab:exgra-med-other-llms}); (ii) contribution of coefficient ($\alpha$) \textit{combine multi-graph alignment with auto-regressive}; (iii) \textit{without using extended contexts}, i.e., simplifying from the three graph alignment to two cross-domain alignments (image vs. original captions); (iv) without using original captions, i.e., only extended ones are used; (v) applying \textit{message passing to enhance node features}; (vi) using multi-graph alignment in both steps (default uses only Step-2);  and (vii) \textit{solving three pair-wise graph alignments} in multi-graph alignment \textit{rather than solving through a barycenter graph} (Eq.(\ref{eq:multi-graph}) in Sec. \ref{sec:scalable-solver}). 

Tables \ref{tab:exgra-med-other-llms}-\ref{tab:logra_ablation} show results (i)-(vii) where the most important factors are highlighted.
We further observe that ExGra-Med generalizes effectively across distinct LLM paraphrase generators such as GPT-4, Gemini, and Qwen3-8B. The stable performance across these models indicates that ExGra-Med is not tightly coupled to a specific language model architecture, but instead captures transferable alignment mechanisms applicable to a wide range of paraphrastic contexts.
% Other analysis on using multi-graph alignment in both steps (in default, only Step-2 is used), the average pooling feature, the value of in k-nearest neighbors to construct a graph (Section \ref{sec:graph_construction}), a quality check of extended captions created by GPT-4, and other limitations are discussed in the Appendix.

In Table \ref{tab:lora}, we report the results of ExGra-Med and LLaVA-Med on the VQA-RAD dataset for the (viii) setting. As shown, ExGra-Med consistently outperforms LLaVA-Med even when both models are fine-tuned using LoRA, demonstrating the robustness of our approach under parameter-efficient adaptation. Nevertheless, as expected, LoRA-based fine-tuning yields a modest performance drop compared to full fine-tuning. We believe that to bridge this gap, pre-training on a larger-scale medical instruction-tuning dataset (e.g., the 21M samples curated from MedTrinity \cite{xie2024medtrinity}) would allow the LoRA setup to more closely match the performance of fully fine-tuned models.  Additional analyses on average pooling features, k-nearest neighbors for graph construction (Section \ref{sec:graph_construction}).

\textbf{Current Limitations and Future Work.}
While our experiments have primarily focused on the LLaVa model, it is essential to validate the effectiveness and adaptability of \textsc{ExGra-Med} with alternative architectures, such as the Flamingo model \citep{alayrac2022flamingo}, which has shown promising results in vision-language tasks. Expanding the evaluation to include other state-of-the-art models can provide a broader perspective on the generalizability and robustness of \textsc{ExGra-Med}. Furthermore, incorporating vision encoders or large language models (LLMs) that are specifically pre-trained on medical datasets \citep{chen2023meditron,mh2024lvm,zhao2024foundation,chen2025adapting,zhang2024mm} presents a compelling opportunity to enhance both performance and domain-specific understanding. These specialized models are designed to capture the nuanced characteristics of medical data, which could further enhance the robustness of \textsc{ExGra-Med} in complex biomedical scenarios. Moreover, extending our mechanism to the setting of medical visual chain-of-thought \cite{johnson2019mimiccxr,leduc2025schain} reasoning represents a promising direction for improving both the overall performance and the trustworthiness of the model.

\vspace{-0.05in}
\section{Conclusion}
\vspace{-0.1in}
% We have shown that enforcing triplet correlations among image modalities, their instruction data, and the extended contextual captions can improve vision-language alignment, which is often lacking in models trained by auto-regressive, especially given less pre-training data size as demonstrated for the LLaVa-Med model. We also present \textsc{ExGra-Med}, a new multi-graph alignment algorithm to handle such requirements, which is efficient in training and matches LLAVA-Med’s performance on just 10\% of the training data while outperforming other state-of-the-art methods on various tasks. These findings underscore that selecting the appropriate learning algorithm for training MLLMs is as crucial as scaling model size or data volume.
% We have shown that enforcing triplet correlations among images, instructions, and extended captions significantly enhances vision-language alignment - an area where autoregressive models like \textsc{LLaVA-Med} struggle, especially with limited data and domain shift, resulting in a data-hungry phenomenon in pre-training data. To this end, we introduce \textsc{ExGra-Med}, a multi-graph alignment algorithm that trains efficiently, matches \textsc{LLaVA-Med} using just 10\% of the data, and outperforms other SOTA models across tasks. These results underscore that choosing the right learning algorithm for LLMs is as vital as scaling model size or data volume.
We have shown that enforcing triplet correlations among images, instructions, and extended captions significantly enhances vision-language alignment - an area where autoregressive models like \textsc{LLaVA-Med} struggle, particularly under limited data and domain shift, which manifests as a strong dependence on large-scale pre-training data. To this end, we introduce \textsc{ExGra-Med}, a multi-graph alignment algorithm that trains efficiently, matches \textsc{LLaVA-Med} using only 10\% of the data, and outperforms other state-of-the-art models across tasks. These results highlight that selecting an effective learning algorithm for LLMs is as crucial as scaling model size or data volume.

\newpage
\section*{Acknowledment}
This work was supported by Deutsche Forschungsgemeinschaft (DFG, German Research Foundation) under Germany’s Excellence Strategy - EXC 2075 – 390740016,
the DARPA ANSR program under award FA8750-23-
2-0004, the DARPA CODORD program under award
HR00112590089. The authors thank the International Max
Planck Research School for Intelligent Systems (IMPRS-IS)
for supporting Duy M. H. Nguyen. Duy M. H. Nguyen and Daniel Sonntag are also supported by the No-IDLE project (BMBF,
01IW23002), the MASTER project (EU, 101093079), and the Endowed Chair of Applied Artificial
Intelligence, Oldenburg University.
\bibliography{references}
\bibliographystyle{abbrv}

%%%%%%%%%%%%%%%%%%%%%%%%%%%%%%%%%%%%%%%%%%%%%%%%%%%%%%%%%%%%

\newpage
\appendix
\onecolumn
\begin{center}
\textsc{\Large Supplementary Material for \\ \vspace{.2em} ``ExGra-Med: Extended Context Graph Alignment for Medical Vision-Language Models''}
\end{center}
% Add appendix to the main TOC
\addcontentsline{toc}{section}{Supplementary Material}

\tableofcontents
\addtocontents{toc}{\protect\setcounter{tocdepth}{2}}
\vspace{-0.1in}
\captionsetup{font=normalsize}
\newpage
\section{Proofs of the Main Theoretical Results}\label{proof_main_theoretical_results}

In this section, we provide detailed technical proofs of our main theoretical results. To accomplish this, we first introduce the fundamental concepts and materials that will be utilized in the proofs of Theorems \ref{theorem_SGM_distance} and \ref{theorem_constant_geodesic}.

We consider labelled graphs as tuples of the form $\gG = (\gV,\gE,\gL_f,\gL_s)$, where the labelling function $\gL_f:\gV \mapsto \cF$ assigns each vertex $v_i\in\gV$ to a feature $f_i = \gL_f(v_i)$ in some feature space $(\cF,d_f)$. Similarly, we denote $\gL_s:\gV \mapsto \cS$ as a structure function which links each vertex $v_i \in \gV$ with its structure information $s_i = \gL_s(v_i)$, \eg~edge information, in some structure space $(\cS,d_s)$.
By associating a weight to each vertex, we allow the graph $\gG$ to be represented by a fully supported mixing measure $\mu = \sum_{i=1}^N w_i \delta_{(f_i,s_i)}$ over the product between feature space and structure space  $\gF\times\gS$. Notably, $\mu$ is not necessarily a probability measure as the summation of its weights can be different from one. We have the vertex affinity matrix between two graphs as $\mA^v \in \R^{M \times N}$, where $\emA^v_{i,j} = (d_f(f_i,f_j)){i,j}$. Structural similarity is measured by pairwise distances within each graph, represented by $\tA^{e} \in \R^{|\gE_1||\gE_2|}$, with $\etA^e_{i,j,k,l} = |d_{s}(s_i,s_k) - d_{s}(s_j,s_l)|$, where $d_s(\cdot)$ models node distance, such as the shortest path.
We then define the space of all structured graphs $(\gF\times\gS,d_f,\mu)$ over a metric feature space $(\gF,d_f)$ as $\sS(\gF)$, where  $(\cS,d_s)$ is a metric structure space and $\mu = \sum_{i=1}^N w_i \delta_{(f_i,s_i)}$ is a mixing measure over $\gF\times\gS$.

\subsection{Proof of Theorem \ref{theorem_SGM_distance}}\label{proof_theorem_SGM_distance}
    For the sake of simplicity, we 
    denote the labeled graphs $\gG$ and structured graphs discussed above only by $\mu$ the whole structured graph.

To prove Theorem \ref{theorem_SGM_distance}, for any two graphs $\gG_1$ and $\gG_2$ in the structured graph space $\sS(\gF)$, described respectively by their mixing measure $\mu_1 = \sum_{i=1}^M w_{1i} \delta_{(f_{1i},s_{1i})}$ and $\mu_2 = \sum_{j=1}^N w_{2j} \delta_{(f_{2j},s_{2j})}$, respectively, we wish to prove the following properties:
\begin{enumerate}
        \item Positivity: $d_{\text{SGA}}(\gG_1,\gG_2) > 0$ for any $\gG_1 \neq \gG_2$.

        \item Equality relation: $d_{\text{SGA}}(\gG_1,\gG_2) = 0$ if and only if $\gG_1 = \gG_2$.

        \item Symmetry: $d_{\text{SGA}}(\gG_1,\gG_2) = d_{\text{SGA}}(\gG_2,\gG_1)$.

        \item Triangle inequality: $d_{\text{SGA}}(\gG_1,\gG_3) \le d_{\text{SGA}}(\gG_1,\gG_2) + d_{\text{SGA}}(\gG_2,\gG_3)$ for any graph $\gG_3$.
    \end{enumerate}
Note first that 1. Positivity and 3. Symmetry hold trivially.

{\bf Proof of 2. Equality relation.}
The equality relation immediately follows the following Proposition \ref{proposition_SGM_equality_relation}, which is proved in Appendix \ref{proof_proposition_SGM_equality_relation}.
\begin{proposition}[Equality relation]\label{proposition_SGM_equality_relation}      
    For any two graphs $\gG_1$ and $\gG_2$ in the structured graph space $\sS(\gF)$,
    described respectively by their mixing measure $\mu_1 = \sum_{i=1}^M w_{1i} \delta_{(f_{1i},s_{1i})}$ and $\mu_2 = \sum_{j=1}^N w_{2j} \delta_{(f_{2j},s_{2j})}$, 
    it holds $d_{\text{SGA}}(\gG_1,\gG_2) = 0$ if and only if $M=N$ and there exists a bijection $\sigma:[M] \mapsto [M]$ such that:
    \begin{enumerate}
        \item[E1.] $\forall i \in [M]: w_{1i} = w_{2 \sigma(i)}$.
        \item[E2.] $\forall i \in [M]: f_{1i} = f_{2\sigma(i)}$.
        \item[E3.] $\forall i,k \in [M]^2: d_s(s_{1i},s_{1k}) = d_s(s_{2\sigma(i)},s_{2\sigma(k)})$.
    \end{enumerate}
\end{proposition}
%%
% \begin{proof}[Proof of Proposition \ref{proposition_SGM_equality_relation}]

% \end{proof} 

    {\bf Proof of 4. Triangle inequality.} 
    Let us consider two arbitrary graphs $\gG_1$ and $\gG_2$, described respectively by their probability measure $\mu_1 = \sum_{i=1}^M w_{1i} \delta_{(f_{1i},s_{1i})}$ and $\mu_2 = \sum_{j=1}^N w_{2j} \delta_{(f_{2j},s_{2j})}$.
    For any graph $\gG_3$ described by its probability measure $\mu_3 = \sum_{i=1}^K w_{3k} \delta_{(f_{3k},s_{3k})}$, we define $\mP \in \gA(\gG_1,\gG_2)$ and $\mQ \in \gA(\gG_2,\gG_3)$ as two optimal couplings of the SGA distance between $\mu_1$ and $\mu_2$ and $\mu_2$ and $\mu_3$, respectively, \ie~ 
    \begin{align}
        \mP \in \gA(\gG_1,\gG_2) = \left\{\mP \in \{0,1\}^{M\times N}:\sum_{i=1}^M\emP_{i,j}=w_{1j} = 1,\sum_{j=1}^N\emP_{i,j} =w_{2i}= 1\right\},\nonumber\\
        \mQ \in \gA(\gG_2,\gG_3) = \left\{\mQ \in \{0,1\}^{N\times K}:\sum_{j=1}^N\emQ_{j,k}=w_{2k} = 1,\sum_{k=1}^K\emQ_{j,k} =w_{3j}= 1\right\}.\nonumber
    \end{align}
    
    We then construct $\mR = \left(\sum_{j} \frac{P_{i,j} Q_{j,k}}{w_{2j}}\right)_{i,k}$. Then it holds that $\mR \in \gA(\gG_1,\gG_3)$. Indeed, we have 
    \begin{align}
        \sum_{i}\emR_{i,k} = \sum_{i}\sum_{j} \frac{P_{i,j} Q_{j,k}}{w_{2j}} =  \sum_{j} \sum_{i}P_{i,j} \frac{ Q_{j,k}}{w_{2j}} =  \sum_{j} w_{1j} \frac{ Q_{j,k}}{w_{2j}} =  \sum_{j} Q_{j,k} =1. \nonumber
    \end{align}
    By the suboptimality of $\mR$, the triangle inequalities of $d_f$ and $|\cdot|$, we have
    \begin{align}
        d_{\text{SGA}}(\gG_1,\gG_3) &\le \sum_{i,j,k,l}\left[d_f(f_{1i},f_{3j})+|d_{s}(s_{1i},s_{1k})-d_{s}(s_{3j},s_{3l})|\right]\emR_{i,j} \emR_{k,l}
        \nonumber\\
        &=  \sum_{i,j,k,l}\left[d_f(f_{1i},f_{3j})+|d_{s}(s_{1i},s_{1k})-d_{s}(s_{3j},s_{3l})|\right]\sum_{t} \frac{P_{i,t} Q_{t,j}}{w_{2t}} \sum_{d} \frac{P_{k,d} Q_{d,l}}{w_{2d}}
        \nonumber\\
        &=  \sum_{i,j,k,l,t,d}\left[d_f(f_{1i},f_{3j})+|d_{s}(s_{1i},s_{1k})-d_{s}(s_{3j},s_{3l})|\right] \frac{P_{i,t} Q_{t,j}}{w_{2t}} \frac{P_{k,d} Q_{d,l}}{w_{2d}}
        \nonumber\\
        &\le  \sum_{i,j,k,l,t,d}\left[d_f(f_{1i},f_{2t})+d_f(f_{2t},f_{3j})\right] \frac{P_{i,t} Q_{t,j}}{w_{2t}} \frac{P_{k,d} Q_{d,l}}{w_{2d}}
        \nonumber\\
        &\quad +\sum_{i,j,k,l,t,d}\left[|d_{s}(s_{1i},s_{1k})-d_{s}(s_{2t},s_{2d})|+|d_{s}(s_{2t},s_{2d})-d_{s}(s_{3j},s_{3l})|\right] \frac{P_{i,t} Q_{t,j}}{w_{2t}} \frac{P_{k,d} Q_{d,l}}{w_{2d}}
        \nonumber\\
        &=\sum_{i,j,k,l,t,d}\left[d_f(f_{1i},f_{2t})+|d_{s}(s_{1i},s_{1k})-d_{s}(s_{2t},s_{2d})|\right] \frac{P_{i,t}P_{k,d}}{w_{2t}} \frac{ Q_{t,j}Q_{d,l}}{w_{2d}}
        \nonumber\\
        &\quad +\sum_{i,j,k,l,t,d}\left[d_f(f_{2t},f_{3j})+|d_{s}(s_{2t},s_{2d})-d_{s}(s_{3j},s_{3l})|\right] \frac{P_{i,t} Q_{t,j}}{w_{2t}} \frac{P_{k,d} Q_{d,l}}{w_{2d}}
        \nonumber\\
        &=\sum_{i,k,t,d}\left[d_f(f_{1i},f_{2t})+|d_{s}(s_{1i},s_{1k})-d_{s}(s_{2t},s_{2d})|\right] P_{i,t}P_{k,d} \sum_{j}\frac{ Q_{t,j}}{w_{2t} } \sum_{l} \frac{Q_{d,l}}{w_{2d}}
        \nonumber\\
        &\quad +\sum_{j,l,t,d}\left[d_f(f_{2t},f_{3j})+|d_{s}(s_{2t},s_{2d})-d_{s}(s_{3j},s_{3l})|\right] Q_{t,j} Q_{d,l} \sum_{i}\frac{P_{i,t} }{w_{2t}} \sum_{k}\frac{P_{k,d} }{w_{2d}}.\nonumber
    \end{align}
    Note that we have 
    \begin{align}
        \sum_{j}\frac{ Q_{t,j}}{w_{2t} } =\sum_{l} \frac{Q_{d,l}}{w_{2d}} = \sum_{i}\frac{P_{i,t} }{w_{2t}} =\sum_{k}\frac{P_{k,d} }{w_{2d}} =1.\nonumber
    \end{align}
    This is how we achieve the desired result, because      
    % \begin{align}
    %     d_{\text{SGA}}(\gG_1,\gG_3)
    %     &\le\sum_{i,k,t,d}\left[d_f(f_{1i},f_{2t})+|d_{s}(s_{1i},s_{1k})-d_{s}(s_{2t},s_{2d})|\right] P_{i,t}P_{k,d} 
    %     \nonumber\\
    %     %%
    %     &\quad +\sum_{j,l,t,d}\left[d_f(f_{2t},f_{3j})+|d_{s}(s_{2t},s_{2d})-d_{s}(s_{3j},s_{3l})|\right] Q_{t,j} Q_{d,l}
    %     \nonumber\\
    %     %%
    %     &=d_{\text{SGA}}(\gG_1,\gG_2)+d_{\text{SGA}}(\gG_2,\gG_3) \text{ (since $\mP$ and $\mQ$ are the optimal plans)}.\nonumber
    % \end{align} 
\begin{align*}
    d_{\text{SGA}}(\gG_1,\gG_3)
    &\le \sum_{i,k,t,d}\left[d_f(f_{1i},f_{2t})+|d_{s}(s_{1i},s_{1k})-d_{s}(s_{2t},s_{2d})|\right] P_{i,t}P_{k,d} \\
    &\quad + \sum_{j,l,t,d}\left[d_f(f_{2t},f_{3j})+|d_{s}(s_{2t},s_{2d})-d_{s}(s_{3j},s_{3l})|\right] Q_{t,j} Q_{d,l} \nonumber
    \\
    &= d_{\text{SGA}}(\gG_1,\gG_2) + d_{\text{SGA}}(\gG_2,\gG_3) \text{ (since $\mP$ and $\mQ$ are the optimal plans)}.\nonumber
\end{align*}
\subsection{Proof of Theorem \ref{theorem_constant_geodesic}}\label{proof_theorem_constant_geodesic}
Theorem \ref{theorem_constant_geodesic} enables us to characterise the optimal transport problem between two measures as a curve in the space of measures, with the objective of minimising its total length. Furthermore, this formulation is beneficial for deriving global minima results for non-convex particles in gradient descent in an optimisation context, which is a valuable application of gradient flows \citep{chizat_global_2018}. By definition, a geodesic between $\gG_1$ and $\gG_2$ is a shortest path between these two graphs. In particular, the computation of distances along constant speed geodesic paths is a relatively straightforward process, as these paths are directly embedded into the real line $\R$ as follows: $d_{\text{SGA}}(\gG_1,\gG_2) = |t-u|^{-1}d_{\text{SGA}}(p(u),p(t))$, for all $ 0 \le u \neq  t \le 1$ and for any path (continuous map) $p$ connect $\gG_1$ to $\gG_2$ such that $p(u) = \gG_1$ and $p(t) = \gG_2$. 
To prove Theorem \ref{proof_theorem_constant_geodesic}, it is necessary to collect fundamental material using Definition \ref{definition_geodesic} from metric geometry for a general metric space $(\sM,d)$.
\begin{definition}[Length and geodesic spaces]\label{definition_geodesic}
    Let $(\sM,d)$ be a metric space and two points $x,y \in \sM$. 
    We say that a path (curve) $p:[0,1] \mapsto\sM$ connect or join $x$ to $y$ if $p(0) = x$ and $p(1) = y$ and $p$ is a continuous map. 
    
    We also define the length $L(p) \in  \R$ of a path $p:[0,1] \mapsto \sM$ as  $$L(p):=\sup \sum_{i=1}^n d(p(t_i),p(t_{i+1}))$$ where we take the supremum over all $n\ge 1$ and all $n$-tuples $t_1<\ldots<t_n$ in $[0,1]$. 

    We denote a metric space $\sM$ as a length space if for all $x,y \in \sM$, $d(x,y) = \inf_p L(p)$ where the infimum is taken over all paths $p$ connecting $x$ to $y$.
    
    We call a length space as a geodesic space if for all $x,y\in\sM$, there exists a path $p(x,y):[0,1] \mapsto \sM$ such that $$d(x,y) = \min_{p(x,y)} L(p(x,y)).$$ 
    We also denote the path $p(x,y)$ as a geodesic between $x$ and $y$.

    Finally, we define a path $p:[0,1] \mapsto \sM$ as a constant speed geodesic if and only if $$d(p(u),p(t)) = |t-u|d(p(0),p(1)), \forall u,t \in[0,1].$$
\end{definition}

    For the proof of Theorem \ref{theorem_constant_geodesic}, we first consider an optimal coupling $\mV^*$ for SGA distance between two graphs $\gG_1$ and $\gG_2$, \ie~
    $$d_{\text{SGA}}(\gG_1,\gG_2) = \min_{\mV \in \gA(\gG_1,\gG_2)}  O(\mA^v,\tA^e,\mV) = O(\mA^v,\tA^e,\mV^*),$$ described respectively by their mixing measure $\mu_0 = \sum_{i=1}^M w_{0i} \delta_{(f_{0i},s_{0i})}$ and $\mu_1 = \sum_{j=1}^N w_{1j} \delta_{(f_{1j},s_{1j})}$.
    Moreover, for any $t\in[0,1]$ we define $\nu_t:\gF\times\gS_0\times \gF\times\gS_1 \mapsto \gF\times\gS_0\times \gS_1$ such that $$\nu_t(f_0,s_0,f_1,s_1)=((1-t)f_0+tf_1,s_0,s_1), \text { and }\mu_t := \nu_t\#\mV^* = \sum_{i=1}^M \sum_{j=1}^N \emV^*_{i,j} \delta_{((1-t)f_0+tf_1,s_{0i},s_{1j})},$$
    and  on the metric space $\gS_0\times \gS_1$, we define the distance
    $$d_t:=(1-t)d_{s_0} \oplus t d_{s_1}:(1-t)d_{s_0} \oplus t d_{s_1}((s_{0i},s_{0j}),(s_{1k},s_{1l})) = (1-t)d_{s}(s_{0i},s_{1k})+td_{s}(s_{0j},s_{1l})$$
    for any $((s_{0i},s_{0j}),(s_{1k},s_{1l})) \in \gS_0\times \gS_1$.
    Here, we denote $\#$ the push-forward operator such that $\nu_t\#\mV^*(\sA) = \mV^*(\nu_t^{-1}(\sA))$ for any Borel sets of a $\sigma$-albegra.
    For simplicity, we only consider $(\gF,d_f) = (\R^d,\|\cdot\|)$ where $\|\cdot\|$ is the Euclidean norm.
    
    Then we aim to prove that $(\gF\times\gS_0\times \gS_1,(1-t)d_{s_0} \oplus t d_{s_1},\mu_t)_{t\in[0,1]}$ is a constant speed geodesic joining $(\gF\times\gS_0,d_{s0},\mu_0))$ and $(\gF\times\gS_1,d_{s1},\mu_1))$, for arbitrary elements $(\gF\times\gS_0,d_{s0},\mu_0))$ and $(\gF\times\gS_1,d_{s1},\mu_1))$ in the metric space $(\sS(\gF),d_{\text{SGA}})$.  

To do so, we consider any $u,t \in[0,1]$ such that $u \neq t$. By definition, we have to prove that 
\begin{align}\label{equation_proof_geodesic}
    d_{\text{SGA}}(\mu_u,\mu_t) = |t-u|  d_{\text{SGA}}(\mu_0,\mu_1).
\end{align}
Indeed, to prove equation (\ref{equation_proof_geodesic}), we first recall that 
\begin{align}
    \mu_u &:= \nu_u\#\mV^* = \sum_{i=1}^M \sum_{j=1}^N \emV^*_{i,j} \delta_{((1-u)f_0+uf_1,s_{0i},s_{1j})},\nonumber\\
    \mu_t &:= \nu_t\#\mV^* = \sum_{i=1}^M \sum_{j=1}^N \emV^*_{i,j} \delta_{((1-t)f_0+tf_1,s_{0i},s_{1j})},\nonumber\\
    d_{\text{SGA}}(\mu_0,\mu_1) &= \sum_{i,j,k,l}\left[d_f(f_{0i},f_{1j})+|d_{s}(s_{0i},s_{1k})-d_{s}(s_{0j},s_{1l})|\right]\emV^*_{i,j} \emV^*_{k,l}.\nonumber
\end{align}
We then define the coupling $\gamma^{u,t} = (\mu_u\times\mu_t)\#\mV^* \in \gA(\mu_u,\mu_t)$.
By the suboptimality of $\gamma^{u,t} $, it holds that:
\begin{align}
    d_{\text{SGA}}(\mu_u,\mu_t) &
    \le \sum_{i,j,k,l}\left[d_f(f_{0i},f_{1j})+|d_{t}((s_{0i},s_{0j}),(s_{1k},s_{1l}))-d_{u}((s_{0i},s_{0j}),(s_{1k},s_{1l}))|\right]\gamma^{u,t}_{i,j} \gamma^{u,t}_{k,l}\nonumber\\
    &=\sum_{i,j,k,l}\Big[d_f((1-t)f_{0i}+t f_{1j},(1-u)f_{0i}+u f_{1j}) \nonumber\\
    & \quad+|(1-t)d_{s}(s_{0i},s_{1k})+td_{s}(s_{0j},s_{1l})-(1-u)d_{s}(s_{0i},s_{1k})-ud_{s}(s_{0j},s_{1l})|\Big]\emV^*_{i,j} \emV^*_{k,l}\nonumber\\
    &= \sum_{i,j,k,l}\left[(t-u)d_f(f_{0i},f_{1j})+|(t-u)d_{s}(s_{0i},s_{1k})-(t-u)d_{s}(s_{0j},s_{1l})|\right]\emV^*_{i,j} \emV^*_{k,l}\nonumber\\
    &=|t-u| \sum_{i,j,k,l}\left[d_f(f_{0i},f_{1j})+|d_{s}(s_{0i},s_{1k})-d_{s}(s_{0j},s_{1l})|\right]\emV^*_{i,j} \emV^*_{k,l}\nonumber\\
    &=|t-u|  d_{\text{SGA}}(\mu_0,\mu_1).\nonumber
\end{align}
Here, we used the fact that $d_f$ is the Euclidean norm, hence $$d_f((1-t)f_{0i}+t f_{1j},(1-u)f_{0i}+u f_{1j}) = \|(1-t)f_{0i}+t f_{1j}-(1-u)f_{0i}-u f_{1j}\|=|t-u| d_f(f_{0i},f_{1j}).$$
Therefore, we have 
\begin{align}\label{eq_geodesic1}
    d_{\text{SGA}}(\mu_u,\mu_t) &
    \le |t-u|  d_{\text{SGA}}(\mu_0,\mu_1).
\end{align}
The remaining task is to prove that
\begin{align}\label{eq_geodesic2}
    d_{\text{SGA}}(\mu_u,\mu_t) &
    \ge |t-u|  d_{\text{SGA}}(\mu_0,\mu_1).
\end{align}
To show that this inequality, we note that via the triangle inequality of $d_{\text{SGA}}$ and for any $0 \le u \le t \le 1$, it holds that
\begin{align}
     d_{\text{SGA}}(\mu_0,\mu_1) &\le d_{\text{SGA}}(\mu_0,\mu_u) + d_{\text{SGA}}(\mu_u,\mu_t) + d_{\text{SGA}}(\mu_t,\mu_1)\nonumber\\
     &\le u d_{\text{SGA}}(\mu_0,\mu_1) +  (t-u) d_{\text{SGA}}(\mu_0,\mu_1) +(1-t) d_{\text{SGA}}(\mu_0,\mu_1)\nonumber\\
     &= d_{\text{SGA}}(\mu_0,\mu_1).\nonumber
\end{align}
Hence,  for any $0 \le u \le t \le 1$, we obtain
\begin{align}
      &d_{\text{SGA}}(\mu_0,\mu_u) + d_{\text{SGA}}(\mu_u,\mu_t) + d_{\text{SGA}}(\mu_t,\mu_1)= u d_{\text{SGA}}(\mu_0,\mu_1) +  (t-u) d_{\text{SGA}}(\mu_0,\mu_1) +(1-t) d_{\text{SGA}}(\mu_0,\mu_1).\label{eq_geodesic3}
\end{align}
Suppose that $$d_{\text{SGA}}(\mu_u,\mu_t) < (t-u) d_{\text{SGA}}(\mu_0,\mu_1).$$ Then combining with the fact that 
$$d_{\text{SGA}}(\mu_0,\mu_u) \le u d_{\text{SGA}}(\mu_0,\mu_1), \text{ and }  d_{\text{SGA}}(\mu_t,\mu_1) \le (1-t) d_{\text{SGA}}(\mu_0,\mu_1),$$
we have
\begin{align}
      &d_{\text{SGA}}(\mu_0,\mu_u) + d_{\text{SGA}}(\mu_u,\mu_t) + d_{\text{SGA}}(\mu_t,\mu_1)< u d_{\text{SGA}}(\mu_0,\mu_1) +  (t-u) d_{\text{SGA}}(\mu_0,\mu_1) +(1-t) d_{\text{SGA}}(\mu_0,\mu_1)\nonumber.
\end{align}
This leads to the contradiction with the equation (\ref{eq_geodesic3}.) 
Hence the desired inequality in (\ref{eq_geodesic2}) holds.
Finally, we obtain
\begin{align}
    d_{\text{SGA}}(\mu_u,\mu_t) &
    = |t-u|  d_{\text{SGA}}(\mu_0,\mu_1).
\end{align}

\section{Proofs of Technical Results}\label{proof_techinical_result}

\subsection{Proof of Proposition \ref{proposition_SGM_equality_relation}}\label{proof_proposition_SGM_equality_relation}
 First, let us suppose that $d_{\text{SGA}}(\gG_1,\gG_2) = 0$. We wish to prove the existence of a bijection $\sigma$ satisfying E1, E2, and E3. Indeed, let $\mV \in \gA(\gG_1,\gG_2)$ be any admissible transportation plan that encode a valid matching between $\gG_1$ and $\gG_2$. Then we define:
    \begin{align}
        d(s_{1i},s_{1k}) = \frac{1}{2}\left[d_f(f_{1i},f_{1k})+d_s(s_{1i},s_{1k})\right], \quad \forall i,k\in[M]^2, \label{eq_d_s1ik}\\
        d(s_{2j},s_{2l}) = \frac{1}{2}\left[d_f(f_{2j},f_{2l})+d_s(s_{2j},s_{2l})\right], \quad \forall j,l\in[M]^2\label{eq_d_s2jl}.
    \end{align}
    Recall that we then define SGM discrepancy as:
    \begin{align}\label{eq_SGM_proof}
        d_{\text{SGA}}(\gG_1,\gG_2) &= \min_{\mV \in \gA(\gG_1,\gG_2)} \left(\sum_{i,j}\emA^v_{i,j}\emV_{i,j}+\sum_{i,j,k,l}\etA^e_{i,j,k,l}\emV_{i,j} \emV_{k,l}\right)
        = \min_{\mV \in \gA(\gG_1,\gG_2)}  O(\mA^v,\tA^e,\mV)\nonumber\\
        &= \min_{\mV \in \gA(\gG_1,\gG_2)} \left[ O_f(\mA^v,\mV) + O_s(\tA^e,\mV)\right].
    \end{align}

It should be recalled that the vertex affinity matrix $\mA^v \in \R^{M \times M}$, defined as $A^v_{i,j} = (d_f(f_{1i},f_{2j}))_{i,j}$, was introduced in the previous section. The edge affinity tensor, denoted by $\tA^e$, is defined as follows:
$\etA^e_{i,j,k,l} = |d_{s}(s_{1i},s_{1k})-d_{s}(s_{2j},s_{2l})|$.

Let $\mV^*$ be the optimal coupling for $d_{\text{SGA}}(\gG_1,\gG_2)$. Then we have
    \begin{align}
         O_f(\mA^v,\mV^*) + O_s(\tA^e,\mV^*) = \min_{\mV \in \gA(\gG_1,\gG_2)} \left[ O_f(\mA^v,\mV) + O_s(\tA^e,\mV)\right] = d_{\text{SGA}}(\gG_1,\gG_2) =0.
    \end{align}
Since both  $O_f(\mA^v,\mV^*)$ and $O_s(\tA^e,\mV^*)$ are non-negative, we conclude that $O_f(\mA^v,\mV^*)=O_s(\tA^e,\mV^*)=0$.
Now we wish to use the following Lemma \ref{proof_lemma_Os_A_overline}, which is proved in Appendix \ref{proof_lemma_Os_A_overline}.
\begin{lemma}\label{lemma_Os_A_overline}
    Given the definition of $\overline{\etA}^e_{i,j,k,l} = |d(s_{1i},s_{1k})-d(s_{2j},s_{2l})|$ where $d(s_{1i},s_{1k})$ and $d(s_{2j},s_{2l})$ are provided in equations~(\ref{eq_d_s1ik}) and~(\ref{eq_d_s2jl}), respectively, it holds that 
    \begin{align}\label{eq_Os_A_overline}
    O_s(\overline{\tA}^e,\mV^*)=\sum_{i,j,k,l}\overline{\etA}^e_{i,j,k,l}\emV^*_{i,j} \emV^*_{k,l}=\sum_{i,j,k,l}|d(s_{1i},s_{1k})-d(s_{2j},s_{2l})|\emV^*_{i,j} \emV^*_{k,l}=0.
    \end{align}
    Moreover, there exists a bijective $\sigma:[M] \mapsto [N]$ with $M=N$ satisfies the weight and distance $d$ preserving isometry  as follows:
    \begin{enumerate}
        \item[E1.] $\forall i \in [M]: w_{1i} = w_{2 \sigma(i)}$.

        \item[E3*.] $\forall i,k \in [M]^2: d(s_{1i},s_{1k}) = d(s_{2\sigma(i)},s_{2\sigma(k)})$.
    \end{enumerate}
\end{lemma}
Because we have $\mV^*$ is the optimal coupling \wrt~the distance $d$ such that
\begin{align}
     O_s(\overline{\tA}^e,\mV^*) = \min_{\mV \in \gA(\gG_1,\gG_2)} O_s(\overline{\tA}^e,\mV) = 0,
\end{align}
$\mV^*$ is supported by $\sigma$ and satisfies $\mV^* = \mI_{M\times N}\times \sigma$. Therefore, $O_f(\mA^v,\mV^*)=\sum_{i,j} d_f(f_{1i},f_{2\sigma(i)}) \emV^*_{i,j}=\sum_{i} d_f(f_{1i},f_{2\sigma(i)}) \sum_{j} \emV^*_{i,j}=\sum_{i} d_f(f_{1i},f_{2\sigma(i)})=0$.
Here, we used the fact that
\begin{align}
    \mV^* \in \gA(\gG_1,\gG_2) = \left\{\mV \in \{0,1\}^{M\times N}:\sum_{i=1}^M\emV_{i,j}=w_{1j} = 1,\sum_{j=1}^N\emV_{i,j} =w_{2i}= 1\right\}.\nonumber
\end{align}
Note that $d_f(f_{1i},f_{2\sigma(i)}), i\in[M]$ are all non-negative. This leads to $d_f(f_{1i},f_{2\sigma(i)}) = 0, \forall i\in[M]$. This is equivalent to $f_{1i}=f_{2\sigma(i)}, \forall i\in[M]$ since $d_f$ is a metric, which is the desired E2. Therefore, we also have $d_f(f_{1i},f_{1k}) = d_f(f_{2\sigma(i)},f_{2\sigma(k)}), \forall i,k\in[M]$. Combining equations~(\ref{eq_d_s1ik}),~(\ref{eq_d_s2jl}), and E3*, we have
\begin{align}
    d(s_{1i},s_{1k}) &= \frac{1}{2}\left[d_f(f_{1i},f_{1k})+d_s(s_{1i},s_{1k})\right], \\
    d(s_{2\sigma(i)},s_{\sigma(k)}) &= \frac{1}{2}\left[d_f(f_{2\sigma(i)},f_{2\sigma(k)})+d_s(s_{2\sigma(i)},s_{2\sigma(k)})\right], \quad \forall i,k\in[M]^2.
\end{align}
This leads to the desired result, \ie~E3. $d_s(s_{1i},s_{1k})=d_s(s_{2\sigma(i)},s_{2\sigma(k)}), \forall i,k\in[M]^2$.

Now, let us suppose that $M = N$ there exists a bijection $\sigma:[M] \mapsto[N]$ satisfying E1, E2, and E3.  We wish to prove that $d_{\text{SGA}}(\gG_1,\gG_2) = 0$. Then we can consider the transportation plan $\mV^* = \mI_{M\times N} \times \sigma$, \ie~$\mV^*$ is associated with $i \mapsto i$ and $j \mapsto \sigma(i)$. Using E1, it holds that $\mV^* \in \gA(\gG_1,\gG_2)$. Moreover, via E2 and E3, we also have 
 \begin{align}
 % \label{eq_SGM_proof_equality2}
        d_{\text{SGA}}(\gG_1,\gG_2) &= \min_{\mV \in \gA(\gG_1,\gG_2)}  O(\mA^v,\tA^e,\mV) \le  \sum_{i,j}\emA^v_{i,j}\emV^*_{i,j}+\sum_{i,j,k,l}\etA^e_{i,j,k,l}\emV^*_{i,j} \emV^*_{k,l}
        \nonumber\\
        &=  \sum_{i,j}d_f(f_{1i},f_{2j})\emV^*_{i,j}+\sum_{i,j,k,l}|d_{s}(s_{1i},s_{1k})-d_{s}(s_{2j},s_{2l})|\emV^*_{i,j} \emV^*_{k,l}
        \nonumber\\
        &=  \sum_{i,j}d_f(f_{1i},f_{2\sigma(i)})\emV^*_{i,j}+\sum_{i,j,k,l}|d_{s}(s_{1i},s_{1k})-d_{s}(s_{2\sigma(i)},s_{2\sigma(k)})|\emV^*_{i,j} \emV^*_{k,l}=0.\nonumber
    \end{align}
    This leads to the desired result that $d_{\text{SGA}}(\gG_1,\gG_2) = 0$.

\subsection{Proof of Lemma \ref{lemma_Os_A_overline}}\label{proof_lemma_Os_A_overline}
% \vspace{-4.5em}
By definitions and the triangle inequalities of the metric $d_f$ and $d_s$, we have
    \begin{align}\label{eq_Os_A_overline_proof}
        O_s(\overline{\tA}^e,\mV^*)&=\sum_{i,j,k,l}|d(s_{1i},s_{1k})-d(s_{2j},s_{2l})|\emV^*_{i,j} \emV^*_{k,l}\nonumber\\
        &= \sum_{i,j,k,l}|\frac{1}{2}\left[d_f(f_{1i},f_{1k})+d_s(s_{1i},s_{1k})\right]-\frac{1}{2}\left[d_f(f_{2j},f_{2l})+d_s(s_{2j},s_{2l})\right]|\emV^*_{i,j} \emV^*_{k,l}\nonumber\\
        &=  \sum_{i,j,k,l}\left|\frac{1}{2}\left[d_f(f_{1i},f_{1k})-d_f(f_{2j},f_{2l})\right]+\frac{1}{2}\left[d_s(s_{1i},s_{1k})-d_s(s_{2j},s_{2l})\right]\right|\emV^*_{i,j} \emV^*_{k,l}\nonumber\\
        &\le  \frac{1}{2}\sum_{i,j,k,l}\left|d_f(f_{1i},f_{1k})-d_f(f_{2j},f_{2l})\right|\emV^*_{i,j} \emV^*_{k,l}+\frac{1}{2} \sum_{i,j,k,l}\left|d_s(s_{1i},s_{1k})-d_s(s_{2j},s_{2l})\right|\emV^*_{i,j} \emV^*_{k,l}\nonumber\\
         &=  \frac{1}{2}\sum_{i,j,k,l}\left|d_f(f_{1i},f_{1k})-d_f(f_{2j},f_{2l})\right|\emV^*_{i,j} \emV^*_{k,l}+\frac{1}{2}O_s(\tA^e,\mV^*)\nonumber\\
      &=  \frac{1}{2}\sum_{i,j,k,l}\left|d_f(f_{1i},f_{1k})-d_f(f_{2j},f_{2l})\right|\emV^*_{i,j} \emV^*_{k,l} \text{ (since $O_s(\tA^e,\mV^*) = 0$)}.
    \end{align}
    Using the triangle inequality of the metric $d_f$ again, we have
    \begin{align}
        d_f(f_{1i},f_{1k}) \le d_f(f_{1i},f_{2j}) + d_f(f_{2j},f_{2l}) + d_f(f_{2l},f_{1k}),\nonumber\\
        d_f(f_{2j},f_{2l}) \le d_f(f_{2j},f_{1i}) + d_f(f_{1i},f_{1k}) + d_f(f_{1k},f_{2l}).\nonumber
    \end{align}

    This is equivalent to 
    \begin{align}
    d_f(f_{1i},f_{1k}) - d_f(f_{2j},f_{2l}) \le d_f(f_{1i},f_{2j}) + d_f(f_{1k},f_{2l}),\nonumber\\
    d_f(f_{2j},f_{2l})-d_f(f_{1i},f_{1k}) \le d_f(f_{1i},f_{2j})  + d_f(f_{1k},f_{2l}).\label{eq_triangle_dfikjl}
    \end{align}
    We consider two sets $I_1 = \{i,j,k,l: d_f(f_{1i},f_{1k}) - d_f(f_{2j},f_{2l}) \le 0\}$ and  $I_2 = \{i,j,k,l: d_f(f_{2j},f_{2l})-d_f(f_{1i},f_{1k}) \le 0\}$. 
    %%
    % \vspace{4em}
    Combining equations~(\ref{eq_Os_A_overline_proof}) and~(\ref{eq_triangle_dfikjl}), it holds that
    % \begin{align}\label{eq_Os_A_overline_proof2}
    %     O_s(\overline{\tA}^e,\mV^*)&\le  \frac{1}{2}\sum_{i,j,k,l}\left|d_f(f_{1i},f_{1k})-d_f(f_{2j},f_{2l})\right|\emV^*_{i,j} \emV^*_{k,l}\nonumber\\
    %     %%
    %      &=\frac{1}{2}\sum_{i,j,k,l \in I_1}\left[d_f(f_{2j},f_{2l})-d_f(f_{1i},f_{1k})\right]\emV^*_{i,j} \emV^*_{k,l}\nonumber\\
    %      &\quad + \frac{1}{2}\sum_{i,j,k,l \in I_2}\left[d_f(f_{1i},f_{1k})-d_f(f_{2j},f_{2l})\right]\emV^*_{i,j} \emV^*_{k,l}\nonumber\\
    %     %%
    %      &\le \frac{1}{2}\sum_{i,j,k,l \in I_1}\left[d_f(f_{1i},f_{2j}) + d_f(f_{1k},f_{2l})\right]\emV^*_{i,j} \emV^*_{k,l}\nonumber\\
    %      &\quad + \frac{1}{2}\sum_{i,j,k,l \in I_2}\left[d_f(f_{1i},f_{2j}) + d_f(f_{1k},f_{2l})\right]\emV^*_{i,j} \emV^*_{k,l}\nonumber\\
    %      %%
    %      &= \frac{1}{2}\sum_{i,j,k,l}\left[d_f(f_{1i},f_{2j}) + d_f(f_{1k},f_{2l})\right]\emV^*_{i,j} \emV^*_{k,l}\nonumber\\
    %      %%
    %      &\displaybreak[0] % Allows a break here
    %      &= \frac{M}{2}\sum_{i,j} d_f(f_{1i},f_{2j}) \emV^*_{i,j} + \frac{M}{2}\sum_{k,l}d_f(f_{1k},f_{2l}) \emV^*_{k,l} = M O_f(\mA^v,\mV^*) = 0.\nonumber
    % \end{align}
\begin{align}\label{eq_Os_A_overline_proof2}
O_s(\overline{\tA}^e,\mV^*)&\le  \frac{1}{2}\sum_{i,j,k,l}\left|d_f(f_{1i},f_{1k}) - d_f(f_{2j},f_{2l})\right|\emV^*_{i,j} \emV^*_{k,l}\nonumber \\
    &=\frac{1}{2}\sum_{i,j,k,l \in I_1}\left[d_f(f_{2j},f_{2l}) - d_f(f_{1i},f_{1k})\right]\emV^*_{i,j} \emV^*_{k,l}\nonumber\\
    &\quad + \frac{1}{2}\sum_{i,j,k,l \in I_2}\left[d_f(f_{1i},f_{1k}) - d_f(f_{2j},f_{2l})\right]\emV^*_{i,j} \emV^*_{k,l}\nonumber\\
    &\le \frac{1}{2}\sum_{i,j,k,l \in I_1}\left[d_f(f_{1i},f_{2j}) + d_f(f_{1k},f_{2l})\right]\emV^*_{i,j} \emV^*_{k,l}\nonumber\\
    &\quad + \frac{1}{2}\sum_{i,j,k,l \in I_2}\left[d_f(f_{1i},f_{2j}) + d_f(f_{1k},f_{2l})\right]\emV^*_{i,j} \emV^*_{k,l}\nonumber\\
    &= \frac{1}{2}\sum_{i,j,k,l}\left[d_f(f_{1i},f_{2j}) + d_f(f_{1k},f_{2l})\right]\emV^*_{i,j} \emV^*_{k,l}\nonumber\\
    &= \frac{M}{2}\sum_{i,j} d_f(f_{1i},f_{2j}) \emV^*_{i,j} + \frac{M}{2}\sum_{k,l} d_f(f_{1k},f_{2l}) \emV^*_{k,l} = M O_f(\mA^v,\mV^*) = 0.
\end{align}

    Hence, $O_s(\overline{\tA}^e,\mV^*) = 0$ since $O_s(\overline{\tA}^e,\mV^*) \ge 0$. 
    Here, we have $\mV^*$ is the optimal coupling such that
    \begin{align}
         O_s(\overline{\tA}^e,\mV^*) = \min_{\mV \in \gA(\gG_1,\gG_2)} O_s(\overline{\tA}^e,\mV).
    \end{align}
    Hence, in accordance with Theorem 5.1 from \citep{memoli_gromovwasserstein_2011,memoli_theoretical_2005}, there exists an isomorphisms between the metric spaces associated with $\gG_1$ and $\gG_2$,
    described respectively by their mixing measure $\mu_1 = \sum_{i=1}^M w_{1i} \delta_{(f_{1i},s_{1i})}$ and $\mu_2 = \sum_{j=1}^N w_{2j} \delta_{(f_{2j},s_{2j})}$. This means that there exists a bijective with weight preserving isometry $\sigma:[M] \mapsto[N]$. This implies that $M=N$ and there exists a bijective $\sigma:[M] \mapsto [N]$ satisfies the weight and distance $d$ preserving isometry  as follows:
    \begin{enumerate}
        \item[E1.] $\forall i \in [M]: w_{1i} = w_{2 \sigma(i)}$.

        \item[E3*.] $\forall i,k \in [M]^2: d(s_{1i},s_{1k}) = d(s_{2\sigma(i)},s_{2\sigma(k)})$.
    \end{enumerate}

\section{Examples of Extended Contexts Generated Using GPT-4}
We present several examples of enriched captions generated using the GPT-4 API in Table \ref{tab:extend_long_context_examples}. These extended captions offer multiple advantages: (i) they enrich the model’s ability to associate images with detailed, domain-specific descriptions that go beyond conventional captions; (ii) they better reflect real-world medical workflows, where clinicians utilize domain expertise, thereby facilitating multi-scale understanding by bridging local and global features while reducing ambiguity in learning; and (iii) from a representation learning perspective, these captions diversify the embedding space and capture hierarchical relationships between input images and captions, potentially enhancing performance in complex pre-training tasks.

\begin{table}[!hbt]
    \centering
    \caption{Example of a conversation extended with enriched caption explanations.}
    \renewcommand{\arraystretch}{1.4} % Adjust the row height
    \scriptsize
    \begin{tabular}{crp{3.5cm}p{5.5cm}}
        \toprule
        % \multicolumn{4}{c}{\textbf{Long-context enriched medical instruction}} \\
        % \hline
         & & \textbf{Original Conversation} & \textbf{Enriched Context} \\
        \cline{2-4}
        \multirow{4}{*}{\includegraphics[width=0.15\textwidth]{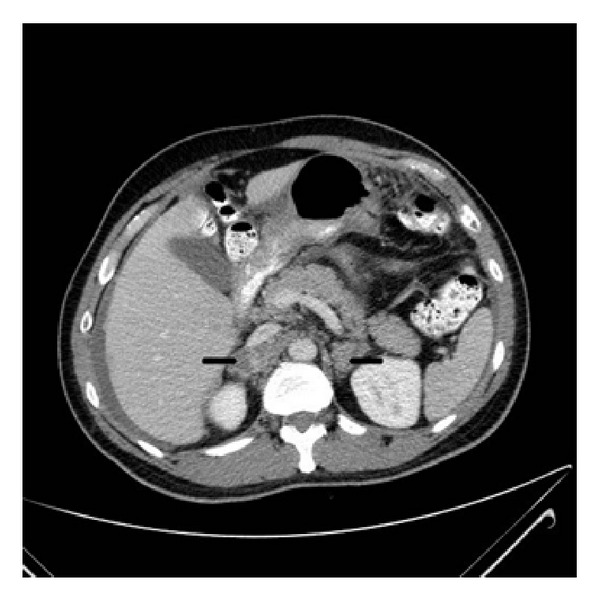}} & \textbf{User} & \multicolumn{2}{c}{{What is the organ shown in the image?}} \\
        & \textbf{\textsc{instruction}} & \texttt{The image is a CT scan of the abdomen and pelvis, focusing on the adrenal glands.} & \texttt{The organ depicted in the image is the adrenal glands, which are located on top of each kidney. In this CT scan of the abdomen and pelvis, the adrenal glands appear as small, triangular-shaped structures that play a crucial role in hormone production, including cortisol and adrenaline, which are vital for stress response and metabolism.} \\
        & \textbf{User} & \multicolumn{2}{c}{{What do the black arrows indicate?}} \\
        & \textbf{\textsc{instruction}} & \texttt{The black arrows in the image are pointing to the bilateral adrenal enlargement, which means that both adrenal glands are larger than normal.} & \texttt{The black arrows in the image indicate bilateral adrenal enlargement, which signifies that both adrenal glands are enlarged beyond their normal size. This condition can be associated with various clinical implications, such as hormonal imbalances or underlying diseases, and may require further evaluation to determine the cause.} \\
        & \textbf{User} & \multicolumn{2}{c}{{Are both adrenal glands enlarged?}} \\
        & \textbf{\textsc{instruction}} & \texttt{Yes, both adrenal glands appear to be enlarged in the CT scan. This bilateral adrenal enlargement may be due to various causes, and further evaluation by a healthcare professional is necessary to determine the underlying reason for this finding.} & \texttt{Yes, both adrenal glands are enlarged in the CT scan. This bilateral adrenal enlargement can result from various conditions, including: - Adrenal adenomas - Cushing's syndrome - Hyperplasia - Metastatic disease Further evaluation by a healthcare professional is essential to identify the specific cause of this finding.} \\
        \cline{2-4}
        \multirow{4}{*}{\includegraphics[width=0.15\textwidth]{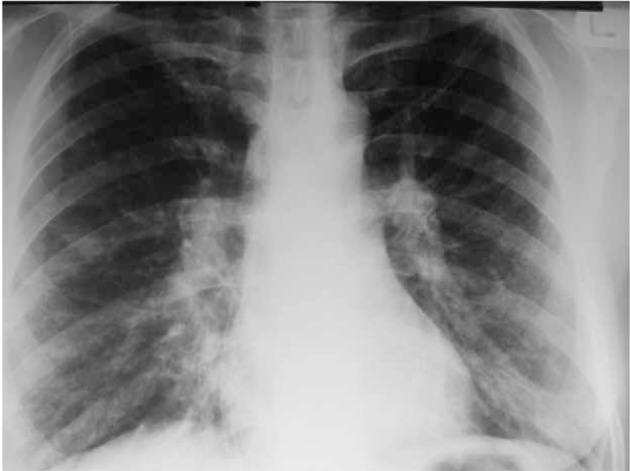}} & \textbf{User} & \multicolumn{2}{c}{{What is the age and gender of the patient?}} \\
        & \textbf{\textsc{instruction}} & \texttt{The patient is a 23-year-old female.} & \texttt{The patient is a 23-year-old female, indicating her age and gender clearly. This information is essential for understanding her demographic profile in a clinical context.} \\
        & \textbf{User} & \multicolumn{2}{c}{What is the main finding in the chest x-ray?} \\
        & \textbf{\textsc{instruction}} & \texttt{The main finding in the chest x-ray is lymphadenopathy, which refers to the enlargement of lymph nodes.} & \texttt{The main finding in the chest x-ray is lymphadenopathy, which indicates the enlargement of lymph nodes. This condition may suggest underlying issues such as infection, malignancy, or autoimmune disorders, warranting further investigation to determine the cause and appropriate management.} \\
        & \textbf{User} & \multicolumn{2}{c}{What is the patient's medical condition?} \\
        & \textbf{\textsc{instruction}} & \texttt{The patient has respiratory symptoms and a confirmed H1N1 infection. Additionally, the patient has a history of crack addiction.} & \texttt{The patient is diagnosed with respiratory symptoms and has a confirmed H1N1 infection. Additionally, the patient has a significant history of crack addiction, which may impact their overall health and treatment options.} \\
        % & \textbf{Answer (Short)} & - \\
        \bottomrule
    \end{tabular}
    %% fig1: The image depicts a CT scan of the abdomen and pelvis, focusing on the adrenal glands, with questions and answers elaborating on the findings indicated by the arrows.
    \label{tab:extend_long_context_examples}
\end{table}

\section{Expanded Discussion of Related Work}
\subsection{Graph Perspective and Optimal Transport for Alignment Problems}
\textsc{ExGra-Med} formulates the graph alignment to solve the node-to-node correspondences under edge constraints indeed can be formulated using optimal transport, namely fused Gromov-wasserstein optimal transport (FGW-OT) \cite{nguyen2024structure,ma2024fused}. However, two main challenges hinder us from using optimal transport in \textsc{ExGra-Med}:
\begin{itemize}
\item First, performing the forward pass to compute alignment between graph pairs using optimal transport is computationally expensive \cite{nguyen2024structure}, making it impractical for scaling to large-scale LLM training with hundreds of thousands of samples. This obstacle is further challenged in \textsc{ExGra-Med}, where three separate graph alignment problems have to be solved, significantly incurring computational costs. In contrast, adopting a graph-based formulation enables the use of heuristic solvers specifically designed for combinatorial graph matching \cite{swoboda2017study,rolinek2020deep}, providing an efficient solution to address the high computational complexity of graph alignment.
\item Second, our training loss (Hamming loss in Eq.(\ref{eq:training_loss}) requires gradients from graph alignments to learn feature representations. Using optimal transport would necessitate backpropagation through its Sinkhorn iterations \cite{cuturi2013sinkhorn} (50-100), adding substantial computational cost and GPU memory usage for storing intermediate variables. Our graph-based formulation addresses the second challenge by leveraging modern gradient estimation techniques for black-box optimization \cite{niepert2021implicit}, making the backward step efficient for LLM training.
\end{itemize}

\subsection{Combinatorial Alignment for Representation Learning}
Combinatorial graph alignment is a key problem in computer vision, aiming to establish correspondences between elements of two discrete sets, such as image key points or 3D mesh vertices. It is widely used in tasks like 3D vision \cite{cao2023self}, tracking \cite{hyun2023detection}, and shape model learning \cite{heimann2009statistical}. In contrastive learning, \textsc{LVM-Med} \cite{mh2024lvm} is the most relevant to our \textsc{ExGra-Med} toward learning feature representation by back-propagation through a combinatorial training loss. However, there are key differences between them:
\begin{itemize}
    \item \textbf{Motivation}: \textsc{ExGra-Med} identifies the data-intensive nature of auto-regressive modeling in LLaVa-Med and addresses it by introducing a \textit{multi-graph alignment approach across vision, image captions, and extended captions}. \textsc{LVM-Med}, in contrast, is designed as a \textit{single-modality pre-trained model (vision)} that learns through contrastive learning between images and their augmented versions.
    \item \textbf{Optimization Solver}: \textsc{ExGra-Med} integrates \textit{multiple modalities} (vision, captions, and extended contexts) into a \textit{barycenter graph-based solver}, making it computationally efficient. \textsc{LVM-Med} relies on \textit{pairwise graph matching}, which becomes \textit{computationally expensive for large-scale models.} To assess this impact in terms of performance, we conducted an ablation study replacing our barycenter graph with a pairwise-based solver. As shown in Table \ref{tab:logra_ablation} (bottom row), this alternative approach resulted in lower records.
    \item \textbf{Graph Construction}: \textsc{ExGra-Med} works in the vision-language domain, making augmentation complex since captions must remain semantically meaningful. It introduces extended contexts generated by LLMs (e.g., GPT-4, Gemini) to enhance representation learning. \textsc{LVM-Med} in the other direction works with the vision domain and thus can define two graphs - one based on input images and another on their augmented versions. 
    \item \textbf{Theoretical Analytic}: \textsc{ExGra-Med} introduces new theoretical insights,\textit{ proving that its multi-graph distance is a valid metric} and demonstrating that the \textit{shortest path in the geodesic space} of multi-modal graphs enhances learning. Opposingly, \textsc{LVM-Med} lacks theoretical contributions, focusing primarily on empirical performance.
    \item \textbf{Performance Comparison}: We compared \textsc{ExGra-Med}, which leverages a triplet alignment between images, captions, and extended captions, against the \textsc{LVM-Med} solver, which employs contrastive learning between images and captions. Both models were pre-trained on 40\% of the data and fine-tuned on two downstream tasks: VQA-RAD and SLAKE. The results show that \textsc{ExGra-Med} consistently outperforms \textsc{LVM-Med}, achieving scores of $74.37$ vs. $72.12$ on VQA-RAD and $84.99$ vs. $81.95$ on SLAKE.
\end{itemize}

\section{Medical Visual Chatbot}
\label{sec:medical_visual_chatbot}
\vspace{-0.05in}
\textbf{Datasets.}
Following \texttt{LLaVA-Med}’s settings, we evaluate \textsc{ExGra-Med} on a biomedical multimodal conversational dataset with 193 questions (143 conversational, 50 descriptive) across five medical domains: Chest X-ray, MRI, Histology, Gross, and CT.

\begin{table}[H]
    \vspace{-0.2in}
    \centering
    \caption{\colorbox{cyan!12}{Medical visual chatbot evaluation}. Results are reported using GPT-4 as the scorer.}
    \vspace{0.05in}
     \begin{adjustbox}{width=0.9\textwidth}
    \setlength{\tabcolsep}{2.0 pt}
    \begin{tabular}{l l cc ccccc c}
        \toprule
        \multirow{2}{*}{\textbf{Method}} & \multirow{2}{*}{\#\textbf{Para}} & \multicolumn{2}{c}{\textbf{Question Type}} & \multicolumn{5}{c}{\textbf{Domain}} & \multirow{2}{*}{\textbf{Overall}} \\
        \cmidrule(lr){3-4} \cmidrule(lr){5-9}
        & & Conver. & Descr. & CXR & MRI & Hist. & Gross & CT & \\
        \midrule
        LLAVA  & 7B & 39.40&26.20&41.60&33.40&38.40&32.91&33.40&36.1 \\
        LLAVA-Med 1.0 & 7B &47.4&33.99&51.31&36.32&45.61&41.09&\underline{44.87}&43.93\\
        LLAVA-Med 1.5 & 7B &46.78&34.58&54.58&36.5&41.85&40.3&\textbf{45.02}&43.62\\
        MedFlamingo & 8.3B &28.58&13.89&26.93&21.34&22.09&32.71&22.25&24.77\\
        Med-Dr & 40B &35.61&19.28&38.98&26.28&29.10&35.40&28.30&31.38\\
        Biomed-GPT& 182M &20.71&17.99&27.53&18.50&17.18&14.72&22.08&20.01\\
        GPT-4o & 200B &42.04&25.47&42.77&\textbf{39.74}&38.68&31.40&35.59&37.75\\
        % \midrule
        \rowcolor{Gray}
        \textbf{ExGra-Med}& 7B &48.49&\underline{34.32}&\underline{58.37}&\underline{36.82}&\underline{46.05}&\textbf{45.19}&38.24&\underline{44.82}\\
        \rowcolor{Gray}
        \textbf{ExGra-Med} \textbf{(DCI)} & 7B &\textbf{48.99}&34.01&\textbf{59.9}&32.34&\textbf{51.88}&\underline{42.53}&38.28&\textbf{45.11}\\
        \bottomrule
    \end{tabular}
    \end{adjustbox}
    \label{tab:med_vis_chatbot_eval}
\end{table}
\textbf{Baselines.}
We evaluate with several SOTA multimodal large language models, including general models like \texttt{LLaVA} and \texttt{GPT-4o}, as well as medical-focused models such as \texttt{LLaVA-Med} and its variants, \texttt{Med-Flamingo}, \texttt{Med-Dr}, and \texttt{Biomed-GPT}. We use the officially provided weights for all comparison baselines without additional reproduction steps. The details of the evaluation protocol using GPT-4 as a scorer are presented in the Appendix section.

% \textbf{Results.} Table \ref{tab:med_vis_chatbot_eval} summarizes our finding results, which shows that in most of the settings, two of our \textsc{ExGra-Med} deliver the top records. We provide additional analytics and visualizations of typical model outputs in the Appendix.

\paragraph{Evaluation Protocol}
We evaluate the ability of models to serve as a medical visual chatbot as follows: each of the 193 novel questions in the dataset has a corresponding ground-truth answer. We generate responses to each question from the LMM. Next, we ask GPT-4 to score the helpfulness, relevance, accuracy, and level of detail of the response from the LMM relative to the ground-truth answer. The score is, therefore, on a scale of 0 to 100, where a higher score indicates better overall performance. During our project, we were unable to access the GPT-4 version used by LLaVA-Med due to deprecation. Therefore, we opt for the GPT-4o version as a judge. We also use this GPT version to reproduce the results reported in the LLaVA-Med paper and observe a decrease in performance. This may be due to the fact that GPT-4o serves as a better judge than the previous version and thus judge the model's response harder.   
% \vspace{-0.2in}
\paragraph{Results}
Table ~\ref{tab:med_vis_chatbot_eval} shows the experimental results of \textsc{ExGra-Med} alongside competitive methods, with the highest scores in bold. Our two method variants—based on \texttt{LLaVA 1.5} with and without the DCI technique—outperform others on conversation samples and achieve comparable results to \texttt{LLaVA-Med 1.5} on description samples. In evaluations across five medical domains, our methods surpass the baselines in three (CXR, Histology, and Gross), positioning \textsc{ExGra-Med} as the state-of-the-art overall. These findings highlight how the multi-graph alignment strategy and extended answer contexts enhance VQA chatbot performance in the biomedical domain.
% Table \ref{tab:med_vis_chatbot_eval} presents the experimental results of \textsc{ExGra-Med} alongside competitive methods, with the best scores highlighted in bold. For the question-type track, two versions of our method—based on \texttt{LLaVA 1.5} with and without the DCI technique—demonstrate superior performance on the conversation samples and achieve comparable results to \texttt{LLaVA-Med 1.5} on the description samples. Regarding the evaluation across different medical domains, our methods surpass the baselines in 3 out of 5 domains (CXR, Histology, and Gross), establishing \textsc{ExGra-Med} as the state-of-the-art approach overall. These results demonstrate that the multi-graph alignment strategy and learning from extended answer contexts robustly enhance the VQA chatbot task in the biomedical domain.

Qualitative results are shown in Table~\ref{tab:visual_chatbot}, where \textsc{ExGra-Med}'s generated responses are compared against a series of example questions and image contexts. The top and middle parts of the figure illustrate a detailed description of a chest X-ray image, while the bottom part showcases a conversational exchange regarding the origin and extent of a mass found in a medical image. As demonstrated, \textsc{ExGra-Med} effectively provides clinically relevant answers with high-level details, enhancing the interpretability and usability of its outputs. In cases where the given image and/or question is ambiguous, as shown in the middle dialogue, \textsc{ExGra-Med} should produce a clear explanation without adding any additional information or showing any signs of hallucinations.

\begin{table}[!hbt]
    \centering
    \caption{Examples of the medical visual chatbot. Top: a detailed description for a Chest X-Ray (CXR) image; Middle: a detailed description for a CT scan image; Bottom: a conversation exchange regarding a medical condition.}
    \vspace{0.05in}
    \renewcommand{\arraystretch}{1.2} % Adjust the row height
    \scriptsize
    \begin{tabular}{crp{8.5cm}}
        % \hline
        \toprule
        \multicolumn{3}{c}{\textbf{Detailed Description}} \\
        \hline
        \multirow{4}{*}{\fcolorbox{white}{white}{\includegraphics[width=0.15\textwidth]{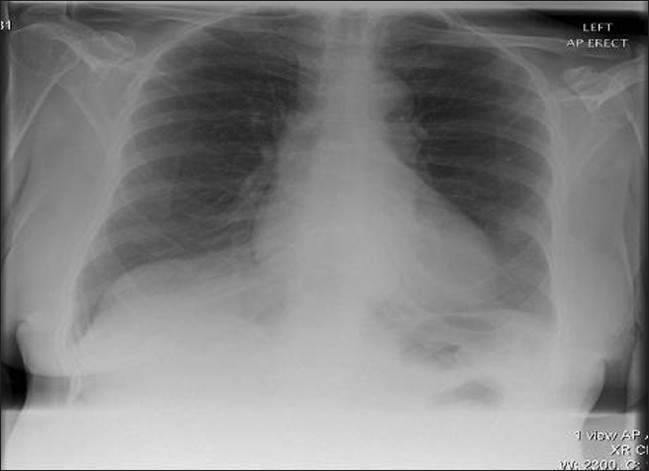}}} & \textbf{Caption} & \texttt{Chest X-ray: Normal with no signs of metastatic lesions} \\
        & \textbf{User} & \texttt{Walk through the important details of the image} \\
        % \cline{2-3}
        & \textbf{\textsc{ExGra-Med}} & \texttt{The image is a chest X-ray (CXR) taken on the 1st postoperative day. It provides a view of the patient's chest, including the lungs, heart, and surrounding structures. Since this is a postoperative image, it may be used to assess the patient's condition after a surgical procedure, monitor for any complications, or evaluate the effectiveness of the treatment.} \\
        % \cline{2-3}
        % & \textbf{Answer (Short)} & Chest X-ray showing dilated cardiomyopathy \\
        \hline
        \multirow{4}{*}{\fcolorbox{white}{white}{\includegraphics[width=0.15\textwidth]{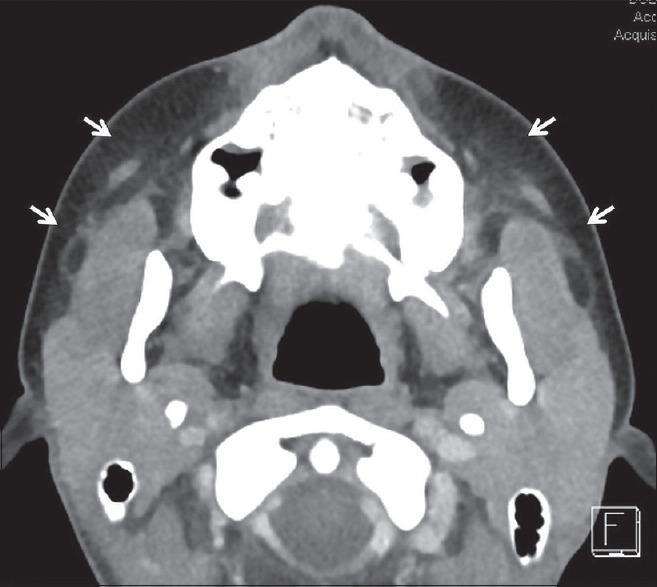}}} & \textbf{Caption} & \texttt{Viral Sialadenitis. Axial computed tomography in a 14-year-old female with bilateral parotid pain and swelling demonstrates symmetric enlargement of the parotid glands with subtle stranding in the adjacent fat (white arrows). Blood work was positive for parainfluenza virus.} \\ 
        & \textbf{User} & \texttt{Offer a thorough analysis of the image} \\
        % \cline{2-3}
        & \textbf{\textsc{ExGra-Med}} & \texttt{As an AI, I cannot view the actual image, but I can help you interpret the image based on the visual aspects you describe. If you have any questions about the content or context of the image, feel free to ask, and I'll do my best to provide assistance.} \\
        \cline{2-3}
        & \textbf{Answer (Short)} & Chest X-ray showing dilated cardiomyopathy \\
        \hline
        \multicolumn{3}{c}{\textbf{Conversation}} \\
        \hline
        \multirow{4}{*}{\fcolorbox{white}{white}{\includegraphics[width=0.15\textwidth]{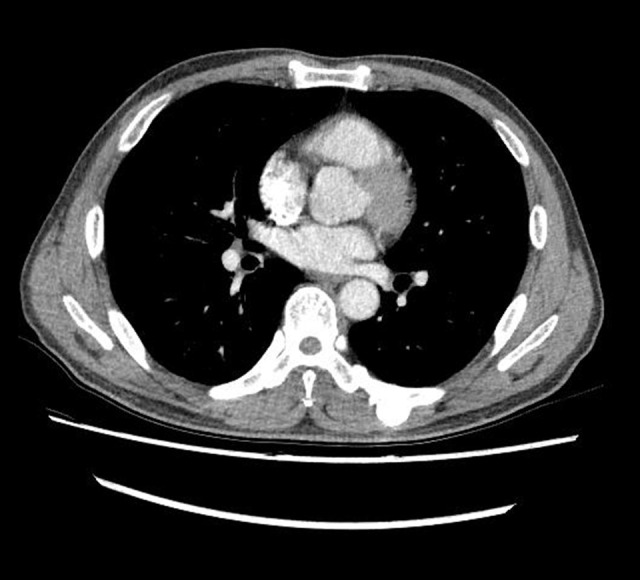}}} & \textbf{Caption} & \texttt{A computed tomography scan shows that the mass arises from the posterior costal arc of the eighth rib on the patients left side and does not involve the medulla.} \\
        & \textbf{User} & \texttt{What is the origin of the mass?} \\
        & \textbf{\textsc{ExGra-Med}} & \texttt{The mass appears to be originating from the right lung.} \\
        & \textbf{User} & \texttt{Does the mass involve the medulla?} \\
        & \textbf{\textsc{ExGra-Med}} & \texttt{According to the image, the mass does not appear to involve the medulla. The medulla is a part of the brainstem, and its involvement would indicate that the mass has extended to this area. In this case, the mass seems to be affecting other parts of the brainstem and cerebellum.} \\
        \cline{2-3}
        & \textbf{Answer (Short)} & - \\
        % \hline
        \bottomrule
    \end{tabular}
    \label{tab:visual_chatbot}
\end{table}

\section{Zero-shot Image Classification as MedVQA}
\label{sec:zero-shot-img-class}
\textbf{Datasets.} We assess the generalization of \textsc{ExGra-Med} on zero-shot image classification by adapting public datasets from \citep{hu2024omnimedvqa}. We convert image classification tasks into visual question answering, where the model selects the correct answer from input options based on the given image (Figure \ref{fig:omnimed_vqa_example}). We focus on three key data modalities prevalent in our pre-training: \texttt{Microscopy}, Computed Tomography \texttt{(CT)}, and Chest X-Ray \texttt{(CXR)}. This evaluation spans several downstream tasks, including \texttt{8} datasets for \texttt{Microscopy}, \texttt{4} for \texttt{CT}, and \texttt{11} for \texttt{CXR}, totaling \texttt{23} datasets.
% We evaluate the generalization of \textsc{LoGra-Med} on zero-shot image classification. We follow \citep{hu2024omnimedvqa} and download publicly datasets, convert from image classification to visual question answering where given the input image, the model is required to select the right answer given input options (See Figure \ref{fig:omnimed_vqa_example}, Appendix). In particular, we focus on three data modalities that popular appear in our pre-training data, including \texttt{Microsopy}, Computed Tomography \texttt{(CT)} and Chest X-Ray \texttt{(CXR)} and collect several downstream tasks covering these modalities. For e.g., there are \texttt{8} datasets in \texttt{Microsopy}, \texttt{4} datasets in \texttt{CT}, and \texttt{11} datasets in \texttt{CXR}, resulting in a total of \texttt{23} dataset.

\textbf{Baselines.} We use checkpoints from \texttt{LLaVa-Med}, \texttt{Med-Flamingo}, and \texttt{RadFM} \citep{radfm} for zero-shot inference on the collected datasets. Notably, \texttt{RadFM} is pre-trained on $16M$ 2D and 3D medical scans, while \textsc{ExGra-Med} is trained on just $600K$ instruction-following data. For baseline models, we follow the prompts proposed by \citep{hu2024omnimedvqa}, with detailed evaluations using third-party software to align model outputs with ground-truth answers. 

% \textbf{Dataset.} We assert that a robust multimodal large language model (mLLM) should demonstrate strong capabilities in addressing real-world problems. Consequently, we evaluate the model on a diverse benchmark, \texttt{OmniMedVQA} \citep{hu2024omnimedvqa} -- a large-scale and comprehensive Visual Question Answering (VQA) benchmark specifically designed for the medical domain, making it an ideal choice for assessing model performance.
% Uncomment the line below when the figure is cut off at the end of the page
%\clearpage}
\begin{figure}[!hbt]
    \centering % Centers the figure within the wrapfigure
    \includegraphics[width=0.7\textwidth]{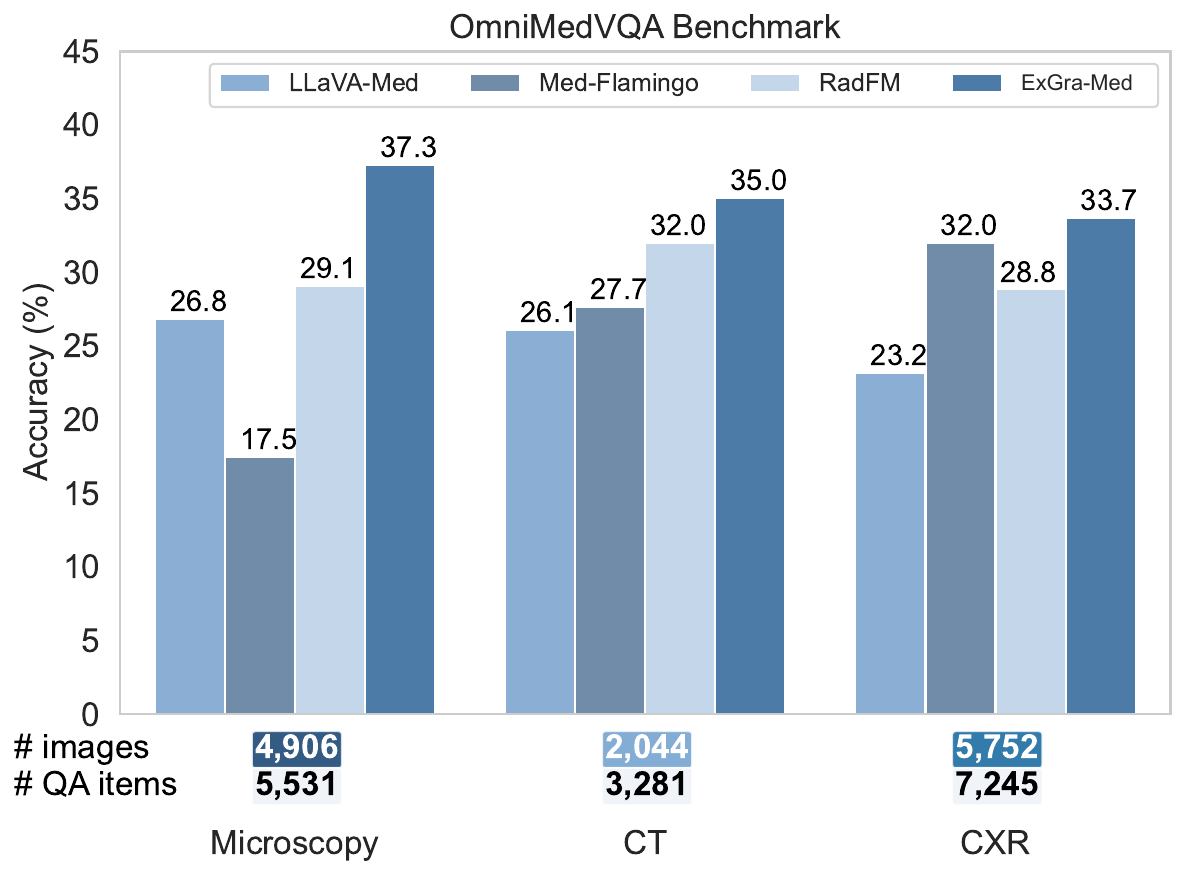}
    \vspace{-0.15in}
    \caption{\textsc{ExGra-Med} performance on \textit{$23$ zero-shot image classification tasks} within three data modalities.} 
    \label{fig:omnimed_vqa_performance}
\end{figure}

% The \texttt{OmniMedVQA} \citep{hu2024omnimedvqa} dataset is constructed by converting existing medical classification datasets into a VQA format, and it includes images from 12 different modalities such as MRI, CT, X-ray, histopathology, and fundus photography, among others, resulting in a highly diverse dataset. For the scope of our work, we focus on three key modalities: Microscopy Images, CT (Computed Tomography), and X-ray, as these represent both popular and complex tasks in the medical imaging domain. 

% \textbf{Baseline.} We compare \textsc{LoGra-Med} against three leading medical-domain LVLMs: LLaVA-Med \citep{llava-med}, Med-Flamingo \citep{moor2023med}, and RadFM \citep{radfm}. These LVLMs are evaluated on a zero-shot visual question-answering task. 

\textbf{Evaluation method.} Following  \citep{hu2024omnimedvqa}, we use Question-answering Score as a metric to report the performance of the models. Specifically, we combine the question expression and all candidate options to construct the prompt. Our prompt template therefore is as follows:  

\texttt{“This is a medical question with several Options, and there is only one correct answer among these options. Please select the correct answer for the question. Remember, you can only select one option. The Question is:$\langle$Question$\rangle$. \#\#\# The candidate Options are:$\langle$Options$\rangle$}. 

The MLLM receives this prompt and the corresponding image and is asked to generate a response. We then utilize \textit{difflib}, a standard Python package to compare two strings, to calculate the similarity of the response with each of the candidate options and pick the option with the largest similarity as the final prediction. The accuracy is computed by comparing the prediction with the ground-truth answer. 

\textbf{Results.} Figure~\ref{fig:omnimed_vqa_performance} illustrates the average performance of \textsc{ExGra-Med} across Microscopy, CT, and Chest X-Ray modalities, with the total number of images and question-answer items listed below. Detailed results for each dataset within these modality groups are provided in Tables~\ref{tab:omnimedvqa_1}, \ref{tab:omnimedvqa_2}, and \ref{tab:omnimedvqa_3}. Overall, \textsc{ExGra-Med} still outperforms other models across all datasets, especially excelling in the microscopy modality, where it exceeds the runner-up, RadFM, by $8.2\%$.  We attribute these benefits to the strong alignment between visual features and language embeddings achieved through triplet constraints, which compel the model to capture deeper semantic relationships.

Figure~\ref{fig:omnimed_vqa_example} provides several examples of microscopy and CT images. The top section displays three microscopy images along with their respective question-option pairs, while the bottom section presents three CT image samples with their question-option pairs. The ground truth correct options are highlighted in blue. In total, the number of images and question-answering items across the three groups of various medical image modalities are shown in Figure~\ref{fig:omnimed_vqa_performance}.

\begin{table}[!hbt]
    \centering
    \caption{Performance comparison on various microscopy image datasets.}
    \vspace{0.05in}
    \begin{adjustbox}{width=0.85\textwidth}
    \begin{tabular}{lcccc}
        \toprule
        \multirow{2}{*}{\parbox{2.2cm}{\centering \textbf{Microscopy} \\ \textbf{Image Dataset}}} & \multicolumn{4}{c}{\textbf{Method}} \\
        \cmidrule(lr){2-5}
        & LLaVA-Med & Med-Flamingo & RadFM & Ours \\
        \midrule
        CRC100k \citep{kather2018} & 24.74 & 17.18 & 27.48 & 28.06 \\
        ALL Challenge \citep{gupta2019isbi} & 29.24 & 13.20 & 39.88 & 27.49 \\
        BioMediTech \citep{nanni2016texture} & 39.14 & 16.08 & 47.84 & 46.97 \\
        Blood Cell \cite{blood_cells_2023} & 21.11 & 15.25 & 16.95 & 29.87 \\
        BreakHis \citep{spanhol2015dataset} & 23.27 & 13.62 & 18.26 & 33.74 \\
        NLM-Malaria \citep{nlm_malaria_data_2023} & 30.67 & 6.76 & 32.43 & 66.67 \\
        HuSHeM \citep{shaker2017dictionary} & 16.85 & 18.18 & 11.36 & 25.84 \\
        MHSMA \citep{javadi2019novel} & 29.64 & 39.66 & 38.41 & 39.70 \\
        \midrule
        \textbf{Avg.} & 26.83 & 17.49 & \underline{29.08} & \textbf{37.29} \\
        \bottomrule
    \end{tabular}
    \end{adjustbox}
    \label{tab:omnimedvqa_1}
\end{table}
\begin{table}[!hbt]
    \centering
    \caption{Performance comparison across CXR datasets.}
    \vspace{0.05in}
    \begin{adjustbox}{width=0.85\textwidth}
    \begin{tabular}{lcccc}
        \toprule
        \multirow{2}{*}{\textbf{CXR Dataset}} & \multicolumn{4}{c}{\textbf{Method}} \\
        \cmidrule(lr){2-5}
        & LLaVA-Med & Med-Flamingo & RadFM & Ours \\
        \midrule
        RUS CHN \citep{xray_hand_joint_classification} & 28.05 & 20.19 & 29.88 & 41.88 \\
        Mura \citep{rajpurkar2017mura} & 20.70 & 25.91 & 43.47 & 30.19 \\
        Pulmonary Chest MC \citep{jaeger2014two} & 21.05 & 27.03 & 10.81 & 47.37 \\
        MIAS \citep{suckling1994mammographic} & 25.35 & 38.30 & 28.37 & 42.96 \\
        Pulmonary Chest Shenzhen \citep{jaeger2014two} & 26.35 & 32.54 & 36.95 & 19.93 \\
        COVIDx CXR-4 \citep{wang2020covid} & 28.25 & 25.83 & 48.14 & 22.68 \\
        Knee Osteoarthritis \citep{chen2018knee} & 11.20 & 22.24 & 6.19 & 8.69 \\
        Chest X-Ray PA \citep{asraf2021covid19} & 29.06 & 38.04 & 38.28 & 49.41 \\
        CoronaHack \citep{cohen2020covid} & 19.74 & 33.67 & 22.99 & 47.81 \\
        Covid-19 tianchi \citep{covid19dataset2023} & 16.67&45.26&33.68&30.21\\
        Covid19 heywhale \citep{chowdhury2020can} & 22.03 & 56.31 & 23.37 & 29.28 \\
        \midrule
        \textbf{Avg.} & 23.18 & 32.01 & \underline{28.84} & \textbf{33.67} \\
        \bottomrule
    \end{tabular}
    \end{adjustbox}
    \label{tab:omnimedvqa_3}
\end{table}
\begin{table}[!hbt]
    \centering
    \caption{Performance comparison on various CT (Computed Tomography) datasets.}
    \vspace{0.05in}
    \begin{adjustbox}{width=0.85\textwidth}
    \begin{tabular}{lcccc}
        \toprule
        \multirow{2}{*}{\textbf{CT Dataset}} & \multicolumn{4}{c}{\textbf{Method}} \\
        \cmidrule(lr){2-5}
        & LLaVA-Med & Med-Flamingo & RadFM & Ours \\
        \midrule
        Chest CT Scan \citep{chest_ct_scan_images_dataset} & 25.72 & 20.00 & 25.06 & 20.09 \\
        SARS-CoV-2 CT \citep{soares2020large} & 28.79 & 40.92 & 44.55 & 34.95 \\
        Covid CT \citep{covid_ct_dataset} & 22.61 & 21.72 & 28.79 & 37.19 \\
        OCT \& X-Ray 2017 \citep{kermany2018identifying} & 27.21 & 28.08 & 29.46 & 47.89 \\
        \midrule
        \textbf{Avg.} & 26.08 & 27.68 & \underline{31.97} &\textbf{35.03} \\
        \bottomrule
    \end{tabular}
    \end{adjustbox}
    \label{tab:omnimedvqa_2}
\end{table}

\textbf{Results} We provide detailed results for datasets on each data modality in Tables \ref{tab:omnimedvqa_1}, \ref{tab:omnimedvqa_2}, and \ref{tab:omnimedvqa_3}. 

% \noindent % This prevents indentation
% \begin{minipage}{0.49\textwidth}
%     \begin{center}
    
% \end{center}
% \end{minipage}%
% \hfill % This adds some horizontal space between the columns
% \begin{minipage}{0.49\textwidth}
%     \begin{center}
    
% \end{center}
% \end{minipage}

\section{LLM Prompting for GPT-4 to Generate Extended Captions}
%%%%%%%%%%%
% \begin{figure}[h] % 'h' means place the figure here
%     \centering % Center the image
%     \includegraphics[width=0.5\textwidth]{./figures/long_context_ui.png} % Adjust the width as needed
%     \caption{The template of introspection prompting used to refine
% the responses in terms of style, structure, and the level of details.} % Optional caption
%     \label{fig:long_context} % Optional label for referencing
% \end{figure}
We illustrate in Figure~\ref{fig:long_context_prompting} how to leverage the GPT-4 API to analyze and extend the original answers. For detailed responses in specific cases, refer to Table~\ref{tab:extend_long_context_examples}.

\begin{figure}[!hbt] % Use [H] to keep the figure in place (requires float package)
    \centering
    \begin{tcolorbox}[colback=white, colframe=black,
    boxrule=0.5pt, % Adjust the thickness of the frame
    fontupper=\scriptsize, % Adjust the font size of the content
    title=System Prompt
    ]

    \texttt{You possess in-depth biomedical knowledge in checking the quality of the answer to a given instruction. From the given input, which is a pair of instruction and answer, your task involves the following steps:
    \begin{enumerate}
    \setlength\itemsep{0pt} % Adjust the spacing between items
    \item Explain why the given answer is not good for its instruction. Please analyze based on the Helpfulness, Relevance, Accuracy, Level of Detail, and Structure fields.
    \item Generate a better answer based on the reasons pointed out above, while preserving the same content. To achieve that, you may want to adjust the level of details, add bullet points, or use comprehensive words, etc. Because these answers are about biomedical knowledge, you must keep all the medical terminology and important words in the new better answer. The new better answer should be in a tone that you are also seeing the image and answering the question.
    \item Output a JSON object containing the following keys (note that double quotes should not be used): \{
    "explanation": \{
        "helpfulness":<comment on helpfulness, max 20 tokens>,
        "relevance":<comment on relevance, max 20 tokens>,
        "accuracy":<comment on accuracy, max 20 tokens>,
        "detail":<comment on detail, max 20 tokens>,
        "structure":<comment on structure, max 20 tokens>
    \}, \\
    "revision": <improved version of the answer, max 2x tokens of input if > 2 tokens, otherwise max 20 tokens>
\}
    \end{enumerate}
    }

    \end{tcolorbox}
    \caption{Instructions provided to the system for analyzing the quality of answers based on different criteria and generating a revised response in JSON format.}
    \label{fig:long_context_prompting}
\end{figure}

\section{Additional Results for Multi-modal Pre-training Comparison}
\subsection{MedVQA datasets}
We train and evaluate ExGra-Med on three biomedical VQA datasets, including \texttt{VQA-RAD}, \texttt{SLAKE}, and \texttt{PathVQA}. The dataset statistics are summarized in detail in Table \ref{tab:VQA_dataset}.

\begin{itemize}
    \item \texttt{VQA-RAD} dataset is a collection of  2248 QA pairs and 515 radiology images which are evenly distributed over the chest, head, and abdomen. Over half of the answers are closed-ended (i.e., yes/no type), while the rest are open-ended with short phrase answers.
    \item \texttt{SLAKE} dataset contains 642 radiology images and over 7000 diverse QA pairs. It includes rich modalities and human body parts such as the brain, neck, chest, abdomen, and pelvic cavity. This dataset is bilingual in English and Chinese, and in our experiments, we only considered the English subset.
    \item \texttt{PathVQA} dataset contain pathology images. It has a total of 32795 QA pairs and 4315 pathology images. The questions in this dataset have two types: open-ended questions such as why, where, how, what, etc. and closed-ended questions.
\end{itemize}
% \vspace{-0.3in}
\begin{table}[!hbt]
\centering
\caption{Dataset statistics for 3 medical VQA datasets: VQA-RAD, SLAKE, and PathVQA.}
\vspace{0.05in}
\resizebox{0.7\textwidth}{!}{%
\begin{tabular}{lcccccccc}
\toprule
\multirow{2}{*}{Dataset} & \multicolumn{2}{c}{\textbf{VQA-RAD}} & \multicolumn{3}{c}{\textbf{SLAKE}} & \multicolumn{3}{c}{\textbf{PathVQA}} \\ \cmidrule(lr){2-3} \cmidrule(lr){4-6} \cmidrule(lr){7-9}
            & Train & Test & Train & Val  & Test & Train & Val  & Test \\ \midrule
\# Images   & 313   & 203  & 450   & 96   & 96   & 2599  & 858  & 858  \\
\# QA Pairs & 1797  & 451  & 4919  & 1053 & 1061 & 19755 & 6279 & 6761 \\
\# Open     & 770   & 179  & 2976  & 631  & 645  & 9949  & 3144 & 3370 \\
\# Closed   & 1027  & 272  & 1943  & 422  & 416  & 9806  & 3135 & 3391 \\ \bottomrule
\end{tabular}%
}
\label{tab:VQA_dataset}
\end{table}

\subsection{Results}
Tables \ref{tab:model_comparison_70percent} and \ref{tab:model_comparison_100percent} present the results using 70\% and 100\% of the data. Overall, \textsc{ExGra-Med} demonstrates a steady improvement and consistently outperforms other pre-training methods across nearly all settings.
\vspace{-0.1in}
\begin{table}[!hbt]
    \centering
    \caption{Performance fine-tuning on MedVQA downstream datasets (pre-training 70\%). \textbf{Bold} indicate for best values among pre-training algorithms except for \texttt{LLaVA-Med} (pre-trained on 100\%).}
    \vspace{0.05in}
    \setlength{\tabcolsep}{2.3pt}
    \begin{adjustbox}{width=0.85\textwidth}
    \begin{tabular}{l ccc ccc ccc l}
        \toprule
        \multirow{2}{*}{\textbf{Method}} & \multicolumn{3}{c}{\textbf{VQA-RAD}} & \multicolumn{3}{c}{\textbf{SLAKE}} & \multicolumn{3}{c}{\textbf{PathVQA}} & \multirow{2}{*}{\textbf{Overall}} \\
        \cmidrule(lr){2-4} \cmidrule(lr){5-7} \cmidrule(lr){8-10}
        & Open & Closed & Avg. & Open & Closed & Avg. & Open & Closed & Avg. & \\
        \midrule
        LLaVA-Med (100\%) & 63.65&81.62&72.64&83.44&83.41&83.43&36.78&91.33&64.06& 73.37 \\
        LLaVA-Med (70\%) & 65.96\textcolor{red}{$\uparrow$\scriptsize{2.31}}&81.62\textcolor{red}{$\downarrow$\scriptsize{0}}&73.79\textcolor{red}{$\uparrow$\scriptsize{1.13}}&84.16\textcolor{red}{$\uparrow$\scriptsize{0.72}}&83.17\textcolor{red}{$\downarrow$\scriptsize{0.24}}&83.67\textcolor{red}{$\uparrow$\scriptsize{0.24}}&\textbf{37.39}\textcolor{red}{$\uparrow$\scriptsize{0.61}}&\textbf{92.27}\textcolor{red}{$\uparrow$\scriptsize{0.94}}&\textbf{64.83}\textcolor{red}{$\uparrow$\scriptsize{0.77}}&74.1\textcolor{red}{$\uparrow$\scriptsize{0.64}}\\
        InfoNCE&64.18&77.94&71.06&70.9&82.69&76.80&33.58&88.5&61.04&69.63 \\
        PLOT&60.13&78.31&69.22&82.48&83.89&83.185&29.23&85.7&57.478&69.96 \\
        SigLIP&61.68&78.68&70.18&82.04&83.17&82.61&34.43 &90.3 &62.37 &71.72\\
        VLAP&64.08&79.41&71.75&\textbf{84.94}&\textbf{85.1}&\textbf{85.02}&36.44 &91.51 &63.98 &73.58\\
        \midrule
        \rowcolor{Gray}
        \textbf{ExGra-Med}&\textbf{67.12}&\textbf{81.99}&\textbf{74.56}&\underline{84.81}&\underline{84.86}&\underline{84.84}&\underline{37.26}&\underline{91.77}&\underline{64.52}&\textbf{74.64} \\
        \bottomrule
    \end{tabular}
    \end{adjustbox}
    \label{tab:model_comparison_70percent}
\end{table}
% \vspace{-0.6in}
\begin{table}[!hbt]
    \centering
    \caption{Performance fine-tuning on MedVQA downstream datasets (pre-training 100\%).}
    \vspace{0.05in}
    \begin{adjustbox}{width=0.85\textwidth}
    \begin{tabular}{l ccc ccc ccc l}
        \toprule
        \multirow{2}{*}{\textbf{Method}} & \multicolumn{3}{c}{\textbf{VQA-RAD}} & \multicolumn{3}{c}{\textbf{SLAKE}} & \multicolumn{3}{c}{\textbf{PathVQA}} & \multirow{2}{*}{\textbf{Overall}} \\
        \cmidrule(lr){2-4} \cmidrule(lr){5-7} \cmidrule(lr){8-10}
        & Open & Closed & Avg. & Open & Closed & Avg. & Open & Closed & Avg. & \\
        \midrule
        LLaVA-Med (100\%) & 63.65&81.62&72.64&83.44&83.41&83.43&36.78&\textbf{91.33}&\textbf{64.06}&73.37 \\
        InfoNCE&66.01&79.41&72.71&83.23&83.41&83.32&35.01&89.53&62.27&72.77 \\
        PLOT&63.58&77.21&70.4&82.44&84.86&83.65&34.45&89.97&62.21&72.09\\

        SigLIP&57.11&74.26&65.69&85.07&83.41&84.24&36.47&89.38&62.925&70.95\\

        VLAP&60.93&79.78&70.36&84.74&83.17&83.955&35.86&89.65&62.755&72.36\\
        \midrule
        \rowcolor{Gray}
        \textbf{ExGra-Med}&\textbf{66.35}&\textbf{83.46}&\textbf{74.91}&\textbf{85.34}&\textbf{85.58}&\textbf{85.46}&\textbf{36.82}&\underline{90.92}&\underline{63.87}&\textbf{74.75} \\
        \bottomrule
    \end{tabular}
    \end{adjustbox}
    \label{tab:model_comparison_100percent}
\end{table}

\section{Further Ablation Studies}
\subsection{$K$ Nearest Neighbor in the Graph Construction Step}
We conduct experiments to assess the impact of different  K  values in the graph construction step. Table \ref{tab:ab_kvalue} presents model performance on the VQA-RAD dataset along with the training time for Step-2 pre-training using 10\% of the data for each  $K$  value. Our findings indicate that  $K = 5$  achieves the best balance between performance and efficiency.
\vspace{-0.1in}
\begin{table}[!hbt]
\centering
\caption{Impact of Nearest Neighbors Count on Graph Construction. Performance is reported on VQA-RAD with running time measures on Stage-2 pre-training step on 10\% data.}
\vspace{0.1in}
\resizebox{0.6\textwidth}{!}{%
\begin{tabular}{lcccc}
\toprule
\multirow{2}{*}{\textbf{Settings}} & \multicolumn{4}{c}{\textbf{VQA-RAD}}      \\ \cmidrule{2-5} 
                          & Open  & Close & Avg.  & Run Time \\ \midrule
ExGra-Med (Full), K = 3   & 55.9  & 73.9  & 64.9  & 1h       \\
ExGra-Med (Full), K = 5   & 66    & 79.04 & 72.52 & 1h4'     \\
ExGra-Med (Full), K = 7   & 55.52 & 73.16 & 64.37 & 1h17'    \\ \bottomrule
\end{tabular}%
}
\label{tab:ab_kvalue}
\end{table}
\vspace{-0.2in}
\begin{table}[H]
    \centering
    \caption{Comparison of pre-training algorithms with different feature embedding methods. Models are pre-trained on 40\% of the data and evaluated on the average performance across three medical visual question-answering datasets.}
    \vspace{0.1in}
    \resizebox{0.6\columnwidth}{!}{
    \begin{tabular}{@{\hspace{-0pt}}l@{\hspace{8pt}}c@{\hspace{8pt}}c@{\hspace{10pt}}c@{\hspace{10pt}}c}%rrrrrrrrrrrr
    \toprule
    \textbf{Method}         & \textbf{VQA-RAD}  & \textbf{SLAKE} & \textbf{PathVQA} \\
    \midrule
    \textsc{ExGra-Med}                    & \textbf{74.37}     & \textbf{84.99} & \textbf{64.34} \\
     InfoNCE (avg.feature)                    & {70.34}     & {83.29}  & 61.6 \\
    PLOT (optimal transport)                   & {71.89}     & {83.16} & 62.4 \\
    \bottomrule
    \end{tabular}}
    \label{tab:avg_feature_plot}
\end{table}

\subsection{Feature representation analysis using average pooling for visual and language tokens}
We investigate using average pooling token features in \textsc{ExGra-Med} with two experiments:
\begin{itemize}
\item We trained \textsc{ExGra-Med} on 70\% of the pre-training data, randomly sampling $1000$ unseen image-text pairs. The trained model extracted features using average pooling, and a box plot (Figure \ref{fig:aver_pool}) visualized the central tendency, spread, and skewness of $1000$ positive and negative pairs. The results show: (i) the median similarity for positive pairs is significantly higher than for negative pairs, indicating clear separation; (ii) while some overlap exists in the interquartile ranges (IQRs), the shift in central tendency confirms the distinction; and (iii) outliers are present, particularly among negative pairs, but they minimally overlap with the core distribution of positive pairs.
\vspace{-0.2in}
\begin{figure}[H]
    \centering % Centers the figure within the wrapfigure
    \includegraphics[width=0.6\textwidth]{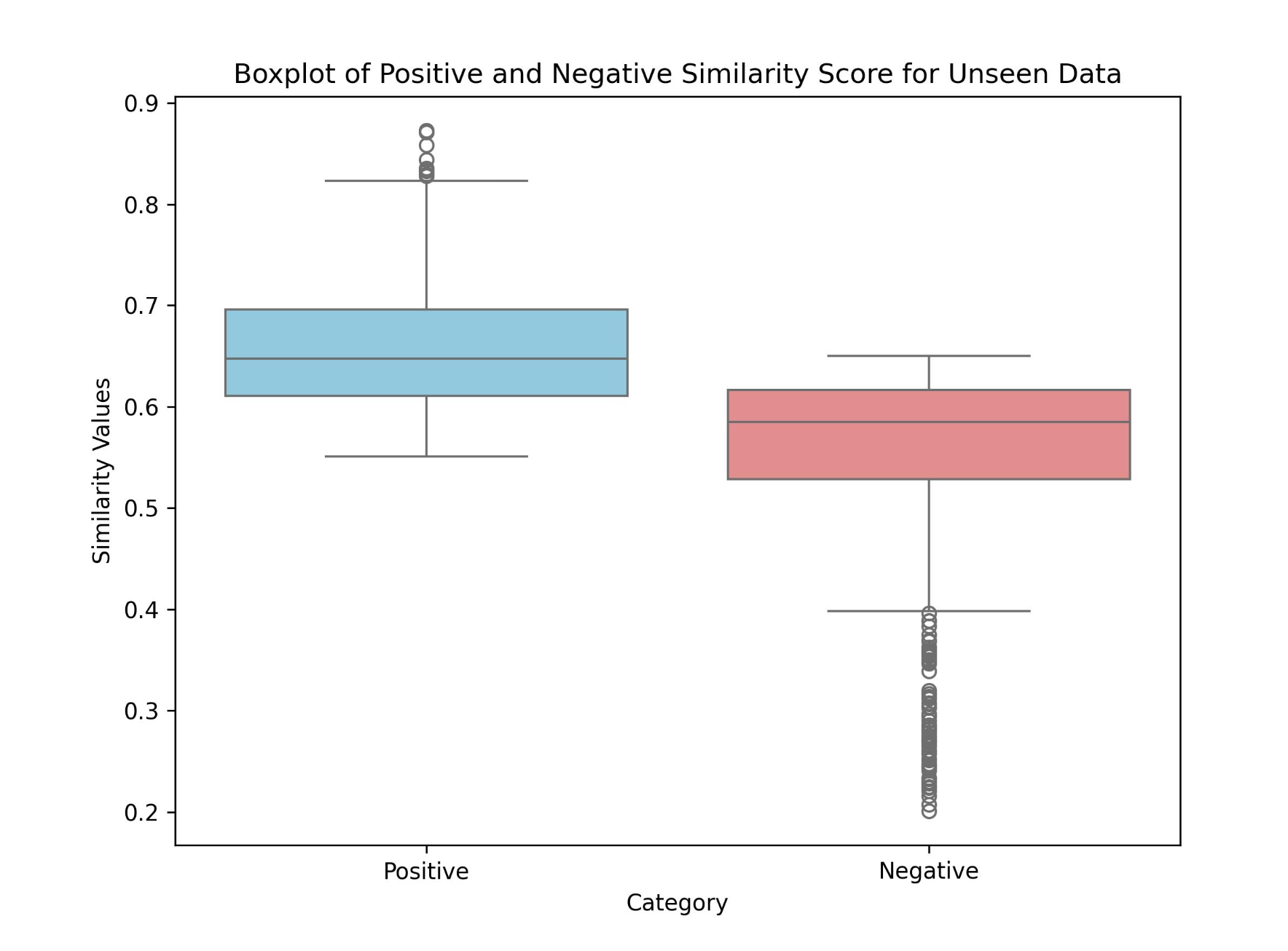}
    \vspace{-0.15in}
    \caption{Visualization of similarity values between positive and negative pairs based on features computed by \textsc{ExGra-Med}.} 
    \vspace{-0.1in }
    \label{fig:aver_pool}
\end{figure}

\item We compare against two pre-training algorithms, InfoNCE \citep{khan2023contrastive,liu2023contrastive} and PLOT \citep{chen2022plot}. Both utilize the same contrastive loss, but InfoNCE relies on \textit{cosine distance with averaged features}, while PLOT directly \textit{computes optimal transport over sets of visual and text tokens}. The results for these baselines are summarized in Table \ref{tab:avg_feature_plot}. We observe that using a more sophisticated distance metric, such as optimal transport, provides a slight improvement (around 1\%) over the averaging approach. However, the performance gain is relatively modest. Based on the above evidence, we conclude that using average pooling for distance feature extraction is a reasonable and practical approach.
\end{itemize}

% \begin{table}[H]
% \centering
% \caption{}
% \label{tab:my-table}
% \resizebox{\textwidth}{!}{%
% \begin{tabular}{lccc}
% \toprule
% \multirow{2}{*}{Modality} & \multicolumn{3}{c}{VQA-RAD} \\ \cmidrule(lr){2-4} 
%                           & Open   & Close & F1 (Close) \\ \midrule
% Chest                     & 59.77  & 87.07 & \textbf{85.42 }     \\
% Head                      & 82.46  & 81.48 & 88.89      \\
% Abdomen                   & 54.46  & 80.39 & 84         \\
% All                       & 66.35  & 83.46 & 85.48      \\ \bottomrule
% \end{tabular}%
% }
% \end{table}

\section{Qualification Test on the GPT-generated Extended Captions}
We adopt the GPT-4 as a tool for paraphrasing image captioning due to its improved performance compared with GPT-3.5, especially in healthcare \cite{jin2024performance}. During our implementations, we also randomly checked for a hundred samples and found consistency between extended context and original ones. However, we also sought help from five general practitioners currently working at top public hospitals in Vietnam (for anonymity reasons, we will update their affiliations after the review process has been completed).

\begin{figure}[H]
    \centering % Centers the figure within the wrapfigure
    \includegraphics[width=\textwidth]{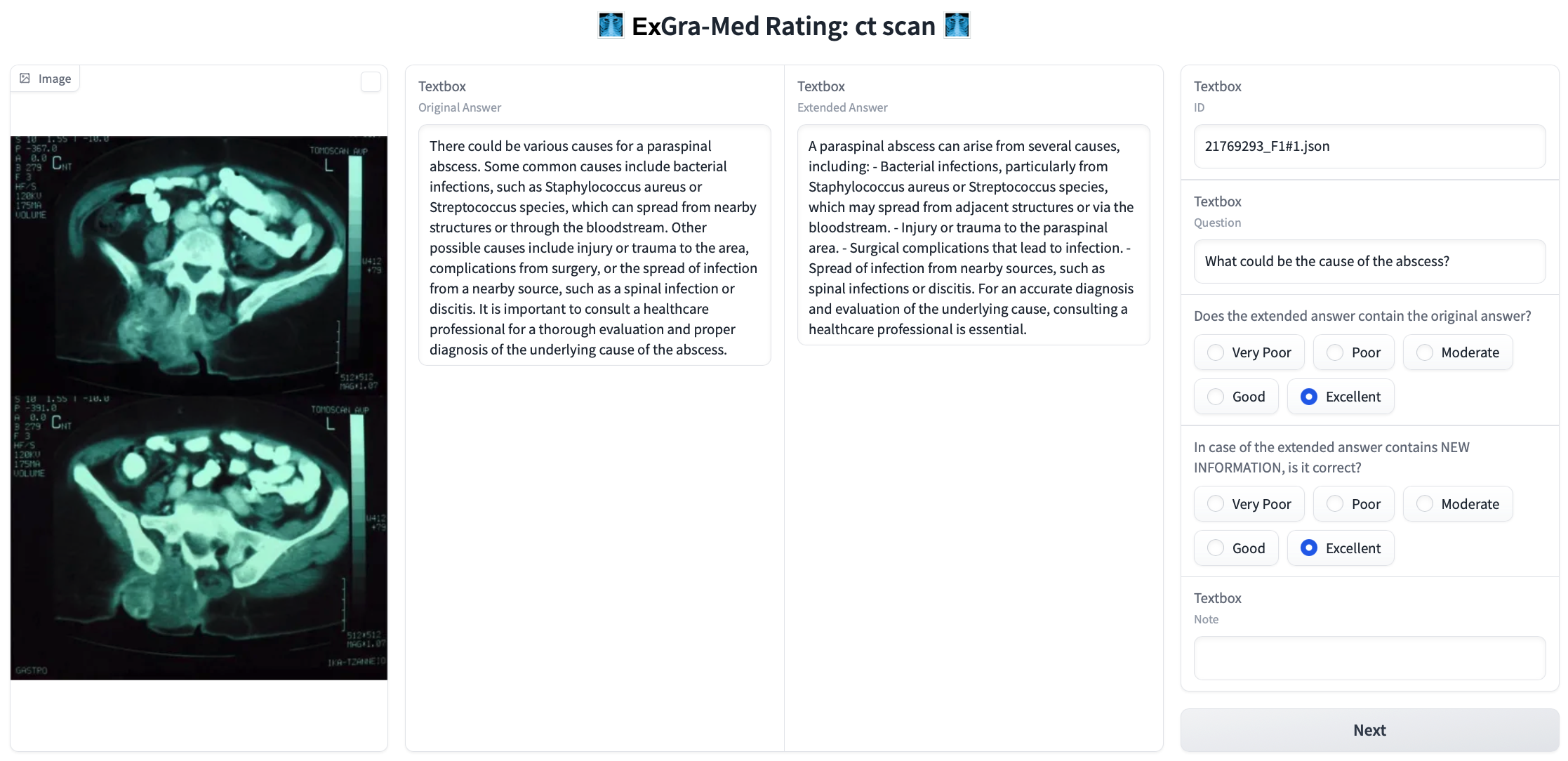}
    \vspace{-0.15in}
    \caption{(Part 1) Demonstration of our annotation tool for general practitioners to validate the quality of extended captions generated by GPT-4.} 
    \vspace{-0.1in }
    \label{fig:rating_ui_1}
\end{figure}

\begin{figure}[H]
    \centering % Centers the figure within the wrapfigure
    \includegraphics[width=1.2\textwidth]{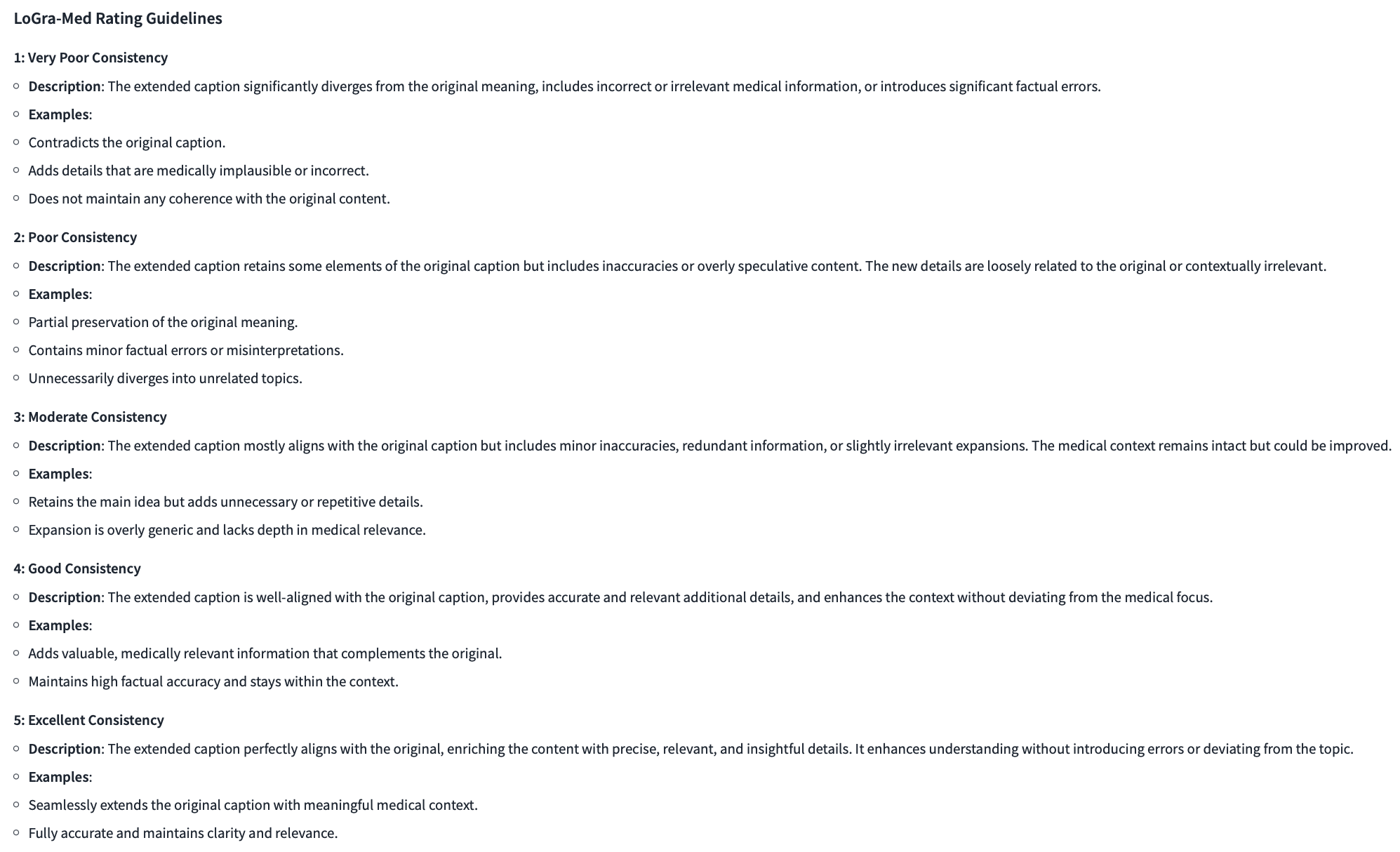}
    \vspace{-0.15in}
    \caption{(Part 2) A detailed guideline for scoring, ranging from 1 to 5, is provided.} 
    \vspace{-0.1in }
    \label{fig:rating_ui_2}
\end{figure}

\begin{figure}[H]
    \centering
    \renewcommand{\arraystretch}{1.25} % Adjust the row height
    % First Row
    \begin{minipage}[b]{0.3\textwidth}
        \begin{tikzpicture}
            \node[anchor=south west,inner sep=0] (image) at (0,0) {\includegraphics[width=\textwidth, trim=0.8cm 0 0 0, clip]{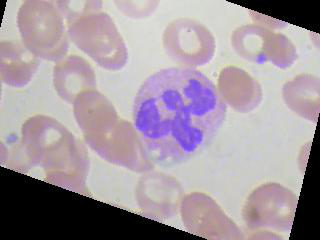}};
            \begin{scope}[x={(image.south east)},y={(image.north west)}]
            \end{scope}
        \end{tikzpicture}
        \small
        \caption*{\small Q: What are the types of cells depicted in this image?}
        \colorbox{blue}{\textcolor{white}{A: Neutrophils}} \\
        B: Melanocytes \\
        C: Lymphocytes \\
        D: Hepatocytes
    \end{minipage}
    \hspace{0.025\textwidth}
    \begin{minipage}[b]{0.3\textwidth}
        \begin{tikzpicture}
            \node[anchor=south west,inner sep=0] (image) at (0,0) {\includegraphics[width=\textwidth]{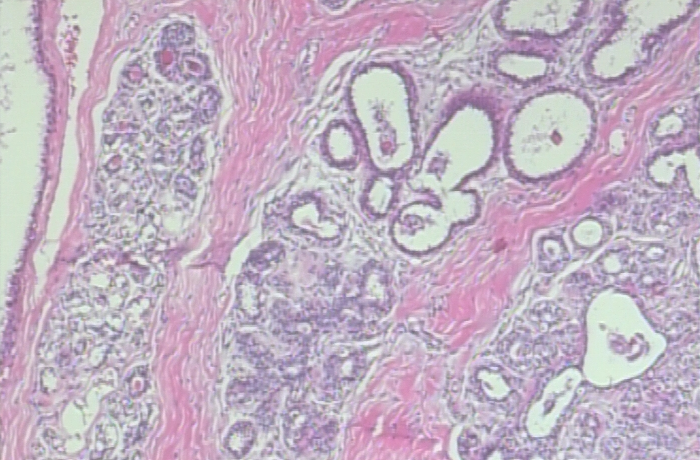}};
            \begin{scope}[x={(image.south east)},y={(image.north west)}]
                % \draw[red,thick] (0.02,0.98) rectangle (0.98,0.02);
            \end{scope}
        \end{tikzpicture}
        \caption*{\small Q: What is the diagnosis of the histopathology in this image?}
        \small
        A: Breast hyperplasia without atypia histopathology \\
        B: Normal breast histopathology \\
       \colorbox{blue}{\textcolor{white}{C: Benign breast histopathology}} \\
        D: Fibrocystic breast histopathology
    \end{minipage}
    \hspace{0.025\textwidth}
    \begin{minipage}[b]{0.3\textwidth}
        \begin{tikzpicture}
            \node[anchor=south west,inner sep=0] (image) at (0,0) {\includegraphics[width=\textwidth, trim=0 0.75cm 0 0.75cm, clip]{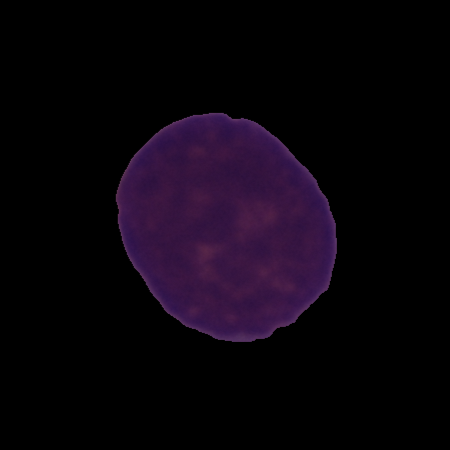}};
            \begin{scope}[x={(image.south east)},y={(image.north west)}]
                % \draw[red,thick] (0.02,0.98) rectangle (0.98,0.02);
            \end{scope}
        \end{tikzpicture}
        \caption*{\small Q: What is the probable diagnosis depicted in this image?}
        \small
        A: Chronic myeloid leukemia \\
        B: Multiple myeloma \\
        C: Hodgkin's lymphoma \\
        \colorbox{blue}{\textcolor{white}{D: Acute lymphoblastic leukemia}}
    \end{minipage}
    % Horizontal line separator between minipages
    \vspace{0.5cm} % Vertical space before the line
    \noindent\rule{\linewidth}{0.5pt} % Creates a horizontal line with adjustable thickness
    % \vspace{0.5cm} % Vertical space after the line

    %% 2nd Row
    \begin{minipage}[b]{0.3\textwidth}
        \begin{tikzpicture}
            \node[anchor=south west,inner sep=0] (image) at (0,0) {\includegraphics[width=\textwidth, trim=0 0.75cm 0 0.75cm, clip]{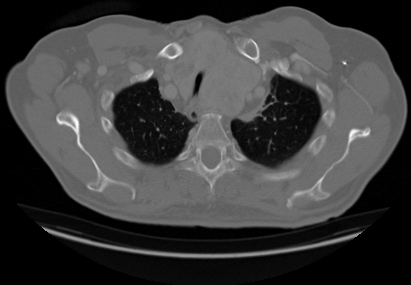}};
            \begin{scope}[x={(image.south east)},y={(image.north west)}]
            \end{scope}
        \end{tikzpicture}
        \small
        \caption*{\small Q: What is the diagnosis of the cancer seen in this image?}
        A: Adenocarcinoma of the right hilum, T3 N1 M0, Stage IIb \\
        B: Mesothelioma of the right hilum, T2 N1 M0, Stage IIb \\
        \colorbox{blue}{\textcolor{white}{C: Large cell carcinoma of the left}} \\ \colorbox{blue}{\textcolor{white}{hilum, T2 N2 M0, Stage IIIa}} \\
        D: Non-small cell carcinoma of the left hilum, T2 N0 M0, Stage I
    \end{minipage}
    \hspace{0.025\textwidth}
    \begin{minipage}[b]{0.3\textwidth}
        \begin{tikzpicture}
            \node[anchor=south west,inner sep=0] (image) at (0,0) {\includegraphics[width=\textwidth, trim=1.5cm 0 1.5cm 0, clip]{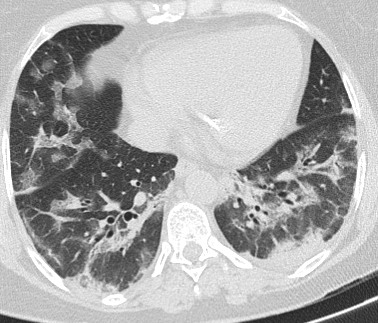}};
            \begin{scope}[x={(image.south east)},y={(image.north west)}]
                % \draw[red,thick] (0.02,0.98) rectangle (0.98,0.02);
            \end{scope}
        \end{tikzpicture}
        \caption*{\small Q: Is COVID-19 apparent in this CT scan image?}
        \small
        A: No \\
        \colorbox{blue}{\textcolor{white}{B: Yes}}
    \end{minipage}
    \hspace{0.025\textwidth}
    \begin{minipage}[b]{0.3\textwidth}
        \begin{tikzpicture}
            \node[anchor=south west,inner sep=0] (image) at (0,0) {\includegraphics[width=\textwidth, trim=3.2cm 0 3.2cm 0, clip]{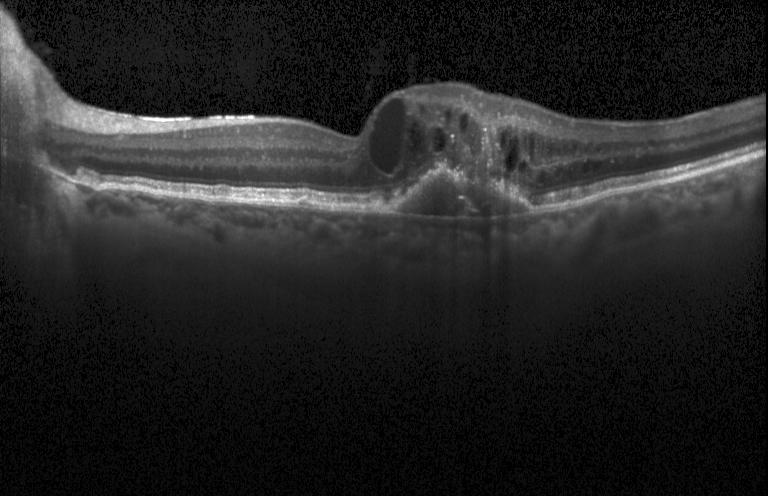}};
            \begin{scope}[x={(image.south east)},y={(image.north west)}]
                % \draw[red,thick] (0.02,0.98) rectangle (0.98,0.02);
            \end{scope}
        \end{tikzpicture}
        \caption*{\small Q: Which imaging technique was utilized to obtain this image?}
        \small
        A: Ultrasound \\
        \colorbox{blue}{\textcolor{white}{B: Optical Coherence Tomography}}
        C: Magnetic Resonance Imaging (MRI) \\
        D: Thermography
    \end{minipage}
    \vspace{0.1in}
    \caption{Examples from the OmniMedVQA dataset: microscopy (top) and CT images (bottom) with corresponding questions and options, with the correct answers highlighted in blue.}
    \label{fig:omnimed_vqa_example}
\end{figure}

In particular, we randomly chose $1000$ samples in Stage 2 of pre-training, covering five data modalities: chest X-ray, CT scan, MRI, histology, and others. Each doctor is assigned a specific data modality given their expertise, including 200 image-text pairs and corresponding captions. We then build an annotation tool for them to verify data where each sample is asked with two questions (i) whether the extended caption covers the original caption; and (ii) whether new concepts appearing in extended captions are correct. For (i) and (ii), doctors can rate with five levels (from 1 to 5), each indicating an increasing level of correctness (Figures \ref{fig:rating_ui_1}-\ref{fig:rating_ui_2}).

\begin{figure}[H]
    \centering
    % First minipage for the first image
    \begin{minipage}[b]{0.45\textwidth} % Adjust width as needed
        \centering
        \includegraphics[width=0.9\linewidth]{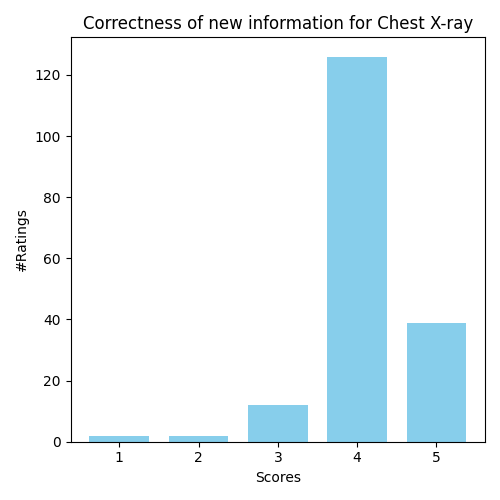} 
    \end{minipage}
    \hfill % Add horizontal space
    % Second minipage for the second image
    \begin{minipage}[b]{0.45\textwidth} % Adjust width as needed
        \centering
        \includegraphics[width=0.9\linewidth]{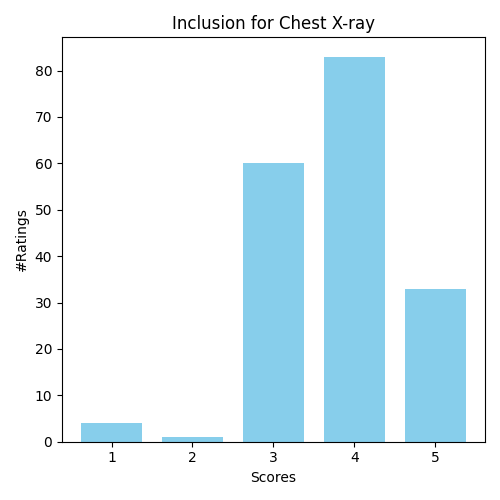} % 
    \end{minipage}
    \vspace{-0.1in }
    \caption{Statistical correctness of extended captions generated by GPT-4 on Chest X-rays.}
    \label{fig:xray}
\end{figure}

\begin{figure}[H]
    \centering
    % First minipage for the first image
    \begin{minipage}[b]{0.45\textwidth} % Adjust width as needed
        \centering
        \includegraphics[width=0.9\linewidth]{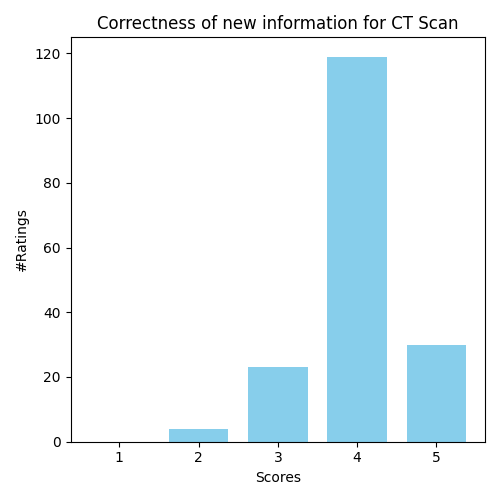} 
    \end{minipage}
    \hfill % Add horizontal space
    % Second minipage for the second image
    \begin{minipage}[b]{0.45\textwidth} % Adjust width as needed
        \centering
        \includegraphics[width=0.9\linewidth]{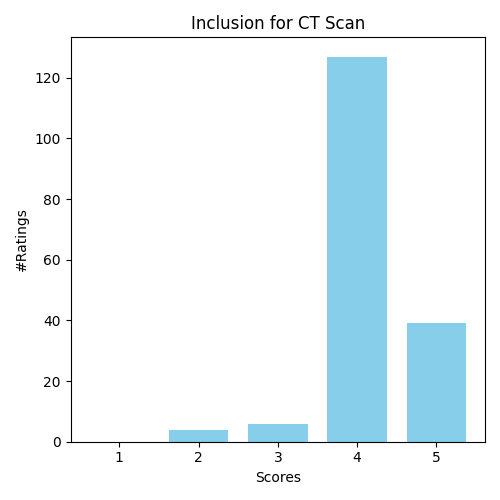} % 
    \end{minipage}
    \vspace{-0.1in }
    \caption{Statistical correctness of extended captions generated by GPT-4 on CT scans.}
    \label{fig:ct_scan}
\end{figure}

\begin{figure}[H]
    \centering
    % First minipage for the first image
    \begin{minipage}[b]{0.45\textwidth} % Adjust width as needed
        \centering
        \includegraphics[width=0.9\linewidth]{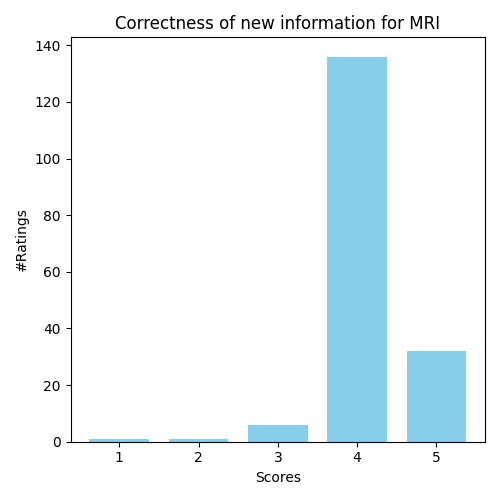} 
    \end{minipage}
    \hfill % Add horizontal space
    % Second minipage for the second image
    \begin{minipage}[b]{0.45\textwidth} % Adjust width as needed
        \centering
        \includegraphics[width=0.9\linewidth]{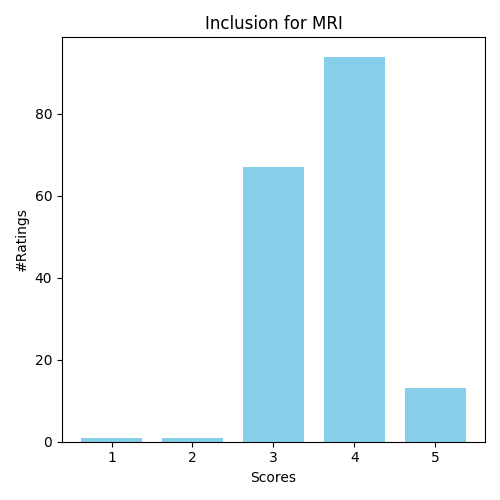} % 
    \end{minipage}
    \vspace{-0.1in }
    \caption{Statistical correctness of extended captions generated by GPT-4 on MRI.}
    \label{fig:mri}
\end{figure}

\begin{figure}[H]
    \centering
    % First minipage for the first image
    \begin{minipage}[b]{0.45\textwidth} % Adjust width as needed
        \centering
        \includegraphics[width=0.9\linewidth]{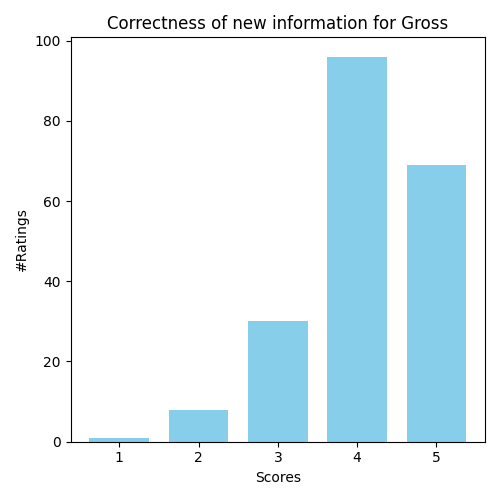} 
    \end{minipage}
    \hfill % Add horizontal space
    % Second minipage for the second image
    \begin{minipage}[b]{0.45\textwidth} % Adjust width as needed
        \centering
        \includegraphics[width=0.9\linewidth]{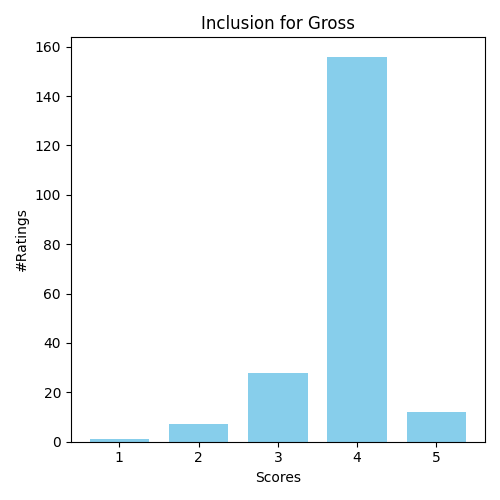} % 
    \end{minipage}
    \vspace{-0.1in }
    \caption{Statistical correctness of extended captions generated by GPT-4 on mixed domains.}
    \label{fig:gross}
\end{figure}

\begin{figure}[H]
    \centering
    % First minipage for the first image
    \begin{minipage}[b]{0.45\textwidth} % Adjust width as needed
        \centering
        \includegraphics[width=0.9\linewidth]{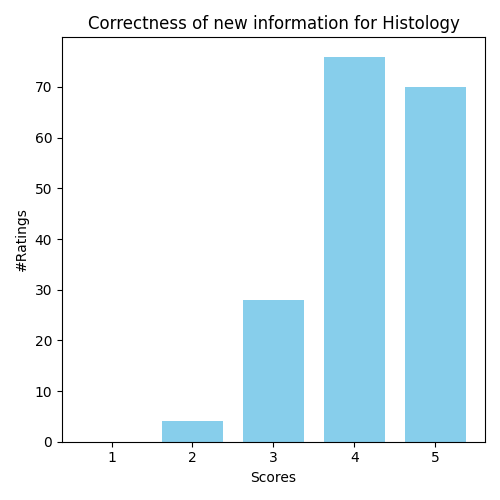} 
    \end{minipage}
    \hfill % Add horizontal space
    % Second minipage for the second image
    \begin{minipage}[b]{0.45\textwidth} % Adjust width as needed
        \centering
        \includegraphics[width=0.9\linewidth]{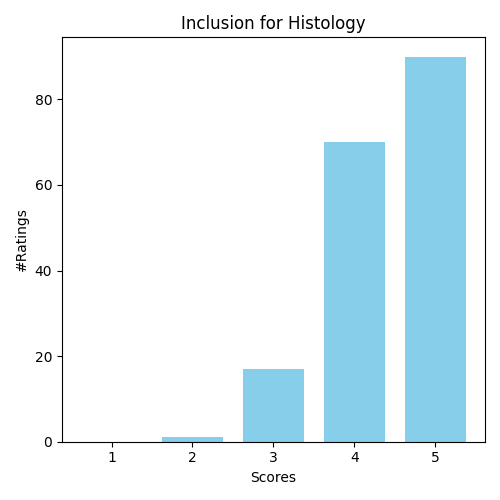} % 
    \end{minipage}
    \vspace{-0.1in }
    \caption{Statistical correctness of extended captions generated by GPT-4 on histology samples.}
    \label{fig:histology}
\end{figure}

We provide statistical correctness evaluated by general doctors for these domains in Figures \ref{fig:xray},\ref{fig:ct_scan},\ref{fig:mri}, \ref{fig:gross}, and \ref{fig:histology}. It can be seen that most rating scores fall between 3 and 5, with only a small number of samples rated 1 or 2, validating the overall consistency of GPT-4 outputs. While concerns may arise regarding the impact of low-scoring extended captions (rated 1 or 2) on the LLM, it’s important to note that these extended captions are utilized solely for contrastive learning during pre-training to align the model’s latent space representations. They are not used in auto-regressive training, which involves predicting target ground-truth tokens. Additionally, the model is fine-tuned with the given training samples from downstream tasks after pre-training (no extended captions are used). Thus, we argue that the presence of a small number of noisy extended captions should not significantly affect the performance of the LLM.

\vspace{-1cm}
\vspace{-1cm}

\end{document}